\documentclass[10pt,twocolumn,letterpaper]{article}

\usepackage{iccv}
\usepackage{times}
\usepackage{epsfig}
\usepackage{graphicx}
\usepackage{amsmath}
\usepackage{amssymb}

\usepackage{bm}
\usepackage{array}
\usepackage{color}
\usepackage{booktabs}
\usepackage{array}
\usepackage{multirow}
\usepackage{stmaryrd}
\usepackage{colortbl}
\usepackage{threeparttable}

\usepackage{algorithmicx,algorithm}
\usepackage{algpseudocode}
\usepackage{makecell}
\usepackage{marvosym}
\usepackage{enumitem}
\usepackage{subcaption}
\usepackage[accsupp]{axessibility}  

\usepackage{boxedminipage}

\usepackage{colortbl}
\definecolor{tbgray}{gray}{.9}
\newcommand{\gray}[1]{\textcolor[rgb]{0.753, 0.753, 0.753}{#1}}
\newcommand{\rc}[0]{\rowcolor[gray]{.9}}

\newcommand{\cparagraph}[1]{\smallskip\noindent\textbf{#1}}

\usepackage[pagebackref=true,breaklinks=true,letterpaper=true,colorlinks,bookmarks=false]{hyperref}

\iccvfinalcopy 

\def\iccvPaperID{5330} 

\ificcvfinal\fi

\begin{document}

\title{Local Context-Aware Active Domain Adaptation}

\author{Tao Sun\\
Stony Brook University\\
{\tt\small tao@cs.stonybrook.edu}
\and
Cheng Lu\\
XPeng Motors\\
{\tt\small luc@xiaopeng.com}
\and
Haibin Ling\\
Stony Brook University\\
{\tt\small hling@cs.stonybrook.edu}
}

\maketitle
\ificcvfinal\thispagestyle{empty}\fi

\pagenumbering{arabic}

\begin{abstract}
   Active Domain Adaptation (ADA) queries the labels of a small number of selected target samples to help adapting a model from a source domain to a target domain. The local context of queried data is important, especially when the domain gap is large. However, this has not been fully explored by existing ADA works. In this paper, we propose a Local context-aware ADA framework, named LADA, to address this issue. To select informative target samples, we devise a novel criterion based on the local inconsistency of model predictions. Since the labeling budget is usually small, fine-tuning model on only queried data can be inefficient. We progressively augment labeled target data with the confident neighbors in a class-balanced manner. Experiments validate that the proposed criterion chooses more informative target samples than existing active selection strategies. Furthermore, our full method clearly surpasses recent ADA arts on various benchmarks. Code is available at \url{https://github.com/tsun/LADA}.
\end{abstract}

\section{Introduction}
Unsupervised Domain Adaptation (UDA)~\cite{ganin2016domain,liu2021cycle} adapts a model from a related source domain to an unlabeled target domain. It has been widely studied in the past decade. Despite its success in many applications, UDA is still a challenging task, especially when the domain gap is large~\cite{sun2022safe}. In practical scenarios, it is often allowable to annotate a small number of unlabeled data. The new paradigm of Active Domain Adaptation (ADA), which queries the label of selected target samples to assist domain adaptation, draws increasing attention recently due to its promising performance with minimal labeling cost~\cite{su2020active,prabhu2021active,xie2022learning,xie2021active}.

Traditional Active Learning (AL) methods select unlabeled samples that are \emph{uncertain} to the model~\cite{joshi2012scalable} or \emph{representative} to the data distribution~\cite{sener2017active}. Some works combine these two principles to design \textit{hybrid} criteria~\cite{joshi2012scalable}. These AL strategies, however, may be less effective in ADA due to the availability of labeled source data and the distribution shift between source and target domains. Recent ADA works seek to use density weighted entropy~\cite{su2020active}, combine transferable criteria~\cite{fu2021transferable}, focus on hard examples~\cite{xie2022learning} or exploit free energy biases~\cite{xie2021active} to select target samples that are beneficial to domain adaptation. 

\begin{figure*}[!t]
	\begin{center}
		\centering
		\includegraphics[width=0.85\linewidth]{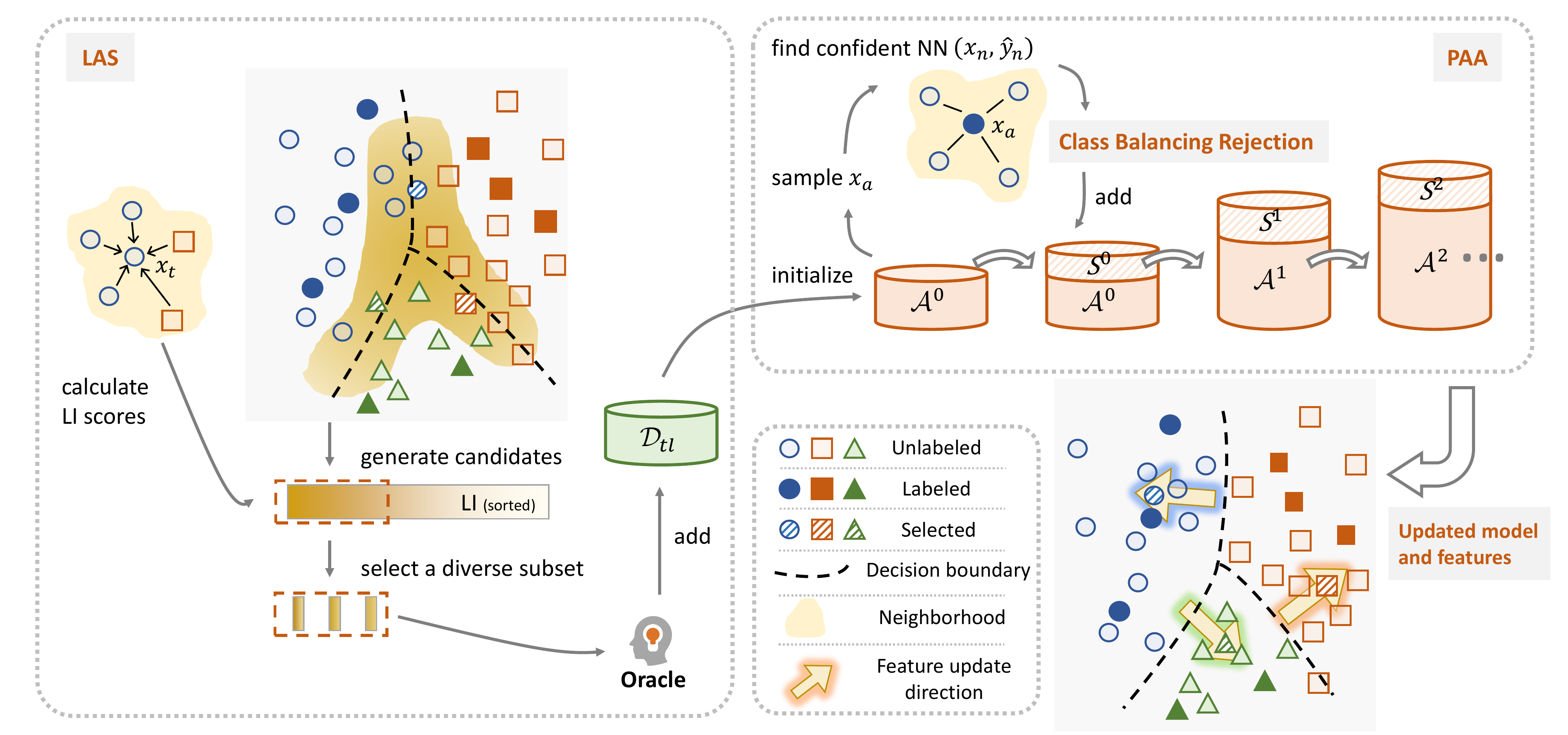}
	\end{center}
	\vspace{-4.5mm}
	\caption{The framework of LADA. For active querying, the LAS (Local context-aware Active Selection) module selects informative target samples based on the Local Inconsistency (LI) of model predictions. For model adaptation, the PAA (Progressive Anchor set Augmentation) module exploits all queried data and their confident neighbors to fine-tune the model. The anchor set $\mathcal{A}$ is expanded progressively during each training epoch and in a class balanced manner.
	}    
	\label{fig:framework}
	\vspace{-3.mm}
\end{figure*}

Despite recent progress in ADA, the local context of queried data has not been fully explored. Different from Semi-Supervised Domain Adaptation (SSDA)~\cite{saito2019semi,li2021cross} where all labeled target data are given at the beginning of training and fixed afterwards, active querying to obtain labeled target data and model update interleave during the training of ADA. The local context can guide the selection of target samples that are uncertain and locally representative. It can also be utilized to update models and reduces the tendency to only memorize these newly queried data during fine-tuning~\cite{zhang2021understanding}. Thus later training rounds can focus on harder cases. In the particular situation when the domain gap is large, it is also safer to trust the neighbors of queried data than other confident but distant samples.

Thus motivated, in this paper, we propose a novel framework of Local context-aware Active Domain Adaptation (LADA). A Local context-aware Active Selection (LAS) module is first designed, with a novel criterion based on the local inconsistency of model predictions. During the active selection stage, a diverse subset from the uncertain regions measured by our criterion is selected for querying labels. Then we design a Progressive Anchor-set Augmentation (PAA) module to overcome issues from the small size of queried data. Since the labeling budget for each round is usually small, it only requires a small fraction of training epoch before the model predicts well on the newly queried data. Another issue is that the labeled data can be imbalanced when the target data are long-tailed or the active selection focuses on partial classes. Our PAA handles the above-mentioned issues through augmenting labeled target data. Specifically, during each training epoch, we initialize an anchor set with all available queried data and progressively augment it with pseudo-labeled confident target data in a class-balanced manner. Target training batches are sampled from the anchor set instead to fine-tune the model. We show that choosing confident samples in the neighborhood of queried data  overcomes the adversarial effects from false confident samples that are distant to labeled data.

To demonstrate the effectiveness of exploiting local context in ADA, we conduct extensive experiments on various domain adaptation benchmarks. We implement several representative active selection strategies, and compare them with LAS under the same configurations. Either with the simple model fine-tuning or a strong SSDA method named MME~\cite{saito2019semi}, LAS consistently selects more informative samples, leading to higher accuracies. Equipped with PAA, the full LADA method outperforms state-of-the-art ADA solutions on various benchmarks on both standard datasets and datasets with class distribution shift.

In summary, we make the following contributions: 
\begin{itemize}[itemsep=0pt,topsep=0pt,parsep=0pt]
	\item We advocate to utilize the local context of queried data in ADA, which may guide active selection and improve model adaptation.
	\item We propose a LAS module with a novel active criterion based on the local inconsistency of class probability predictions. It selects more informative samples than existing active selection criteria.
	\item We design a PAA module to overcome issues from the small size of queried data. It progressively supplements labeled target data with confident samples in a class-balanced manner.
	\item Extensive experiments show that the full LADA method outperforms state-of-the-art ADA solutions.
\end{itemize}

\section{Related Work}

\vspace{-1mm}\paragraph{Domain Adaptation.} 
Domain Adaptation aims to adapt a model from a source domain to a related but different target domain. Early works seek to align the feature distributions through matching statistics~\cite{tzeng2014deep} or adopt adversarial learning~\cite{ganin2016domain,sun2023domain}. Recent studies show their intrinsic limitations when label distribution shift (LDS) exists~\cite{li2020rethinking,prabhu2021sentry}. Aligning feature distributions could also fail to learn discriminative class boundaries in SSDA where a few labeled target data are available~\cite{saito2019semi,yang2021deep}. Self-training~\cite{liu2021cycle,sun2022safe} that learns from model predictions has become a promising approach in UDA with LDS~\cite{li2020rethinking}, SSDA~\cite{yang2021deep}, source-data free UDA~\cite{liang2020we,sun2022prior}, \etc. Exploiting the structure of target data to improve adaptation performance is also an important research topic~\cite{tang2020unsupervised,yang2021exploiting,li2021cross}. Since UDA can be challenging when the domain gap is large, some recent works try to explore Transformer~\cite{dosovitskiy2020image} architecture for stronger feature representation~\cite{sun2022safe}, and leverage prior knowledge of target class distribution to rectify model predictions~\cite{sun2022prior}.


\vspace{-3mm}\paragraph{Active Learning.}
Active Learning (AL) selects a small number of unlabeled samples for annotation. There are two major paradigms: \emph{uncertainty}-based methods select uncertain samples with criteria like entropy, confidence~\cite{lewis1994heterogeneous}, prediction margin~\cite{balcan2007margin}, \textit{etc.}; \emph{representativeness}-based methods select samples that can represent the entire data distribution. Popular methods include clustering~\cite{xu2003representative} and CoreSet~\cite{sener2017active}. While most traditional AL algorithms use a one-by-one query method, deep AL~\cite{ren2021survey} adopts batch-based query to avoid extra computation and model overfitting~\cite{ash2019deep,kim2021task}. Recently, the ideas of Bayesian learning~\cite{kirsch2019batchbald}, adversarial learning~\cite{sinha2019variational}, and reinforcement learning~\cite{liu2019deep} have also been introduced into deep AL.  

\vspace{-3mm}\paragraph{Active Domain Adaptation.}
Active Domain Adaptation queries the label of selected target samples to assist domain adaptation. It was first studied in the task of sentiment classification from text data~\cite{rai2010domain}, where samples are selected with uncertainty and a domain separator. More recently, AADA~\cite{su2020active} studies ADA in vision tasks. It combines active learning with adversarial domain adaptation and selects samples with a diversity cue from importance weight plus an uncertainty cue.  CLUE~\cite{prabhu2021active} performs uncertainty-weighted clustering to select samples that are both uncertain and diverse. S$^3$VAADA~\cite{rangwani2021s3vaada} uses a submodular criterion combining three scores to select a subset.  TQS~\cite{fu2021transferable} adopts an ensemble of transferable committee, transferable uncertainty and transferable domainness as the selection criterion. EADA~\cite{xie2021active} trains an energy-based model, and selects samples based on energy values. SDM~\cite{xie2022learning} focuses on hard cases. It consists of a maximum margin loss and a margin sampling function. LAMDA~\cite{hwang2022combating} addresses the issue of label distribution mismatch in ADA by seeking target data that best approximate the entire target distribution.  

Although various ADA methods have been proposed, the local context of queried data has not been fully explored. The closest to ours is~\cite{wang2023mhpl} that proposes a Minimum Happy (MH) points learning for Active Source Free Domain Adaptation. The MH points are selected to have low neighbor purity and high neighbor affinity, where purity is measured with the entropy of hard neighbor labels. Different from theirs, we use the local inconsistency of class probability predictions to keep more information. Furthermore, they adopt one-shot querying in order to determine source-like samples as source data are unavailable, while we follow previous ADA works to conduct several rounds of active selection. We show that our active learning strategy is more compact and effective under ADA setting.

\section{Local Context-Aware ADA}
\paragraph{Problem Formulation.}
In ADA, there is a source domain with labeled data $\mathcal{D}_s=\{({x}_i^s, y_i^s)\}$  and a target domain with unlabeled data $\mathcal{D}_t=\{({x}_i^t)\}$. Meanwhile, we can actively select a few target samples to query their labels under a labeling budget $B$, which is usually much smaller than the total amount of target data (\ie, $B\ll |\mathcal{D}_t|$). Let the obtained labeled target data (\textit{a.k.a.}~queried data) be $\mathcal{D}_{tl}=\{({x}_i^{tl}, y_i^{tl})\}$, with $|\mathcal{D}_{tl}|=B$. The remaining unlabeled target data are $\mathcal{D}_{tu}=\mathcal{D}_{t}\setminus \mathcal{D}_{tl}$. The goal of ADA is to learn a model $h=g\circ f$ that generalizes well on the target domain, through querying most informative samples. $f$ is a feature extractor and $g$ is a task predictor. This work focuses on $C-$way classification. Hence labels $y\in\mathcal{Y}=\{0,1,\cdots, C-1\}$ are categorical variables. In practice, it is more efficient to conduct active querying in $R$ rounds, with a label budget of $b=B/R$ for each round. 

It is worth mentioning that some ADA methods~\cite{fu2021transferable,prabhu2021active,xie2022learning} train the model using only labeled data $\mathcal{D}_s\cup \mathcal{D}_{tl}$, while others~\cite{rangwani2021s3vaada,hwang2022combating,wang2023mhpl} also require unlabeled data. When the labeling budget is small or there exists label distribution shift, the unlabeled data are critical to the model performance. For our work, we present results under both settings.


\vspace{-3mm}\paragraph{Method Overview.} 

We exploit the local context of queried data with the aim to select more informative target samples and improve model adaptation. The framework of our Local context-aware Active Domain Adaptation (LADA) is illustrated in Fig.~\ref{fig:framework}. The LAS (Local context-aware Active Selection) module sorts all unlabeled target samples by their Local Inconsistency of model predictions (LI scores), and then selects a diverse subset with the largest scores for querying their ground-truth labels. The PAA (Progressive Anchor set Augmentation) module progressively supplements query set with confident target samples. A Class Balancing strategy is adopted to add samples from different classes evenly. The two modules run alternatively for several rounds until the labeling budget is used up. Alg.~\ref{alg:LADA} summarizes the training procedure. 

\begin{figure}[!t]
\vspace{-4mm}
\begin{algorithm}[H]
	\renewcommand{\algorithmicrequire}{\textbf{Input:}}
	\renewcommand{\algorithmicensure}{\textbf{Initialization:}}
	\caption{Local context-aware ADA.} 
	\label{alg:LADA} 
	\begin{algorithmic}[1]
		\Require Source data $\mathcal{D}_{s}$, target data $\mathcal{D}_{t}$, training epochs $E$, iterations per epoch $I$
		\Ensure $\mathcal{D}_{tl}=\emptyset$, $\mathcal{D}_{tu}=\mathcal{D}_{t}$
		\For{$e=0$ \textbf{to} $E$}	
		\If{need active querying}	
		\State Obtain query set $\mathcal{D}_{q}$ with  ground-truth labels from Oracle as described in Sec.~\ref{sec:LAS}
		\State $\mathcal{D}_{tl}\leftarrow \mathcal{D}_{tl}\cup \mathcal{D}_q$, ~~ $\mathcal{D}_{tu}\leftarrow \mathcal{D}_{tu}\setminus \mathcal{D}_q$ 
		\EndIf		
		\State Initialize $\mathcal{A}=\mathcal{D}_{tl}$, $\mathcal{S}=\emptyset$
		\For{$i=0$ \textbf{to} $I$}				
		\State Sample mini-batches $\mathcal{B}_s$ from $\mathcal{D}_s$, $\mathcal{B}_a$ from $\mathcal{A}$
		\State Update model parameters with Eq.~\ref{eq:obj}
		\State Update $\mathcal{S}$ as described in Alg.~\ref{alg:PAA}.
		\If{iteration over $\mathcal{A}$ finishes}
		\State $\mathcal{A}\leftarrow \mathcal{A}\cup \mathcal{S}$, $\mathcal{S}=\emptyset$
		\EndIf	
		\EndFor
		\EndFor	
	\end{algorithmic}
\end{algorithm}
\vspace{-8mm}
\end{figure}

\subsection{Local Context-aware Active Selection}\label{sec:LAS}

\emph{Uncertainty} and \emph{Representativeness} are two general principles in traditional active learning. Uncertainty-based strategies aim to select samples whose model predictions are less confident, while representativeness-based strategies aim to select samples to better represent the entire data distribution. In ADA, the source domain provides auxiliary labeling information, making it cost-inefficient to sample in the well-aligned regions of the two domains. Our empirical experiments indicate that it is more beneficial to focus on the uncertain target samples in ADA tasks.

Many criteria (\eg, \textit{entropy} and \textit{margin}) have been proposed to select uncertain samples. However, these measures only rely on the sample-wise model prediction while ignoring its local context, which may include some outliers during active selection. To address this issue, we design a novel criterion based on the local inconsistency of model predictions. Let $p(\cdot)=\sigma(h(\cdot))$ denote the class probability after \texttt{softmax} operation {$\sigma(\cdot)$}. For an unlabeled target sample $x_{t}$, its Local Inconsistency (LI) is measured as
\begin{equation}\label{eq:Q0}
\vspace{-0.5mm}
	\mathrm{LI}(x_{t})=-\frac{1}{K}\sum_{k=1}^{K} w_{k} p(x_{tk})^{\top}p(x_{t})
 \vspace{-0.5mm}
\end{equation} 
where  $\{x_{tk}\}$ are $K$ nearest neighbors of $x_{t}$ based on the cosine similarity $\langle f(x_{tk}),f(x_{t}) \rangle$, and $w_k\propto f(x_{tk})^{\top}f(x_{t})$ is the normalized weight. Here we use the soft probability $p$, as pseudo labels are unreliable for uncertain samples near decision boundaries. When the neighbors are very close to the center sample, $\mathrm{LI}(x_{t})\approx -p(x_{t})^{\top}p(x_{t})= p(x_{t})^{\top}(1-p(x_{t}))-1$, which is essentially the Gini Impurity. In real ADA benchmarks, the data distribution is not dense and the neighborhood size $K$ can be adjusted accordingly. To removes outlier values and enhance the clustering structure of candidate uncertain regions, we additionally smooth the scores among neighborhood by
\begin{equation}
\vspace{-0.5mm}
	\mathrm{LI}(x_{t})\leftarrow \mathrm{LI}(x_{t})+\frac{1}{K}\sum_{k=1}^{K} w_{k}\mathrm{LI}(x_{tk})
  \vspace{-0.5mm}\label{eq:LI_sm}
\end{equation}

\begin{figure}[!t]
\vspace{-4mm}
\begin{algorithm}[H]
	\renewcommand{\algorithmicrequire}{\textbf{Input:}}
	\renewcommand{\algorithmicensure}{\textbf{Return:}}
	\caption{Progressive Anchor set Augmentation.} 
	\label{alg:PAA} 
	\begin{algorithmic}[1]
		\Require Confidence threshold $\tau$, reject probability $q_{\mathcal{A}}$, supplementary set $\mathcal{S}$, labeled mini-batch $\mathcal{B}_a$, target data $\mathcal{D}_{t}$, neighborhood size $K$ 
		\For {$x_a\in \mathcal{B}_a$}
		\State Randomly sample $x_n$ from the $K$ nearest neighbors of $x_a$ for LAA (or from $\mathcal{D}_{t}$ for RAA)
		\State $p_n = \sigma(h(x_n))$, $\hat{y}_n=\arg\max p_n$
		\State $\xi \sim U(0,1)$ \Comment{\gray{$U$ is uniform distribution}}
		\If{$p_n[\hat{y}_n]\geq \tau$ and $\xi \geq q_{\mathcal{A}}[\hat{y}_n]$}
		\State $\mathcal{S}\leftarrow \mathcal{S}\cup \{(x_n,\hat{y}_n)\}$
		\EndIf		
		\EndFor		
		\State Return $\mathcal{S}$
	\end{algorithmic}
\end{algorithm}
\vspace{-8mm}
\end{figure}

\begin{table*}[!t]
	\caption{Accuracies (\%) on \textbf{Office-Home} using 5\%-budget. (The highest accuracies across all methods are \textbf{bolded} and within each section are \underline{underlined}.)}
	\footnotesize
	\centering
 \vspace{-2mm}
	\scalebox{0.9}{
		\begin{tabular}{p{2cm}@{}p{2.5cm}@{}p{1.05cm}<{\centering}@{}p{1.05cm}<{\centering}@{}p{1.05cm}<{\centering}@{}p{1.05cm}<{\centering}@{}p{1.05cm}<{\centering}@{}p{1.05cm}<{\centering}@{}p{1.05cm}<{\centering}@{}p{1.05cm}<{\centering}@{}p{1.05cm}<{\centering}@{}p{1.05cm}<{\centering}@{}p{1.05cm}<{\centering}@{}p{1.05cm}<{\centering}@{}p{1.05cm}<{\centering}}
			\toprule
			AL method & DA method & Ar$\shortrightarrow$Cl & Ar$\shortrightarrow$Pr & Ar$\shortrightarrow$Rw &  Cl$\shortrightarrow$Ar & Cl$\shortrightarrow$Pr & Cl$\shortrightarrow$Rw &  Pr$\shortrightarrow$Ar & Pr$\shortrightarrow$Cl & Pr$\shortrightarrow$Rw &      
			Rw$\shortrightarrow$Ar & Rw$\shortrightarrow$Cl & Rw$\shortrightarrow$Pr & Avg.   \\ 	
			\midrule
			RAN & \multirow{9}{*}{ft w/ CE loss} & 58.9 & 77.6 & 78.7 & 61.9 & 74.2 & 73.0 & 62.9 & 56.0 & 77.9 & 70.1 & 60.1 & 83.4 & 69.6 \\ 
			ENT &  & 61.3 & 81.1 & \underline{82.4} & 64.9 & 78.6 & 77.0 & 64.0 & 58.8 & 81.7 & 73.5 & 61.9 & 87.1 & 72.7 \\ 
			MAR &  & 62.2 & 81.5 & \underline{82.4} & 64.4 & 79.5 & 77.2 & 64.2 & 60.5 & 81.7 & 73.7 & 64.2 & 87.5 & 73.3 \\ 
			CoreSet &  & 58.2 & 77.7 & 79.2 & 61.7 & 73.6 & 73.3 & 60.6 & 55.1 & 78.9 & 70.3 & 60.0 & 82.8 & 69.3 \\ 
			BADGE &  & 63.4 & 81.9 & 81.5 & 65.1 & 79.9 & 77.0 & 64.4 & 61.0 & 82.0 & 73.8 & 63.5 & 88.0 & 73.5 \\ 
			AADA &  & 60.4 & 81.5 & 82.3 & 64.5 & 79.0 & 76.9 & 63.2 & 58.9 & 81.9 & 73.3 & 63.2 & 87.4 & 72.7 \\ 
			CLUE &  & 63.0 & 81.7 & 81.1 & 63.2 & 79.3 & 76.2 & 64.6 & 59.7 & 81.5 & 73.1 & 63.5 & 86.8 & 72.8 \\
			TQS &  & 58.6 & 81.1 & 81.5 & 61.1 & 76.1 & 73.3 & 61.2 & 54.7 & 79.7 & 73.4 & 58.9 & 86.1 & 70.5 \\ 
			MHPL & & 63.5 &	81.7 &	82.1 &	65.0 &	79.2 &	77.2 &	65.0 &	61.7 &	82.2 &	73.8 &	65.1 &	87.7 &	73.7 \\
			\rc
			LAS & & \underline{66.1} & \underline{83.8} & \underline{82.4} & \underline{66.9} & \underline{82.5} & \underline{78.6} & \underline{66.8} & \underline{63.1} & \underline{82.4} & \underline{74.9} & \underline{66.7} & \underline{89.6} & \underline{75.3} \\		
			\midrule
			RAN & \multirow{5}{*}{MME} & 64.0 & 82.3 & 81.2 & 68.4 & 80.8 & 77.3 & 68.2 & 63.5 & 82.0 & 76.0 & 65.1 & 86.2 & 74.6 \\
			AADA &   & 64.8 & 84.2 & 84.7 & 71.8 & 82.3 & 80.4 & {71.7} & 62.7 & 85.1 & 79.2 & 67.2 & 88.6 & 76.9 \\ 
			CLUE &   & 65.5 & 84.9 & 83.8 & 71.1 & 82.4 & 79.5 & 71.4 & 62.9 & 85.2 & 79.0 & 66.9 & 88.4 & 76.7 \\ 
			MHPL &  & 66.8 &	82.3 &	84.0 &	71.1 &	84.2 &	\underline{\textbf{80.6}} &	\underline{\textbf{71.9}} &	65.5 &	84.5 &	78.1 &	67.6 &	89.3 &	77.2 \\	
			\rc
			LAS & & \underline{\textbf{68.6}} & \underline{\textbf{86.7}} & \underline{\textbf{85.0}} & \underline{\textbf{72.1}} & \underline{\textbf{84.6}} & \underline{\textbf{80.6}} & 71.4 & \underline{\textbf{65.6}} & \underline{\textbf{85.5}} & \underline{\textbf{79.4}} & \underline{\textbf{68.4}} & \underline{\textbf{89.9}} & \underline{\textbf{78.2}} \\ 	
			\bottomrule
	\end{tabular} }
	\label{tab:officehome}
	\vspace{-1mm}
\end{table*}

\begin{table*}[!t]
	\caption{Accuracies (\%) on \textbf{Office-Home} using 10\%-budget. ($^\dagger$Using importance sampling)}
	\footnotesize
	\centering
\vspace{-2mm}
	\scalebox{0.9}{
		\begin{tabular}{p{2cm}@{}p{2.5cm}@{}p{1.05cm}<{\centering}@{}p{1.05cm}<{\centering}@{}p{1.05cm}<{\centering}@{}p{1.05cm}<{\centering}@{}p{1.05cm}<{\centering}@{}p{1.05cm}<{\centering}@{}p{1.05cm}<{\centering}@{}p{1.05cm}<{\centering}@{}p{1.05cm}<{\centering}@{}p{1.05cm}<{\centering}@{}p{1.05cm}<{\centering}@{}p{1.05cm}<{\centering}@{}p{1.05cm}<{\centering}}
			\toprule
			AL method & DA method & Ar$\shortrightarrow$Cl & Ar$\shortrightarrow$Pr & Ar$\shortrightarrow$Rw &  Cl$\shortrightarrow$Ar & Cl$\shortrightarrow$Pr & Cl$\shortrightarrow$Rw &  Pr$\shortrightarrow$Ar & Pr$\shortrightarrow$Cl & Pr$\shortrightarrow$Rw &      
			Rw$\shortrightarrow$Ar & Rw$\shortrightarrow$Cl & Rw$\shortrightarrow$Pr & Avg.   \\ 		 
			\midrule
			BADGE & \multirow{5}{*}{ft w/ CE loss}  &  71.3 & 88.6 & 86.3 & \underline{70.6} & {86.7} & 82.0 & 71.6 & 69.7 & \underline{86.7} & \underline{79.8} & 72.4 & 91.9 & 79.8 \\
   			AADA &  &  69.0 & 87.5 & \underline{87.4} & 70.1 & 85.3 & 81.8 & 70.3 & 67.2 & \underline{86.7} & 79.1 & 69.6 & 90.6 & 78.7 \\ 
			CLUE &  & 69.7 & 87.6 & 85.7 & 69.8 & 86.3 & 81.0 & 69.8 & 68.4 & 85.5 & 78.1 & 71.8 & 91.1 & 78.7 \\ 
			TQS &  & 68.0 & 87.7 & 85.7 & 67.0 & 83.0 & 78.7 & 69.3 & 64.5 & 83.9 & 77.8 & 68.9 & 90.6 & 77.1 \\
			\rc
			LAS & &  \underline{73.6} & \underline{90.0} & 87.0 & \underline{70.6} & \underline{88.7} & \underline{82.6} & \underline{72.4} & \underline{71.8} & 86.3 & 79.3 & \underline{73.8} & \underline{92.2} & \underline{80.7} \\ 
			\midrule
                BADGE & \multirow{4}{*}{CDAC} & 68.8 & 86.4 & 84.4 & 75.3 & 88.2 & 83.6 & 75.2 & 71.3 & 87.1 & 79.6 & 73.5 & 91.3 & 80.4 \\ 
                AADA & & 69.7 & 87.0 & 86.2 & 74.8 & 87.0 & 84.5 & 74.8 & 69.7 & 88.1 & 79.7 & 71.2 & 90.6 & 80.3 \\ 
                CLUE & & 72.3 & 87.6 & 85.8 & 74.0 & 88.6 & 84.3 & 74.4 & 72.6 & 87.2 & 79.4 & 73.3 & 91.1 & 80.9 \\
                \rc
                LAS &  & \underline{73.6} & \underline{89.3} & \underline{86.8} & \underline{76.3} & \underline{89.6} & \underline{85.9} & \underline{76.8} & \underline{74.9} & \underline{88.5} & \underline{81.1} & \underline{76.5} & \underline{92.3} & \underline{82.6} \\
			\midrule
			TQS & \multirow{3}{*}{DANN$^\dagger$} & 68.7 & 80.1 & 83.1 & 64.0  & 83.1  & 76.9 & 67.7 & 71.0 & 84.4 & 76.4 & 72.7 & 90.0 & 76.5 \\
			S$^3$VAADA	&  & 65.5 & 79.6 & 80.0 & 65.4 & 82.2 & 75.5 & 68.4 & 68.1 & 84.0 & 73.5 & 70.7 & 88.6 & 75.1 \\
			LAMDA &  & 74.8 & 88.5 & 86.9 & 73.8 & 88.2 & 83.3 & 74.6 & 75.5 & 86.9 & 80.8 & 77.8 & 91.7 & 81.9\\
			\midrule               
                 & MCC & 77.2 & 91.0 & \textbf{88.6} & 77.1 & 90.7 & 86.8 & 76.4 & 76.4 & 89.1 & 81.9 & 77.7 & 93.3 & 83.9 \\  
			\rc
			 \multirow{1}{*}{LAS}  & RAA & \textbf{77.8} &	91.8 &	88.4 &	\textbf{77.7} &	\textbf{91.5} &	\textbf{87.7} &	\textbf{78.1} &	\textbf{79.1} &	\textbf{89.5} &	\textbf{83.4} &	\textbf{79.8} &	\textbf{94.1} &	\textbf{84.9} \\
			\rc
			 & LAA &  77.2 &	\textbf{91.9} &	88.1 &	76.9 &	91.1 &	86.8 &	76.6 &	78.1 &	88.3 &	82.0 &	79.0 &	93.8 &	84.2 \\	 
			\bottomrule
	\end{tabular} }
	\label{tab:officehome_p10}
	\vspace{-4mm}
\end{table*}

Data with high $\mathrm{LI}$ scores generally have large sample-wise uncertainty and inconsistent predictions to neighbors. An adjoint effect is that those highly scored target samples tend to co-occur in specific local regions. Active selection solely based on $\mathrm{LI}$ scores would include highly similar sample pairs, leading to a waste of labeling budgets. Fortunately, we can select a diverse subset from an over-sampled candidate set. Let $b$ be the labeling budget for current round and $M$ be a predefined ratio. A clustering process is run on the top $(1+M)b$ samples with highest $\mathrm{LI}$ scores, and then the $b$ cluster centroids are selected as the query set. The diverse sampling can also be implemented with determinant point process, though clustering is simple and works good enough. Note that unlike CLUE~\cite{prabhu2021active} that runs an uncertainty-weighted clustering on all unlabeled target data, ours focuses only on a small portion of uncertain samples and runs more efficient.

\subsection{Progressive Anchor Set Augmentation} \label{sec:LMA}

After each round of active querying, the model can be updated for 1-2 epochs. A common way is to fine-tune the model using all labeled data~\cite{fu2021transferable,xie2022learning,xie2021active} with the objective\footnote{Some works sample training data from a joint labeled set $\mathcal{D}_{l}=\mathcal{D}_s\cup \mathcal{D}_{tl}$, which is more efficient but less generalizable.}
\begin{equation}\label{eq:losso}
	\mathcal{L}=\mathbb{E}_{(x,y)\sim \mathcal{D}_s}\ell_{\rm ce}(h(x),y)+\mathbb{E}_{(x,y)\sim \mathcal{D}_{tl}}\ell_{\rm ce}(h(x),y)
\end{equation}
where $\ell_{\rm ce}$ is the cross entropy loss. 

The number of queried data for each round is usually small, \eg, only 1\% target data are added to $\mathcal{D}_{tl}$ after active selection when the total labeling budget is 5\%. This causes two issues: it only requires a small fraction of training epoch before the model predicts well on those queried samples; $\mathcal{D}_{tl}$ can be class imbalanced when the target data are imbalanced or the active selection focuses on a few classes. Consequently, it fails to fully utilize the training resources and to provide more informative target samples to the following rounds of active selection. 


To supplement the limited queried data, we propose a Progressive Anchor set Augmentation (PAA) module to incorporate confident target data. An anchor set $\mathcal{A}$ is initialized with current $\mathcal{D}_{tl}$. At each training iteration, a target mini-batch $\mathcal{B}_a$ is sampled from $\mathcal{A}$ instead of $\mathcal{D}_{tl}$ for supervised training. Then some selected confident samples in the neighborhood of $\mathcal{B}_a$ are added to a temporal set $\mathcal{S}$. Once the sampling iteration over $\mathcal{A}$ concludes, we reinitialize the anchor set as $\mathcal{A}\leftarrow\mathcal{A}\cup \mathcal{S}$ and clear $\mathcal{S}$. The same procedure repeats until the end of this epoch. We name this as Local context-aware Anchor set Augmentation (LAA) as it exploits the local region of queried data. We also consider Random Anchor set Augmentation (RAA), where confident samples are randomly selected from the entire $\mathcal{D}_{t}$.

To obtain a class balanced anchor set, we reject a confident target sample with a higher probability if there already exist many data from the same class in current $\mathcal{A}$. Specifically, let $p_{\mathcal{A}}[c]=\sum_{(x_a,y_a)\in \mathcal{A}} \mathbb{I}[y_a=c]/|\mathcal{A}|$, $p_{\mathcal{A}}^{\rm max}=\max_{c}p_{\mathcal{A}}[c]$ and $p_{\mathcal{A}}^{\rm min}=\min_{c}p_{\mathcal{A}}[c]$. For a confident target sample with pseudo label $c$, we reject it with a probability
\begin{equation}
 \vspace{-0.5mm}
	q_{\mathcal{A}}[c]=\frac{p_{\mathcal{A}}[c]-p_{\mathcal{A}}^{\rm min}}{p_{\mathcal{A}}^{\rm max}}
 \vspace{-0.5mm}
\end{equation}
The detailed procedure of progressive anchor set augmentation is described in Alg.~\ref{alg:PAA}.

The anchor set $\mathcal{A}$ contains not only all labeled target data but also some pseudo-labeled confident data. With the class balancing rejection, $\mathcal{A}$ is expected to contain adequate number of target samples from each class, thus effectively overcome the drawbacks from the small size of queried data. Later rounds of active selection may focus on more challenging samples. To further exploit $\mathcal{A}$, we incorporate a strong image transformation with RandAugment~\cite{cubuk2020randaugment}.   The objective becomes
\begin{equation}\label{eq:obj}
	\mathcal{L}=\mathbb{E}_{(x,y)\sim \mathcal{D}_s}\ell_{\rm ce}(h(x),y)+\mathbb{E}_{(x,y)\sim {\mathcal{A}}}\ell_{\rm ce}(h(\tilde{x}),y)
\end{equation}
where $\tilde{x}=\beta x + (1-\beta)\alpha(x) $ for some random augmentation operator $\alpha(\cdot)$, and $\beta\sim Beta(0.2,0.2)$ is a random variable. It can be viewed as a specific kind of \textit{Mixup} between a weak augmentation and a strong augmentation of the same image. Comparing Eq.~\ref{eq:obj} to Eq.~\ref{eq:losso}, we see that the memory usage and computation cost are not increased.

It is worth mentioning that $\mathcal{A}$ is reinitialized as $\mathcal{D}_{tl}$ at each training epoch. This reduces cumulative biases from noisy pseudo labels. Another reason is that after each round of active selection, restarting from $\mathcal{D}_{tl}$ ensures to train the model sufficiently on the newly queried target samples.

\vspace{-1mm}
\section{Experiments}
We conduct experiments on four widely-used domain adaptation benchmarks:  \textbf{Office-31}~\cite{saenko2010adapting}, \textbf{Office-Home}~\cite{venkateswara2017deep}, \textbf{VisDA}~\cite{peng2017visda} and \textbf{DomainNet}.  \textbf{Office-Home RSUT}~\cite{tan2019generalized} is a subset of Office-Home created with the protocol of Reverse-unbalanced Source and Unbalanced Target to have a large label distribution shift. 

All experiments are implemented with Pytorch. In consistent with previous works~\cite{fu2021transferable,xie2021active,xie2022learning}, we use a pre-trained ResNet-50~\cite{he2016deep} backbone and perform 5 rounds of active selection with $B=5\%$ or $B=10\%$ of all target data. For a fair comparison, we reproduce the results of several traditional active learning criteria including random (RAN), least confidence (CONF), entropy (ENT), prediction margin (MAR)~\cite{joshi2012scalable}, CoreSet~\cite{sener2017active}, BADGE~\cite{ash2019deep}, and ADA methods including AADA~\cite{su2020active} and CLUE~\cite{prabhu2021active} in a unified framework. Results of recent state-of-the-art ADA methods like S$^3$VAADA~\cite{rangwani2021s3vaada}, TQS~\cite{fu2021transferable} and LAMDA~\cite{hwang2022combating} are borrowed from the corresponding papers whenever applicable.

We set the confidence threshold $\tau$ to 0.9 for all datasets. The neighborhood size $K$ is 5 for Office-31 and 10 for other datasets. We set $M$ with an empirical formulation $\lceil \frac{55}{100B}-1\rceil$, leading to a candidate set of about $\sim$12\% unlabeled target data. Since VisDA has a huge number of data in each class, we reduce the empirical value by a half. Additional implementation details and analyses can be found in the supplementary.

\begin{table}[t]
	\caption{Accuracies (\%) on \textbf{Office-31} using 5\%-budget.} 
	\centering
	\footnotesize
	\scalebox{0.85}{
		\begin{tabular}{p{1.2cm}p{1.0cm}p{0.52cm}<{\centering}p{0.52cm}<{\centering}p{0.52cm}<{\centering}p{0.52cm}<{\centering}p{0.52cm}<{\centering}p{0.52cm}<{\centering}p{0.52cm}<{\centering} }
			\toprule
			AL & DA & A$\shortrightarrow$D & A$\shortrightarrow$W & D$\shortrightarrow$A & D$\shortrightarrow$W & W$\shortrightarrow$A & W$\shortrightarrow$D & Avg. \\ 
			\midrule
			RAN &  \multirow{10}{*}{\makecell[c]{ft w/ \\ CE loss}} &  82.5 & 87.6 & 73.4 & 98.6 & 76.2 & 99.6 & 86.3 \\  
			ENT &   &  90.8 & 92.5 & 75.8 & 99.9 & 76.9 & \underline{\textbf{100.}} & 89.3  \\ 
			MAR &   &  89.4 & 92.5 & 78.5 & 99.8 & 79.0 & 99.9 & 89.9 \\ 
			CoreSet &   & 83.8 & 85.9 & 74.8 & 97.6 & 76.1 & 99.6 & 86.3 \\ 
			BADGE &   &88.0 & 90.9 & 79.2 & 99.8 & 80.5 & 99.9 & 89.7 \\  
			AADA &   & 88.7 & 91.4 & 76.3 & \underline{\textbf{100.}} & 77.3 & \underline{\textbf{100.}} & 88.9 \\ 
			CLUE &   & 89.8 & 91.8 & 79.1 & \underline{\textbf{100.}} & 79.8 & 99.9 & 90.1\\ 
			TQS &   & \underline{92.8} & 92.2 & 80.6 & \underline{\textbf{100.}} & 80.4 & \underline{\textbf{100.}} & 91.1 \\  
			MHPL & & 90.0 &	91.0 &	78.3 &	99.4 &	78.7 &	99.9 &	89.5  \\
			\rc
			LAS &  & 91.6 &	\underline{93.9} &	\underline{81.5} &	99.7 &	\underline{81.8} &	99.6 &	\underline{91.4}  \\  
             \midrule
                 RAN & \multirow{5}{*}{MME} & 89.9 & 92.6 & 78.1 & 99.2 & 78.9 & 99.8 & 89.7 \\
                 AADA & & 95.9 & 94.1 & 81.4 & \underline{99.5} & 81.3 & \underline{\textbf{100.}} & 92.0 \\ 
                 CLUE & & 96.1 & \underline{96.3} & 82.1 & 99.3 & 82.0 & \underline{\textbf{100.}} & 92.6 \\ 
                 MHPL & & 95.2 &	95.6 &	82.2 &	99.3 &	82.1 &	\textbf{100.} &	92.4 \\
                 \rc
                 LAS & & \underline{96.7} &	96.1 &	\underline{84.8} &	99.3 &	\underline{84.8} &	\underline{\textbf{100.}} &	\underline{93.6} \\ 
			\midrule
              RAN & \multirow{5}{*}{CDAC} &  90.6 & 91.7 & 75.0 & 98.2 & 77.0 & 99.7 & 88.7 \\
              AADA & & 94.7 & 96.1 & 77.8 & 99.5 & 79.6 & 99.5 & 91.2 \\ 
              CLUE  & & 94.9 & 95.3 & 78.8 & 99.5 & 79.5 & \underline{\textbf{100.}} & 91.4 \\ 
              MHPL & & 94.2 & 95.6 & 76.8 & 99.5 & 79.2 & 99.6 & 90.8 \\ 
              \rc
              LAS & & \underline{96.2} & \underline{96.6} & \underline{80.8} & \underline{99.6} & \underline{82.2} & 99.8 & \underline{92.5} \\  
              \midrule
               & DANN & 95.0 & 96.4 & 83.3 & 99.0 & 83.7 & 99.8 & 92.9 \\ 
                & MCC & 94.2 & 95.5 & \textbf{85.1} & \textbf{100.} & \textbf{86.0} & 99.8 & 93.4 \\ 
                \rc
			  &  RAA & 96.9 &	97.6 &	84.2 &	\textbf{100.} &	\textbf{86.0} &	\textbf{100.} &	\textbf{94.1} \\  
               \rc
               \multirow{-4}{*}{LAS}  &  LAA  & \textbf{97.8} &	\textbf{98.5} &	82.8 &	\textbf{100.} &	85.2 &	\textbf{100.} &	94.0 \\ 
			\bottomrule
	\end{tabular} }
	\label{tab:office31}
	\vspace{-4mm}
\end{table}

\subsection{Main Results}
\cparagraph{Comparison with ADA methods on standard datasets.} 
Results on Office-Home using 5\% budget is listed in Tab.~\ref{tab:officehome}. Among the active selection criteria, uncertainty-based criteria (\textit{e.g.}, ENT and MAR) generally obtain higher accuracies than representativeness-based criteria (\textit{e.g.}, CoreSet), indicating that it is inefficient to select target data that are well-aligned with the source domain. When the semi-supervised solver MME~\cite{saito2019semi} is used, the performance gaps among these criteria become smaller. With either strategy, our proposed LAS consistently obtains the best scores, showing that it can select more informative target samples. Table~\ref{tab:officehome_p10} lists the results using 10\%-budget. When using fine-tuning or CDAC~\cite{li2021cross}, LAS obtains better accuracies than other active selection criteria. Compared with the recent LAMDA method, our LAS w/ LAA as a unified solution boosts the accuracy by +2.3\%. LAS w/ RAA is slightly better. 

Table.~\ref{tab:office31} shows results on Office-31 using 5\%-budget. When fixing the adaptation method as fine-tuning, MME or CDAC, LAS consistently performs better than other active selection criteria. When fixing the active selection method as LAS, the proposed RAA/LAA performs better than MME, MCC~\cite{jin2020minimum}, and CDAC. Table~\ref{tab:visda_domainnet} lists results on VisDA using 10\%-budget. LAS w/ LAA outperforms LAMDA by +1.3\%.

\cparagraph{Comparison with ADA methods on label-shifted datasets.} 
Since label distribution mismatch between source and target domains raises a critical issue in ADA, we compare with the LAMDA~\cite{hwang2022combating} devised to address this issue on label-shifted datasets. Following their settings, we use 10\%-budget. Tables~\ref{tab:officehome_rsut},\ref{tab:visda_domainnet} list the comparison results. LAMDA applies importance sampling on source data to match the label distribution of source and target domains. This is particularly useful to domain adversarial methods like DANN. Our RAA/LAA belong to self-training methods. We use Class Balancing Rejection to create class-balanced training data. Given LAS, RAA/LAA achieve higher accuracies than other adaptation methods. The full LAS w/ LAA method improves over LAMDA by +1\%.

\begin{table}[!t]
	\caption{Accuracies (\%) on \textbf{Office-Home RSUT} using 10\%-budget. ($^\dagger$Using importance sampling)} 
	\centering
	\footnotesize
\vspace{-2mm}
	\scalebox{0.85}{
		\begin{tabular}{p{1.4cm}p{1.4cm}p{0.4cm}p{0.4cm}p{0.4cm}p{0.4cm}p{0.4cm}p{0.4cm}p{0.4cm}<{\centering} }
			\toprule
			AL & DA & C$\shortrightarrow$P & C$\shortrightarrow$R & P$\shortrightarrow$C & P$\shortrightarrow$R & R$\shortrightarrow$C & R$\shortrightarrow$P & Avg.\\
			\midrule		
			S$^3$VAADA	& VAADA & 73.0 & 63.0 & 50.7 & 69.6 & 52.6 & 78.3 & 64.5 \\
			TQS & DANN$^\dagger$ & 67.6 & 61.4 & 54.8 & 74.7 & 53.6 & 77.6 & 64.9 \\
			LAMDA & DANN$^\dagger$ & 81.2 & 75.7 & \textbf{64.1} & 81.6 & 65.1 & 87.2 & 75.8 \\
                \midrule
                \multirow{6}{*}{LAS}  & DANN & 80.2 & 69.4 & 58.4 & 76.9 & 60.8 & 85.3 & 71.8 \\ 
			 & MME & 75.6 &	68.9 &	56.6 &	76.9 &	57.9 &	87.2 &	70.5 \\	
              & MCC & 82.3 & 72.0 & 62.6 & 81.2 & \textbf{65.9} & 88.3 & 75.4 \\ 
              & CDAC & 79.6 & 71.9 & 62.1 & 81.4 & 63.7 & 85.6 & 74.1 \\ 
			\rc
			 & RAA & \textbf{83.8} &	73.6 &	64.0 &	82.6 &	65.2 &	\textbf{88.6} &	76.3 \\	
			\rc
			 & LAA & 83.2 &	\textbf{77.2} &	63.8 &	\textbf{83.0} &	65.4 &	88.1 &	\textbf{76.8} \\		
			\bottomrule
			\vspace{5mm}
		\end{tabular}
	}
	\label{tab:officehome_rsut}
	\end{table}
 
\begin{table}[!t]
		\caption{Accuracies (\%) on \textbf{VisDA} and \textbf{DomainNet} using 10\%-budget. ($^\dagger$Using importance sampling)} 
 \vspace{-1mm}
	\centering
	\footnotesize
\vspace{-1mm}
	\scalebox{0.85}{
		\begin{tabular}{p{1.4cm}p{1.4cm}p{0.6cm}<{\centering}p{0.4cm}p{0.4cm}p{0.4cm}p{0.4cm}p{0.4cm}<{\centering} }
			\toprule
			\multirow{2}{*}{AL} & \multirow{2}{*}{DA} &  \multirow{2}{*}{VisDA} & \multicolumn{5}{c}{DomainNet} \\			
			& & & R$\shortrightarrow$C & C$\shortrightarrow$S & S$\shortrightarrow$P & C$\shortrightarrow$Q & Avg. \\
			\midrule
			TQS & DANN$^\dagger$ & 87.7 & 59.3 & 50.9 & 52.4 & 41.5 & 51.0 \\	
			LAMDA  & DANN$^\dagger$ & 91.8 & 65.3 & 56.1 & 58.1 & 48.3 & 57.0 \\
                \midrule
                 & DANN & 91.3 & 62.1 & 53.1 & 53.1 & 42.4 & 52.7 \\ 
			   & MME & 92.2 & 65.9 &	54.1 &	55.9 &	42.9 &	54.7 \\
			\rc
			   & RAA & 93.0 & \textbf{70.3} &	\textbf{58.1} &	\textbf{60.3} &	48.9 &	 \textbf{59.4} \\
			\rc        
		  \multirow{-4}{*}{LAS} & LAA & \textbf{93.1} & 69.4	&   57.5 &	59.9 &	\textbf{49.1} &	59.0 \\
			
			\bottomrule
	\end{tabular} }
	\label{tab:visda_domainnet}
	\vspace{-4mm}
\end{table}

\begin{figure*}[!t]
	\centering	
	\begin{subfigure}[]{0.175\textwidth}
		\includegraphics[width=\textwidth]{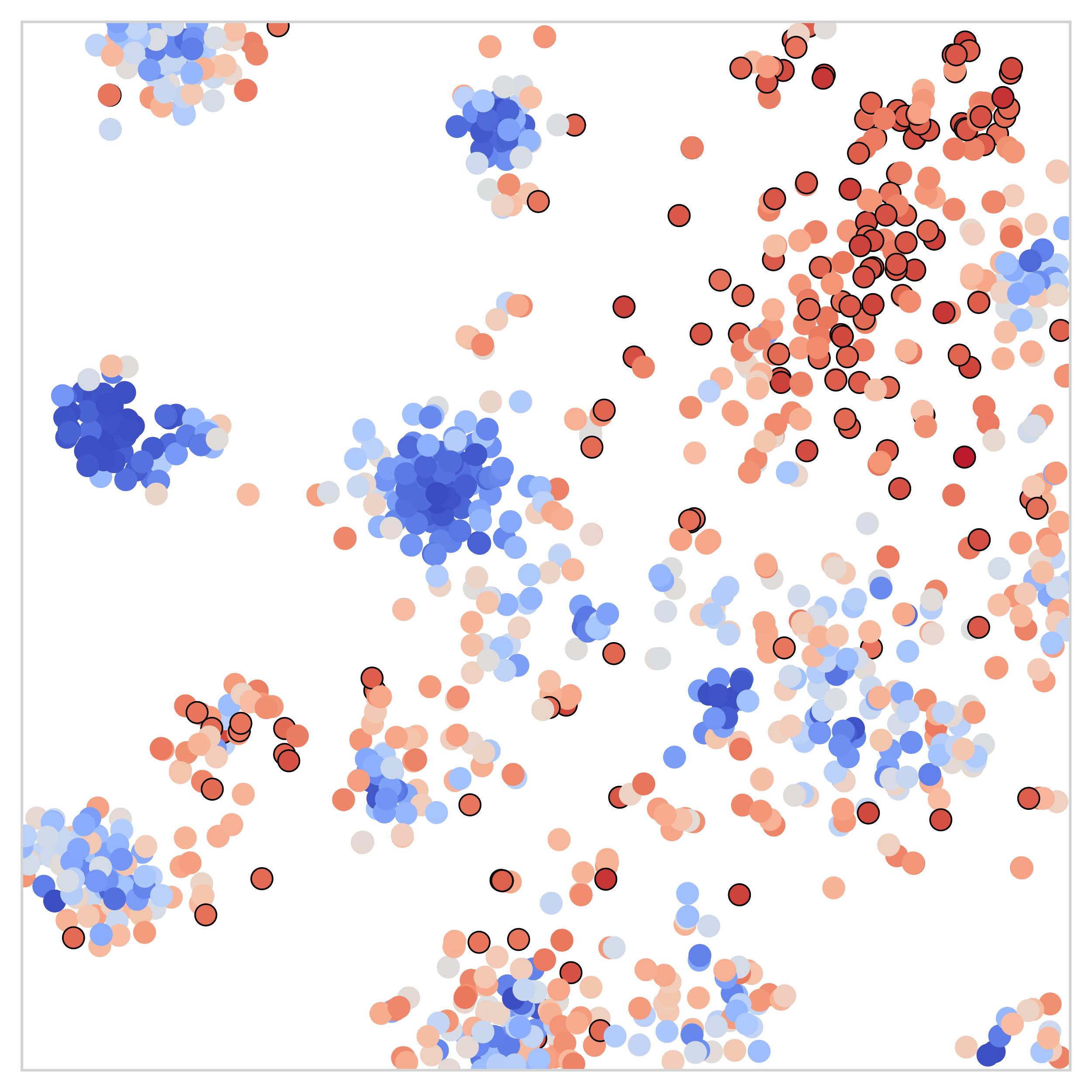}
		\caption{entropy}
		\label{fig:tsne:entropy}
	\end{subfigure}	
	\begin{subfigure}[]{0.175\textwidth}
		\includegraphics[width=\textwidth]{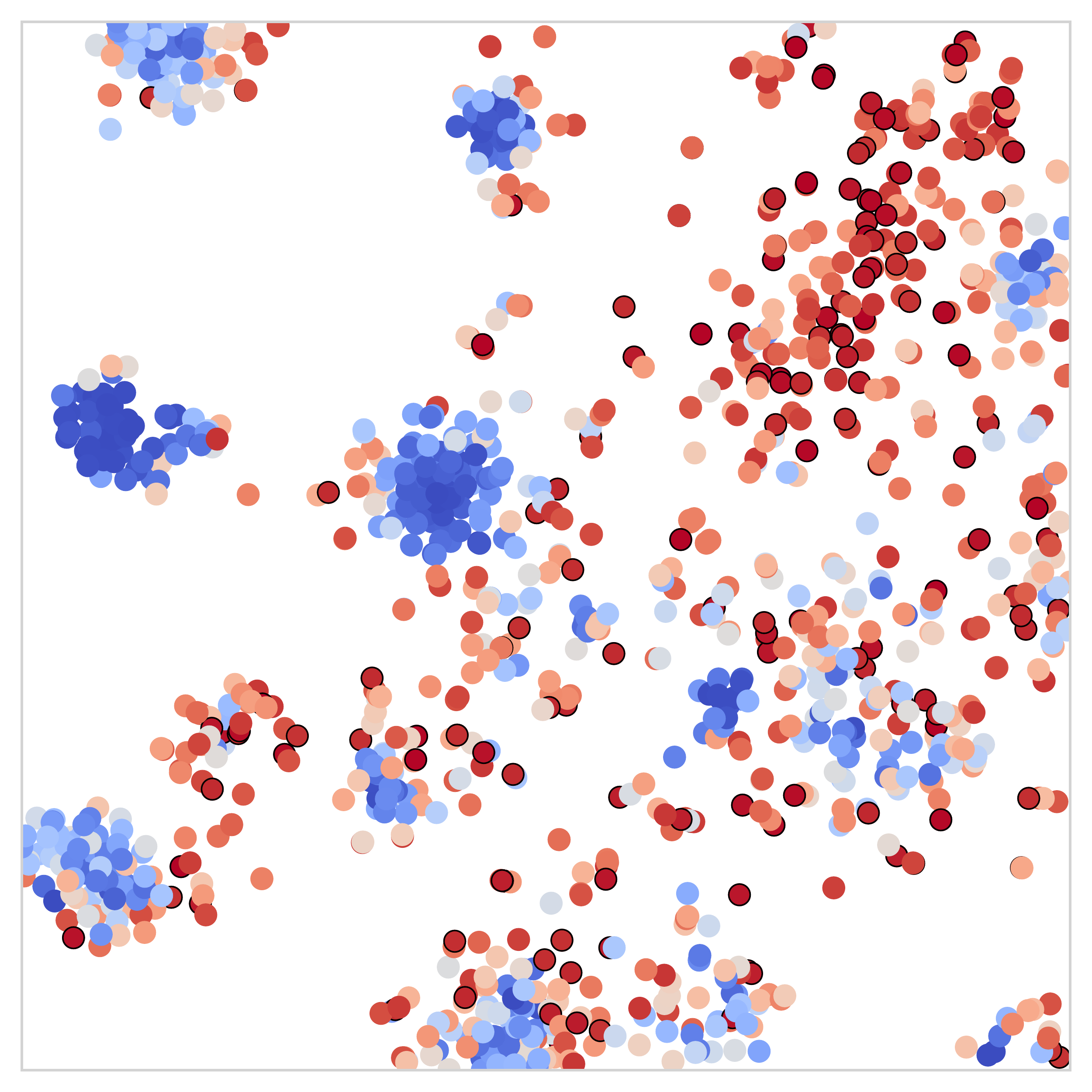}
		\caption{margin}
		\label{fig:tsne:margin}
	\end{subfigure}	
	\begin{subfigure}[]{0.175\textwidth}
		\includegraphics[width=\textwidth]{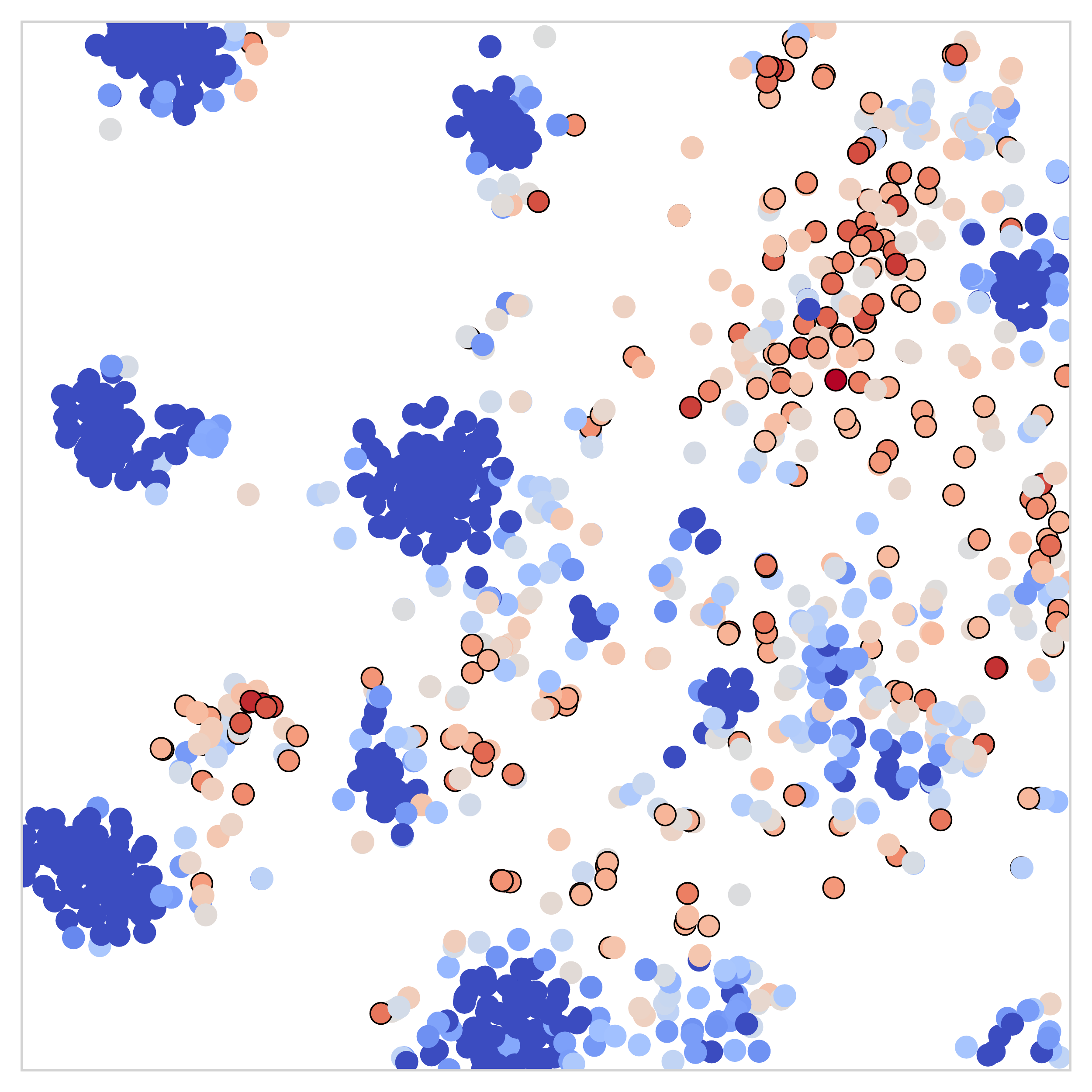}
		\caption{NAU}
		\label{fig:tsne:NAU}
	\end{subfigure}	
	\begin{subfigure}[]{0.175\textwidth}
		\includegraphics[width=\textwidth]{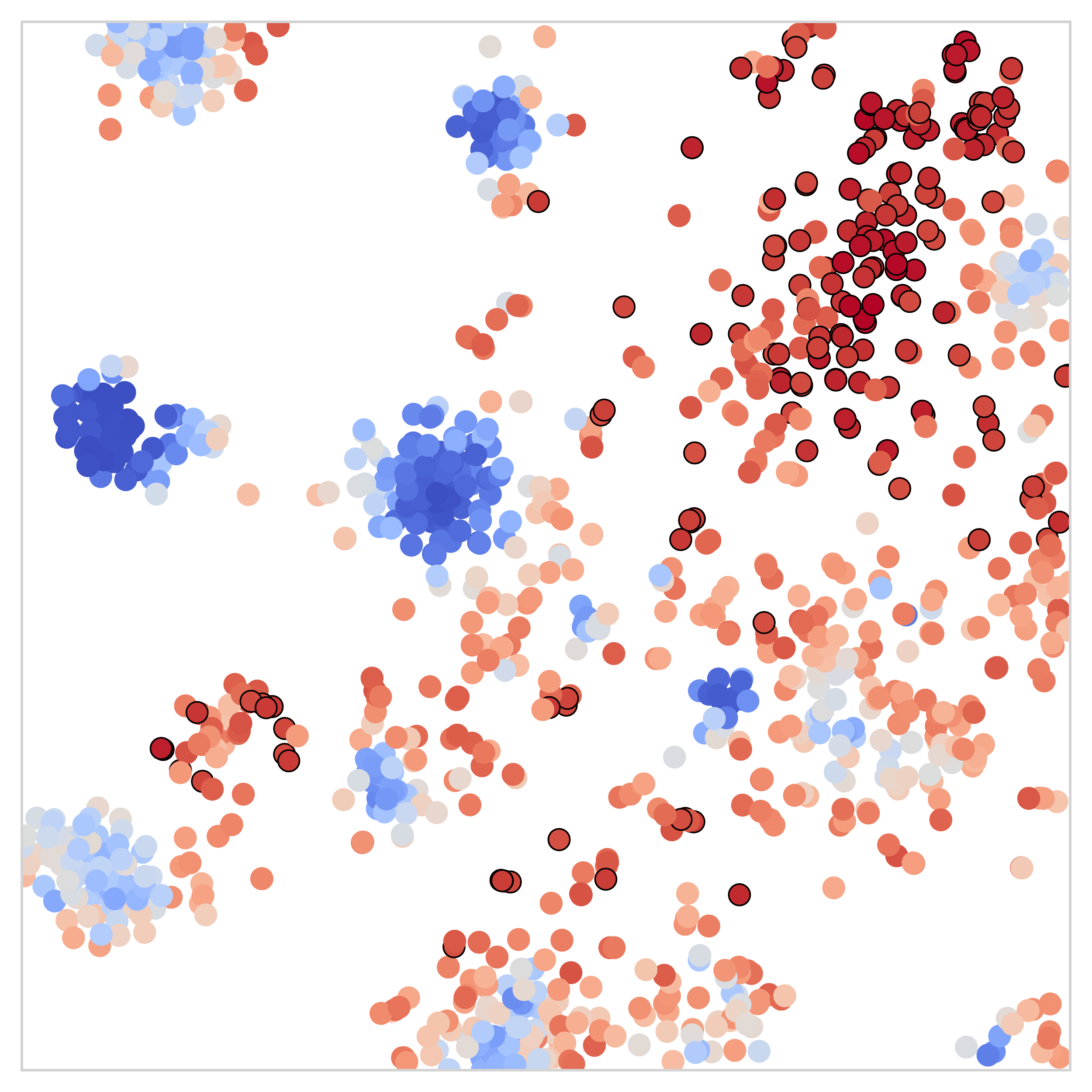}
		\caption{LI}
		\label{fig:tsne:LI}
	\end{subfigure}	
	\begin{subfigure}[]{0.028\textwidth}
		\includegraphics[width=\textwidth]{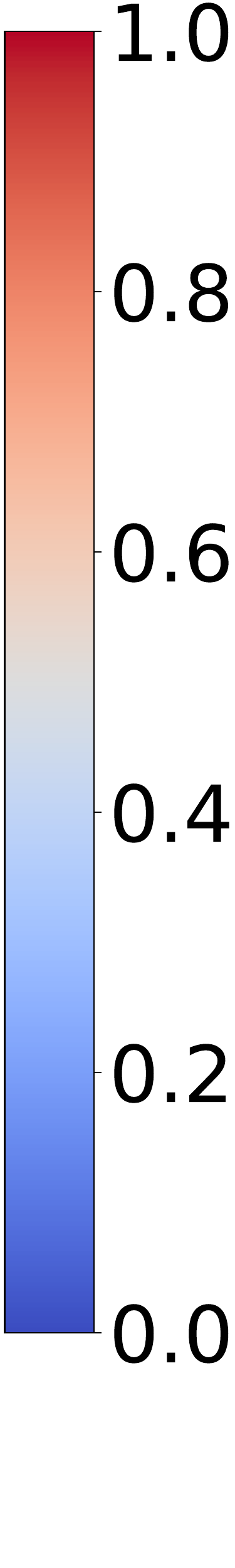}
	\end{subfigure}	
	\begin{subfigure}[]{0.24\textwidth}
		\includegraphics[width=\textwidth]{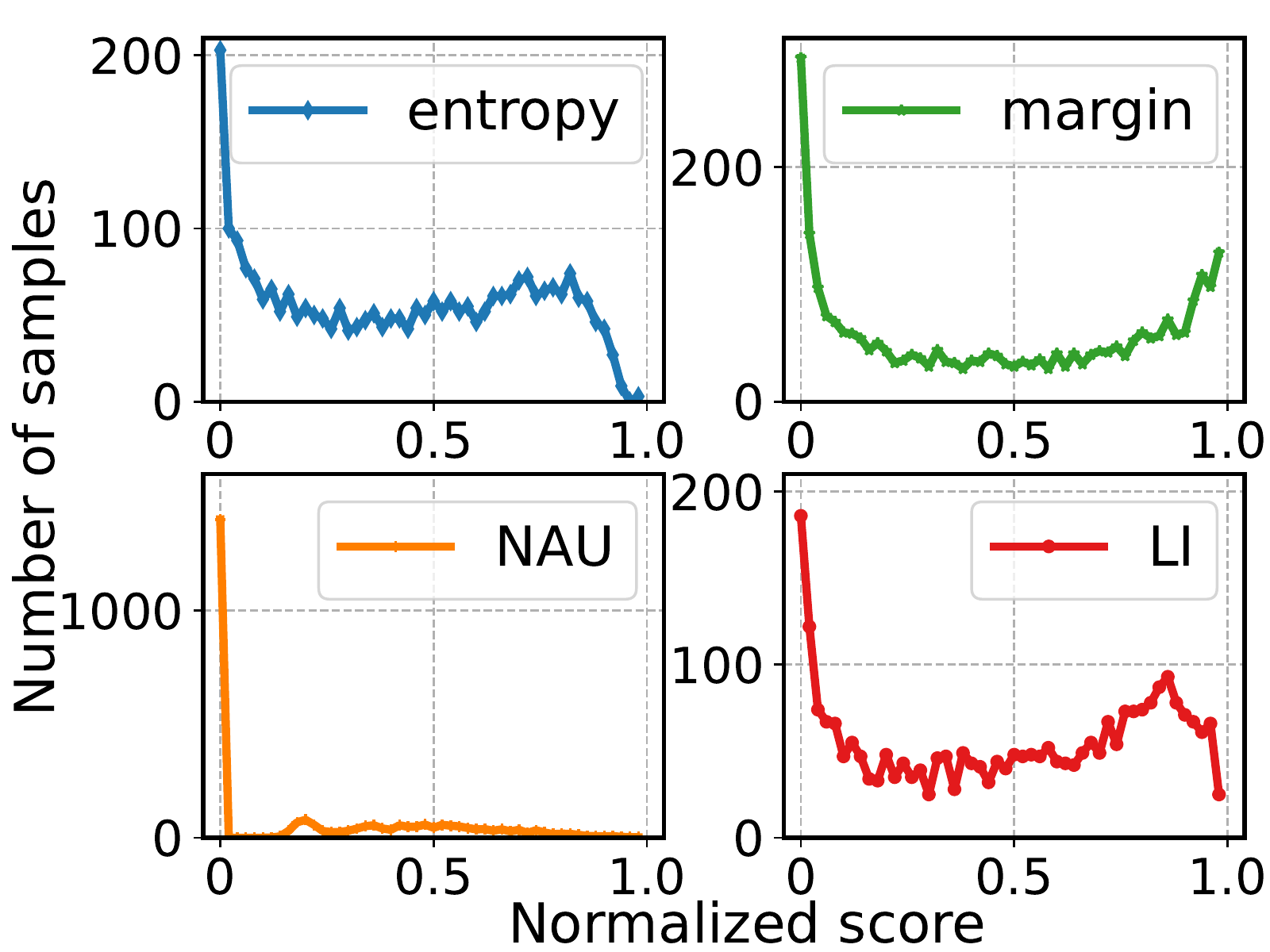}
		\caption{histogram}
		\label{fig:tsne:histogram}
	\end{subfigure}	
	\vspace{-3mm}\caption{(a-d) $t$-SNE visualization of target features on Office-31 W$\rightarrow$A. Samples are colored according to their normalized uncertainty scores, where red indicates large values and blue indicates small values. The top 10\% samples with highest scores are marked with black boarders. (e) Histogram of target samples by normalized scores. }
	\label{fig:tsne}
	\vspace{-3mm}
\end{figure*}

\begin{figure}[!t]
	
	\begin{center}
	\includegraphics[trim=2in 0.25in 0.25in  2in,clip,width=.355\linewidth]{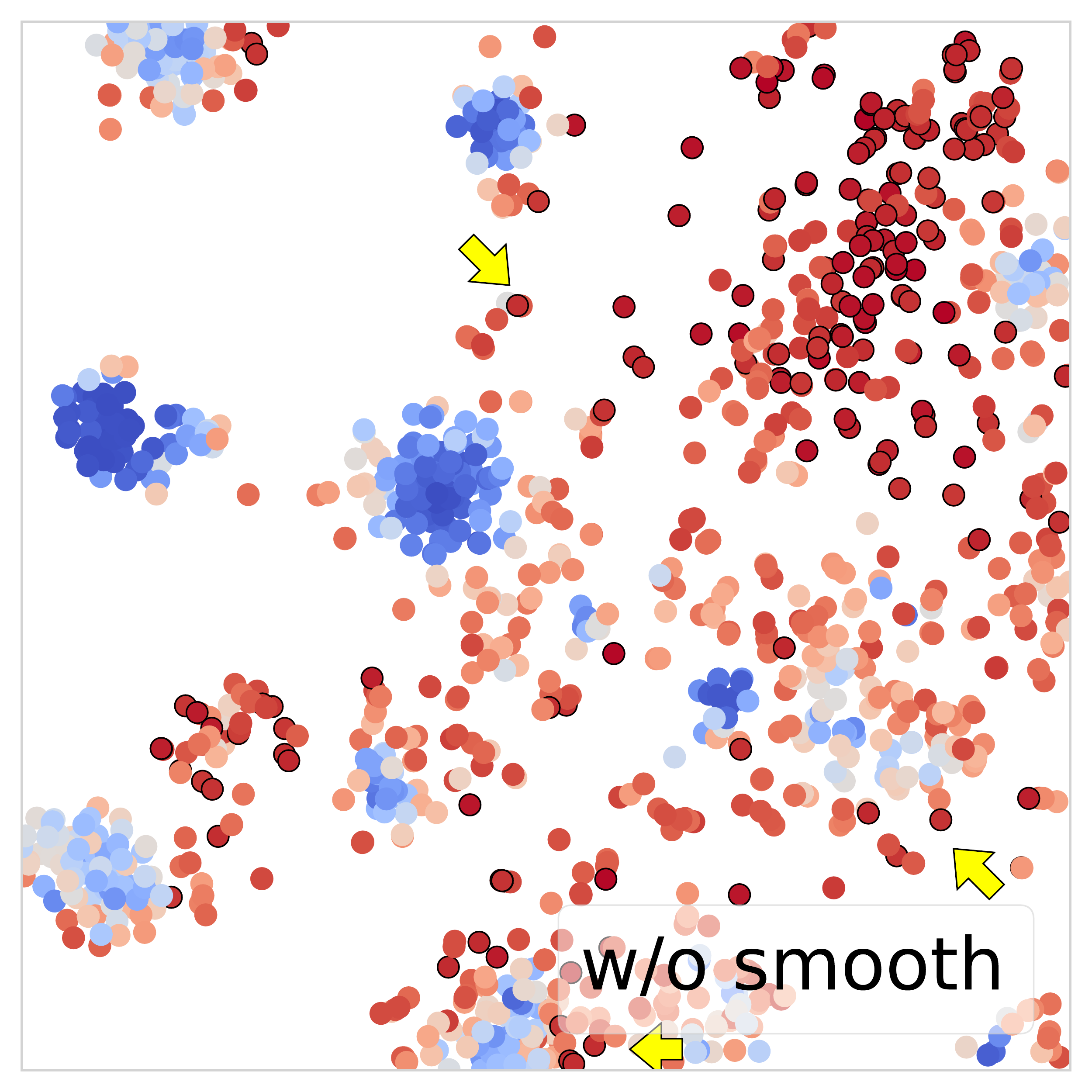}
        \includegraphics[trim=2in 0.25in 0.25in  2in,clip,width=.355\linewidth]{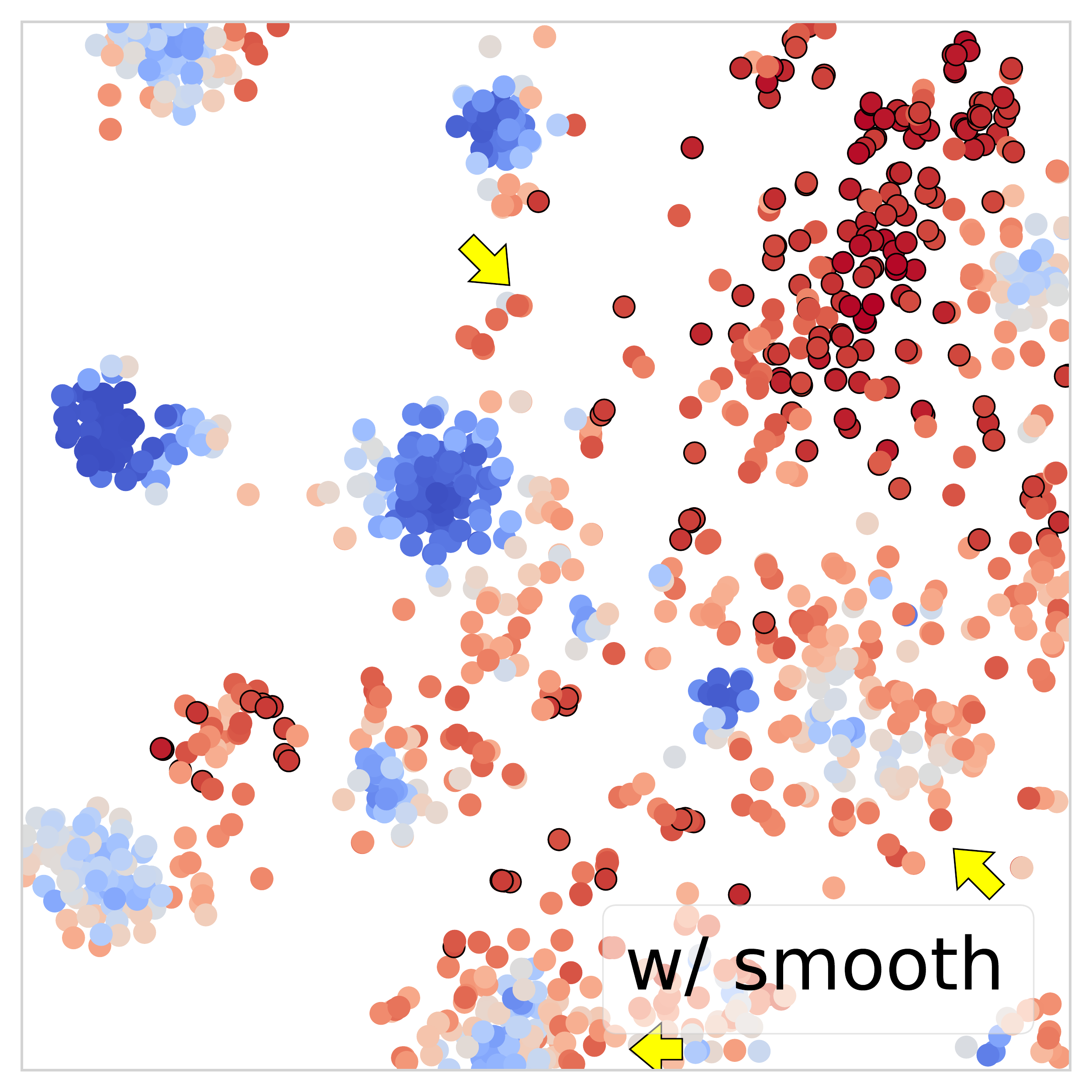}
        \hfill
        \includegraphics[width=.27\linewidth]{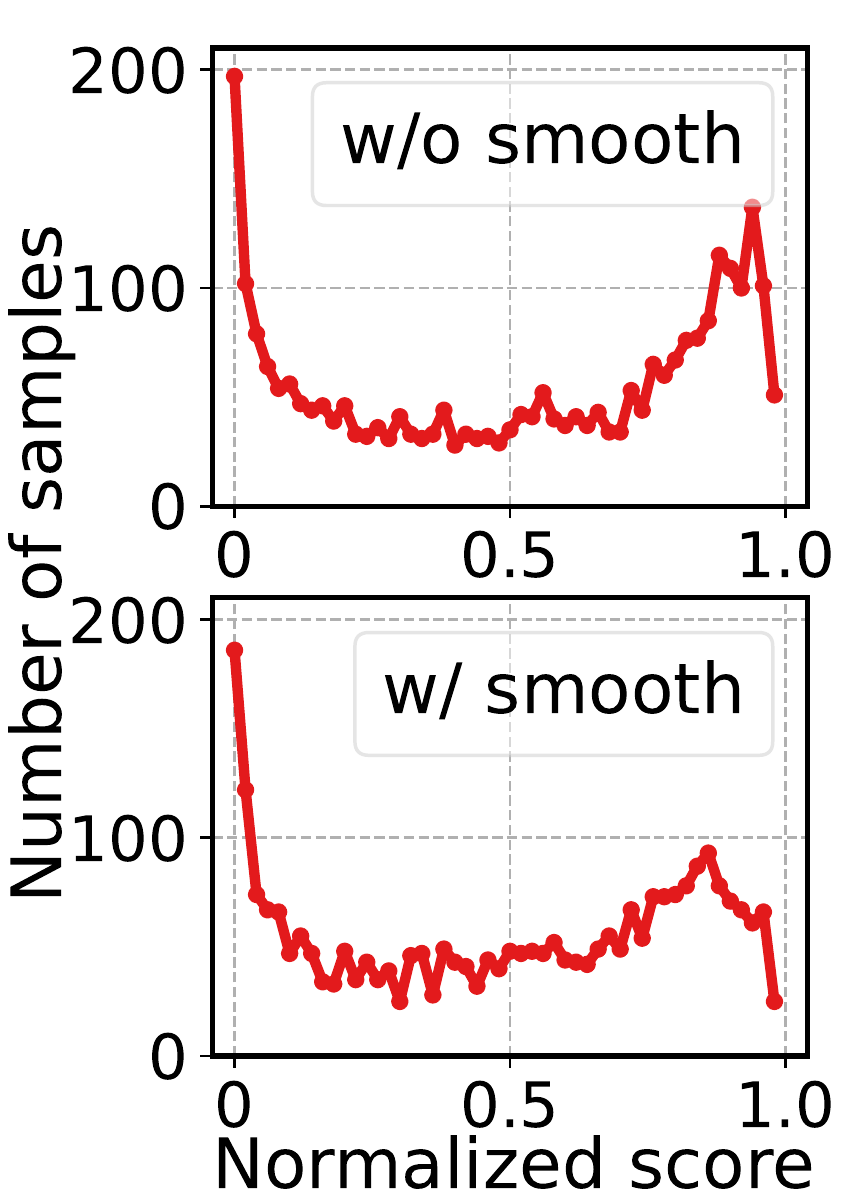}
	\end{center}
      \vspace{-4mm}
        \caption{Zoomed in $t$-SNE visualization of target features and histograms on Office-31 W$\rightarrow$A for LI. The smoothing in Eq.~\ref{eq:LI_sm} removes some outliers (\eg, samples annotated with yellow arrows).}
	\label{fig:LI_smooth}
  \vspace{-3mm}
\end{figure}

\subsection{Analysis on LAS}

\begin{figure}[!t]
	\centering		
	\begin{subfigure}[]{0.45\linewidth}
		\includegraphics[width=\linewidth]{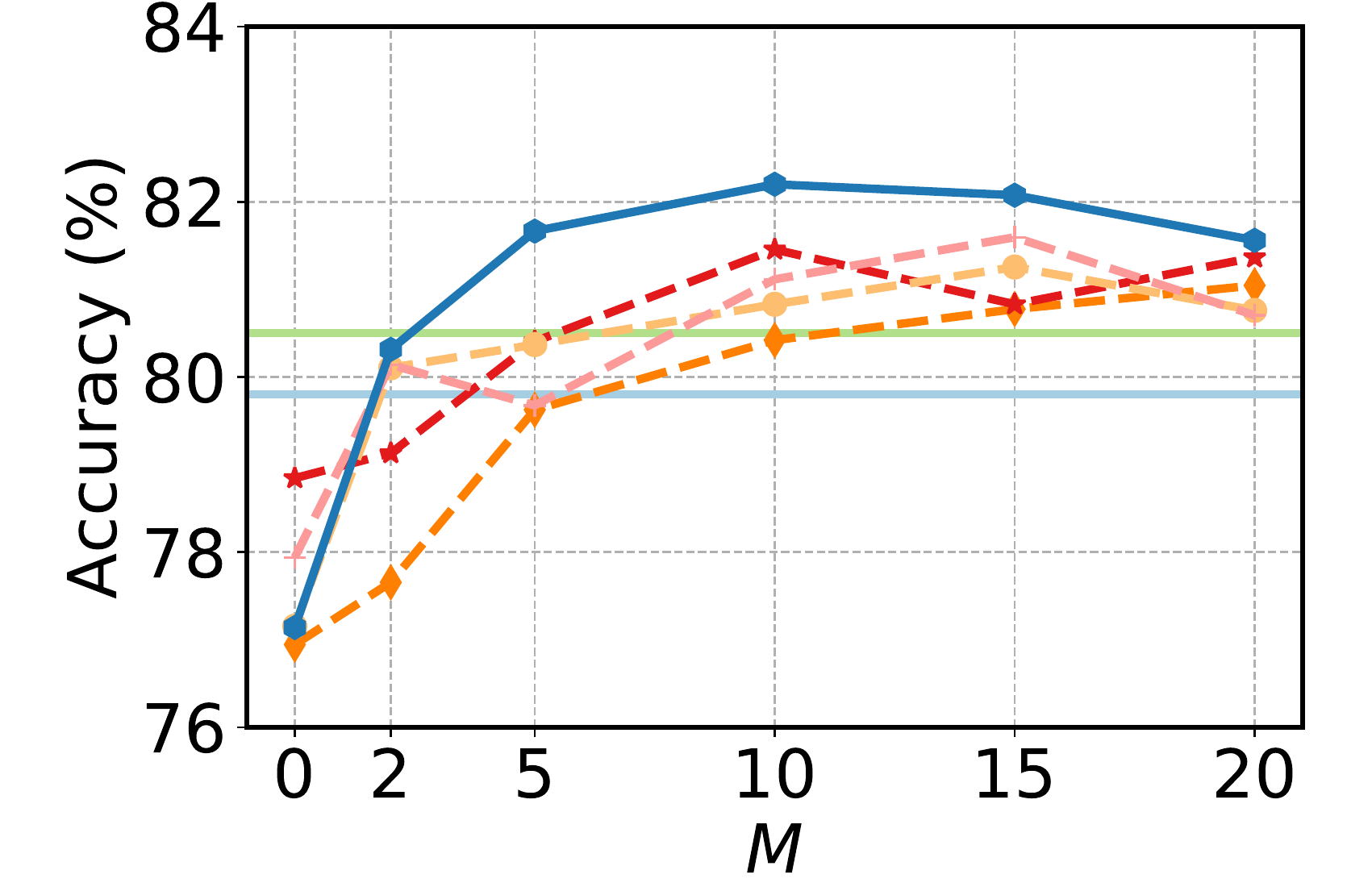}\vspace{-1mm}\caption{Office-31 W$\rightarrow$A}	
		\label{fig:uncertainty:w2a}
	\end{subfigure}	
	\begin{subfigure}[]{0.54\linewidth}
		\includegraphics[width=\linewidth]{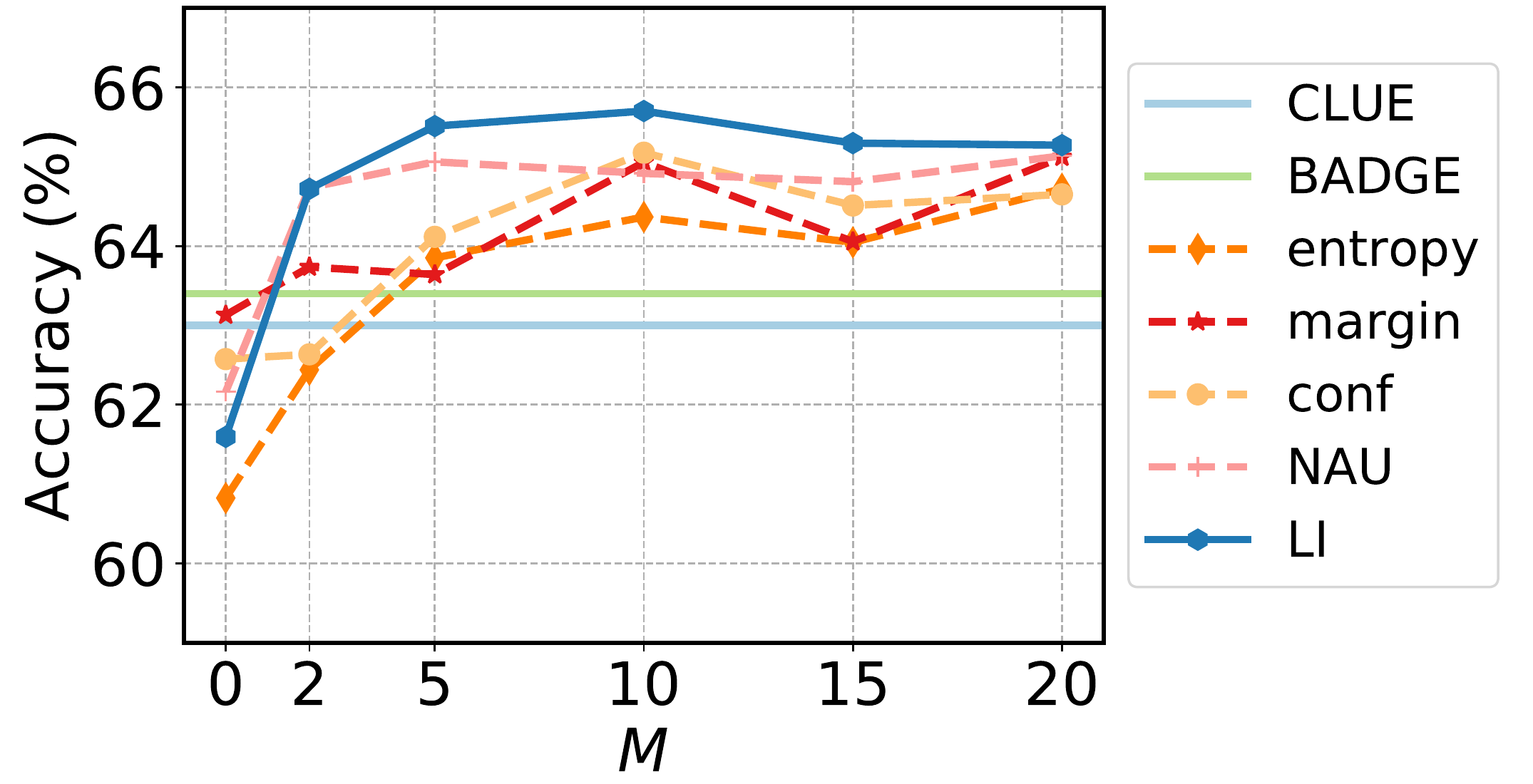}\vspace{-1mm}\caption{Office-Home Ar$\rightarrow$Cl}	
		\label{fig:uncertainty:ar2cl}
	\end{subfigure}	
	\begin{subfigure}[]{0.4\linewidth}	
		\centering
		\footnotesize
		\scalebox{0.8}{		
			\begin{tabular}{p{1.2cm}<{\centering}|p{0.5cm}<{\centering}p{0.5cm}<{\centering}}
				\toprule
				Criterion & OH & O31\\
				\midrule
				CLUE & 72.8 & 90.1 \\
				BADGE & 73.5 & 89.7 \\
				\midrule
				entropy & 74.3 & 90.2\\			
				margin & 74.2 & 90.5\\				
				conf & 74.2 & 90.3\\
				NAU & 74.4 & 90.5 \\
				\rc
				LI & \textbf{75.3} & \textbf{91.3}\\				
				\bottomrule
		\end{tabular} }
		\caption{}
		\label{fig:uncertainty:M10}		
	\end{subfigure}	
	\begin{subfigure}[]{0.52\linewidth}
		\includegraphics[width=\linewidth]{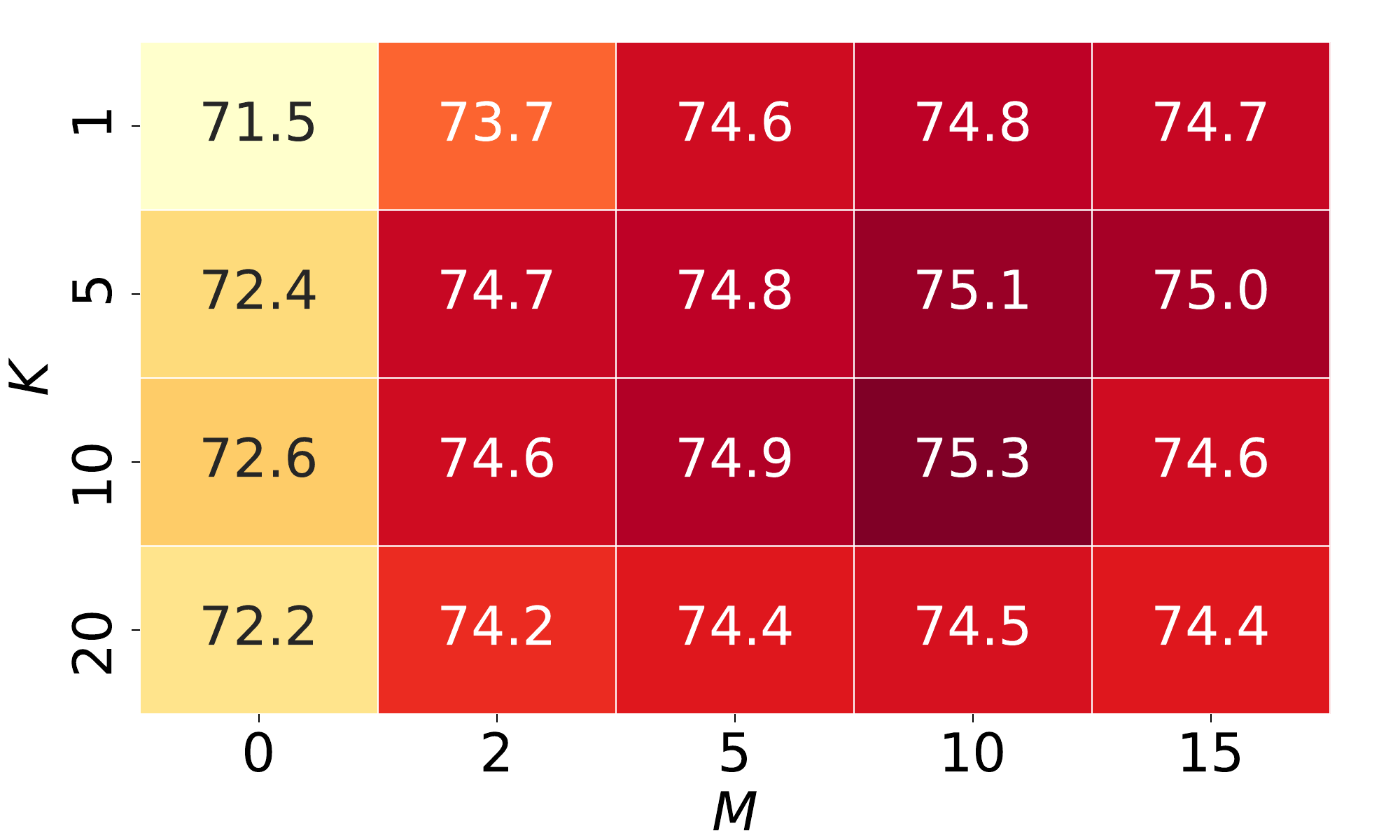}        
		\caption{}
		\label{fig:uncertainty:MK}		
	\end{subfigure}
    \vspace{-2mm}
 \caption{(a,b) Replacing LI score in LAS with different uncertainty criteria; (c) comparison results when $M=10$; (d) varying $K$ and $M$ in LAS on Office-Home.}
	\label{fig:uncertainty}	
	\vspace{-4mm}
\end{figure}

\cparagraph{Uncertainty measures in LAS.} LAS discovers uncertain regions with the LI score. To analyze its properties, we replace the LI score in LAS with other uncertainty measures while keeping the same diverse sampling procedure. Comparison measures include \textit{prediction margin}, \textit{entropy}, \textit{prediction confidence} and \textit{NAU score}\cite{wang2023mhpl}. Fig.~\ref{fig:uncertainty:w2a} and Fig.~\ref{fig:uncertainty:ar2cl} plots the performances under different oversampling rate $M$. From the figure, using a large $M$ is beneficial to all uncertainty measures, indicating the necessity to balance between uncertainty and diversity. CLUE and BADGE are also hybrid methods, but use a different way to LAS. CLUE runs a clustering on all target samples based on entropy weighted distances, and BADGE runs a clustering on all target samples based on the gradient embeddings. Both of them rely on the entire target data. In contrast, LAS focuses on some local uncertain regions, and involves only about 12\% target data in our experiments. When $M\geq 10$, LAS and its variants surpass CLUE and BADGE on the two tasks. In Fig.~\ref{fig:uncertainty:M10} when $M=10$, LI leads to the best scores on Office-Home and Office-31. Fig.~\ref{fig:uncertainty:MK} studies the parameter sensitivity on $K$ and $M$ in LAS. Performances are close in the proximal of optimal values. 

\begin{figure*}[!t]
	\centering	
	\includegraphics[width=0.36\textwidth]{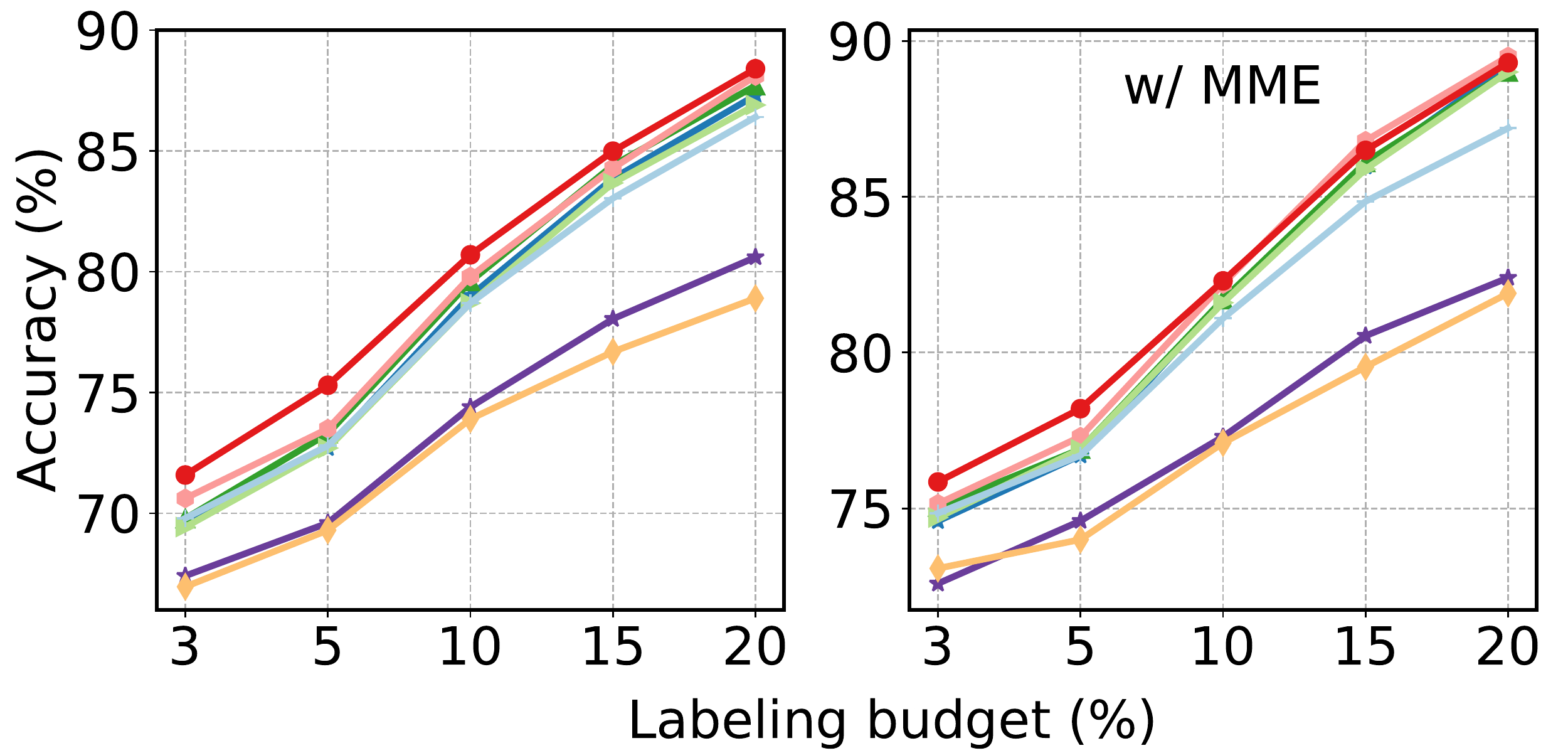}	
	\hfil
	\includegraphics[width=0.28\textwidth]{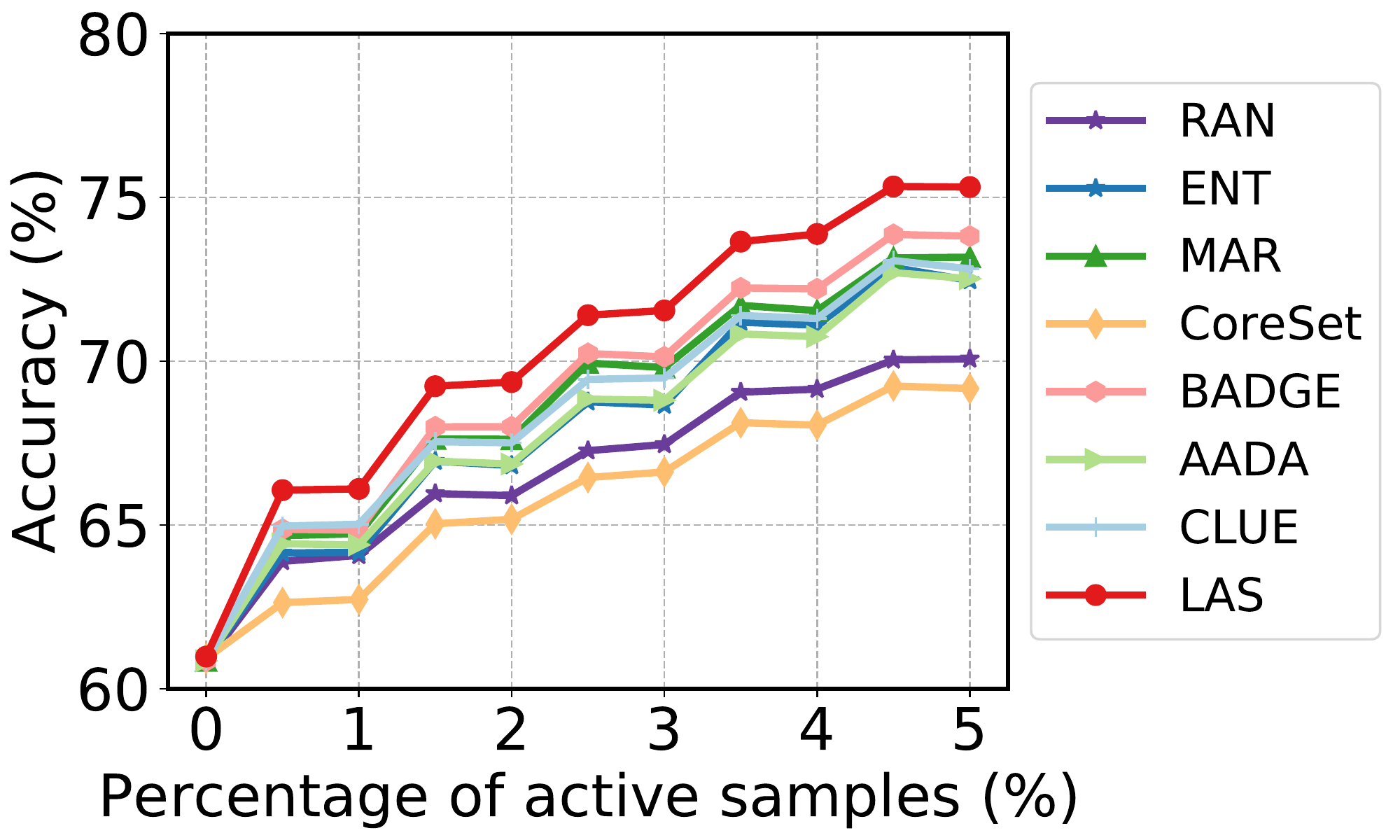}	
	\hfil
	\includegraphics[width=0.3\textwidth]{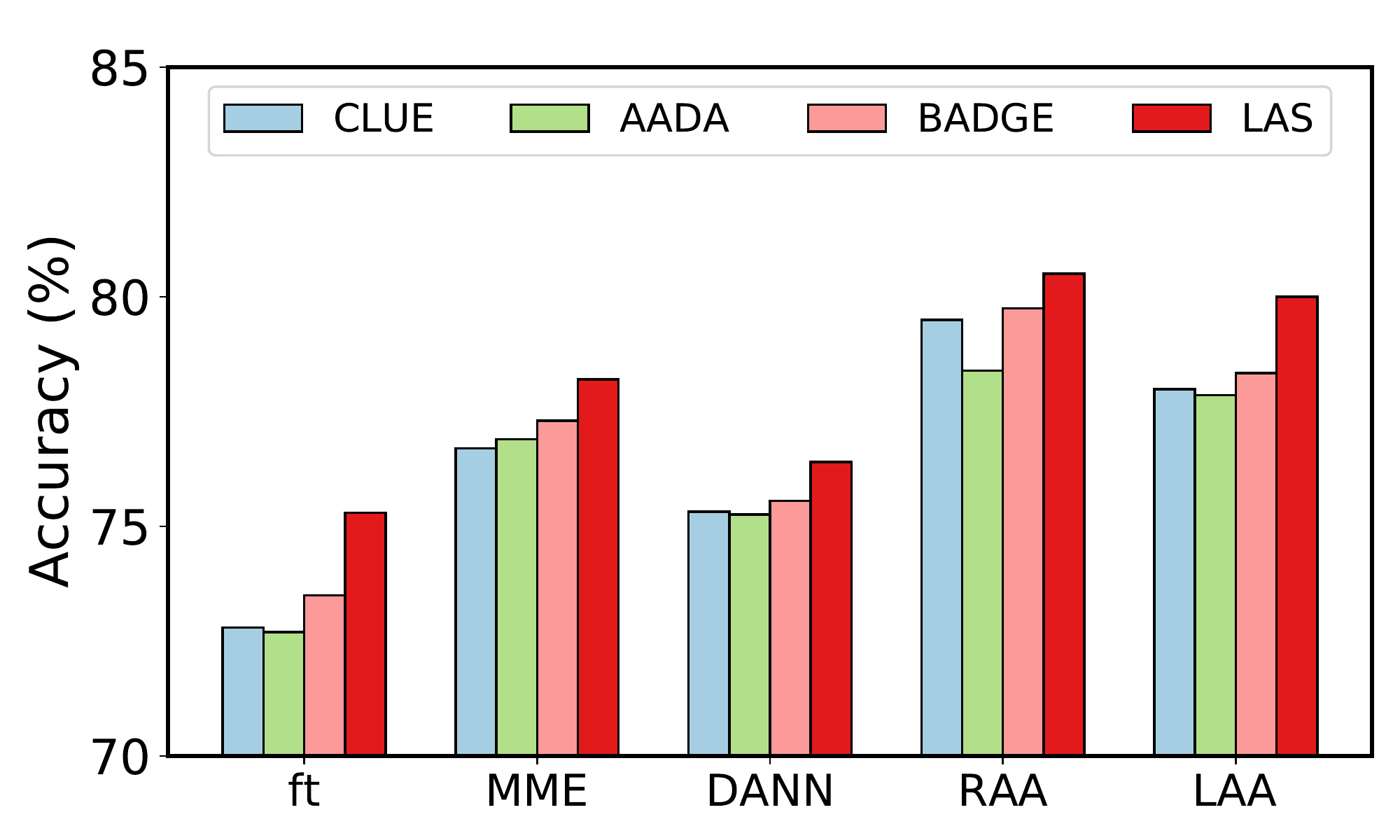}	
	\vspace{-2.5mm}\caption{Analysis on Office-Home. (Left) varying labeling budget; (Center) accuracy curves using 5\%-budget; (Right) combining AL criteria with different DA strategies using 5\%-budget.}
	\label{fig:OH_budgets_DA}
     \vspace{-4mm}
\end{figure*}

\cparagraph{Advantages of LAS over other criteria.} To better show the advantage of LAS over other active learning criteria, Fig.~\ref{fig:OH_budgets_DA} plots the accuracies under different labeling budgets and domain adaptation methods. In the left figure, LAS consistently achieves better or comparable accuracies than other criteria with both fine-tuning and MME. The improvement is more significant when the labeling budget is small. The center figure plots the accuracy curves using 5\%-budget. LAS outperforms other criteria through training process. In the right figure, no matter which DA strategy is used, LAS obtains the best scores.   

To further analyze the difference between these uncertainty measures, Fig.~\ref{fig:tsne} visualizes the target features on Office-31 W$\rightarrow$A. We normalize all uncertainty scores to $[0,1]$ and color each point accordingly. Meanwhile, the top 10\% samples with highest scores are marked with black boarders. Entropy and margin do not consider the local-context. Thus they tend to include some outliers as shown in Fig.~\ref{fig:tsne:entropy} and Fig.~\ref{fig:tsne:margin}. NAU is defined as $\mathrm{NAU}=\mathrm{NP}\times \mathrm{NA}$, where $\mathrm{NP}$ is the entropy of class distribution among neighboring samples and $\mathrm{NA}$ is the average similarity between a sample and its neighbors. Since the number of neighbors is limited, $\mathrm{NP}$ has discrete values and tend to be small. Shown in Fig.~\ref{fig:tsne:NAU} and Fig.~\ref{fig:tsne:histogram}, the majority target data have small NAU scores. In contrast, uncertainty samples tend to have large LI scores and are more gathered. A peak around 0.8 can be observed in the histogram of LI. This also explains why a diverse sampling is important for LI. In the zoomed visualization of Fig.~\ref{fig:LI_smooth}, we can see that the smoothing in LI helps as it can remove some outliers.  

\subsection{Analysis on RAA/LAA}

\begin{figure}[!t]
	\centering	
	\begin{subfigure}[]{0.22\textwidth}
		\includegraphics[width=\textwidth]{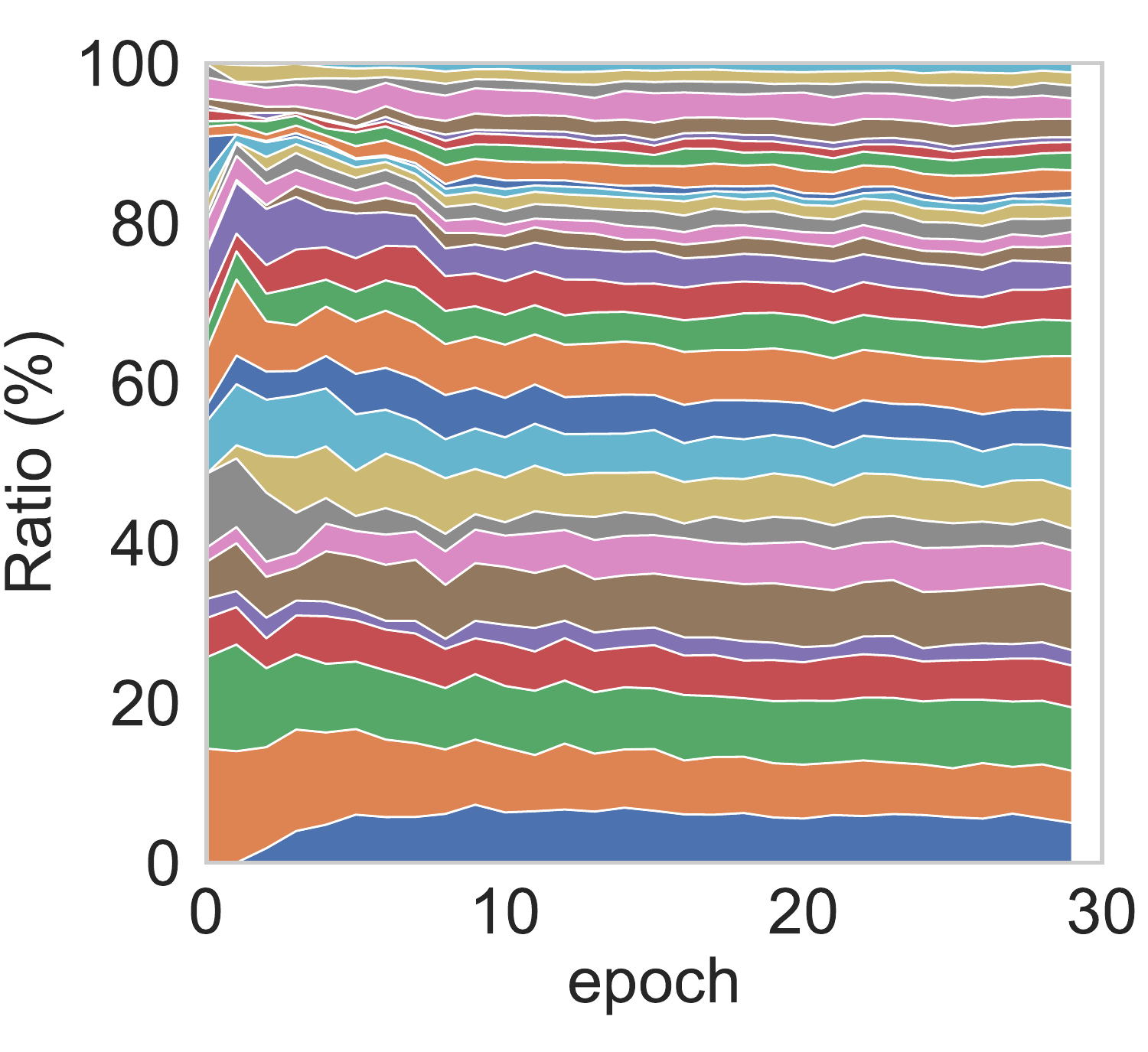}
		\caption{w/o CBR}	
	\end{subfigure}
	\begin{subfigure}[]{0.22\textwidth}
		\includegraphics[width=\textwidth]{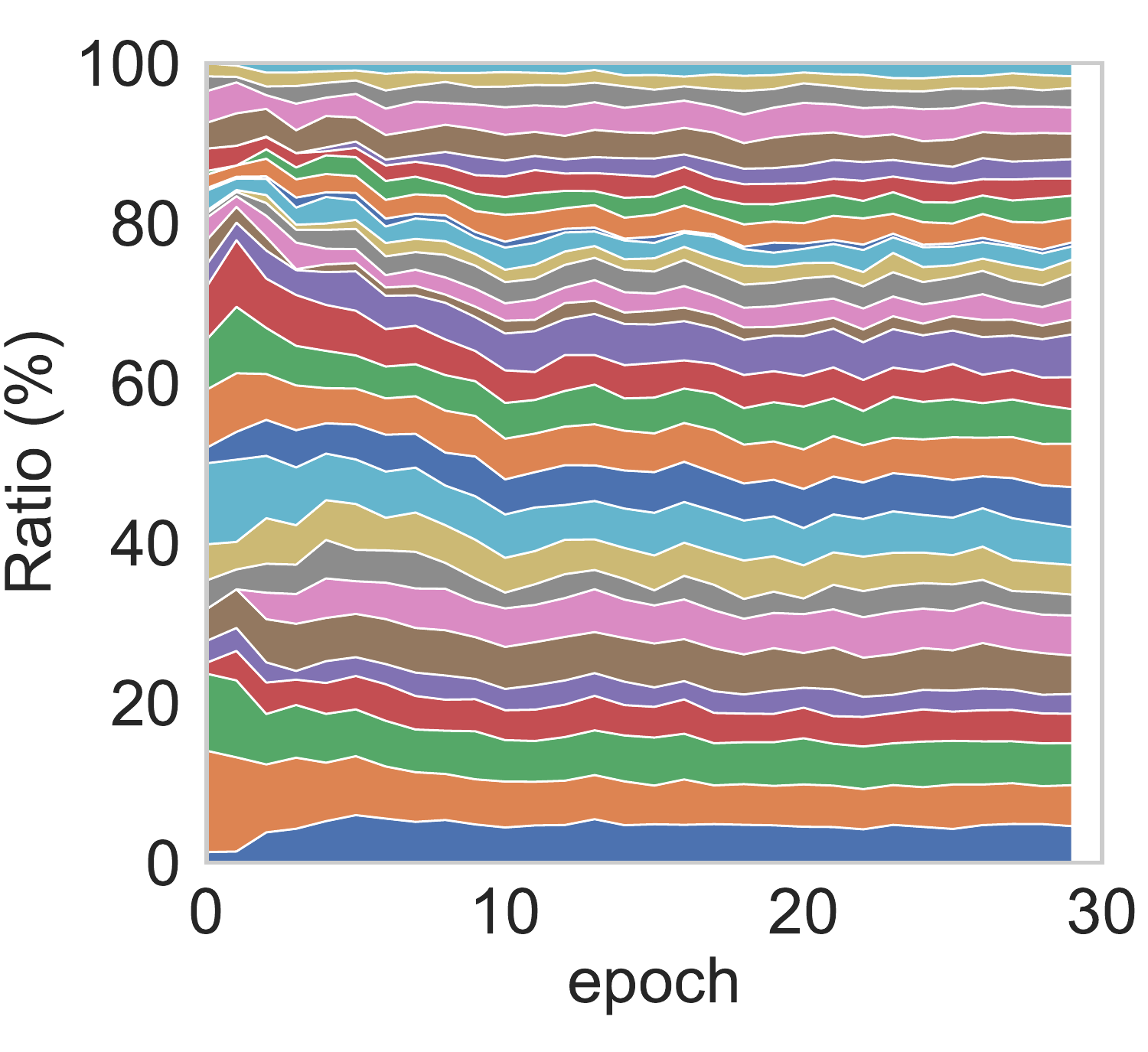}
		\caption{w/ CBR}	
	\end{subfigure}
	\vspace{-2mm}\caption{Effects of Class Balancing Rejection (CBR) in LAA on the ratio of samples per class among $\mathcal{A}$. (Averaged over six tasks on Office-Home RSUT. For better visualization, only 15 major and 15 minor classes are included.)}
	\vspace{-4mm}
	\label{fig:CB_dist}
\end{figure}

\cparagraph{Effects of Class Balancing Rejection (CBR).} To handle issues from label distribution shift, an effective way is to create a class balanced training set. We realize this through using different rejection probabilities for target samples from major or minor classes when creating the anchor set $\mathcal{A}$. Figure~\ref{fig:CB_dist} visualizes the ratio of samples per class among $\mathcal{A}$. As can be seen, $\mathcal{A}$ is dominated by major classes without using CBR. In contrast, the ratio of samples from minor classes increases when CBR is used. This helps to train each class in a more balanced manner, as evident from +1.0\% increase in per-class average accuracy in Tab.~\ref{tab:abl}

\begin{figure}[!t]
	\centering
	\begin{subfigure}[]{0.42\linewidth}
		\includegraphics[width=\linewidth]{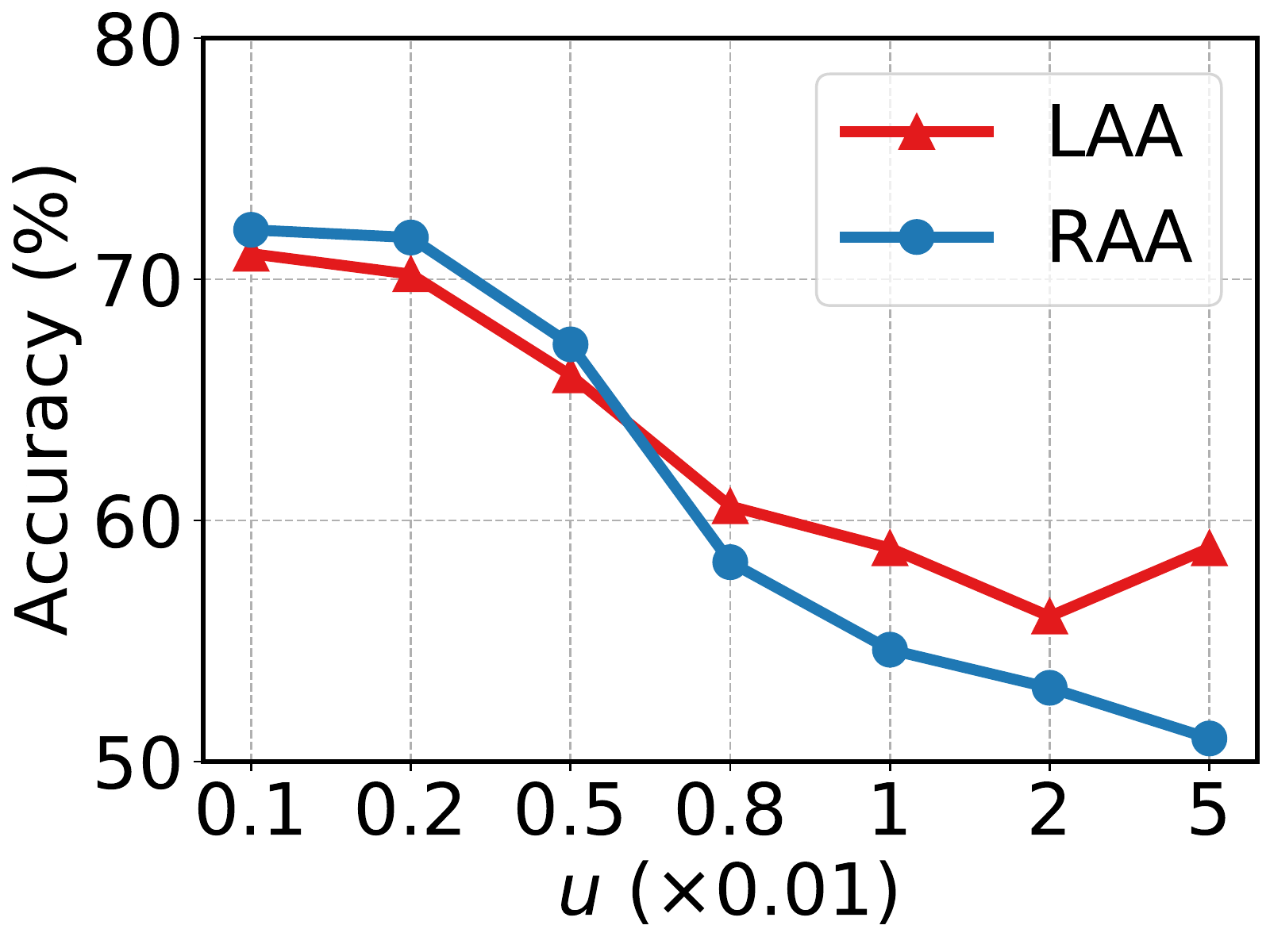}
	\end{subfigure}	
	\begin{subfigure}[]{0.56\linewidth}
		\includegraphics[width=\linewidth]{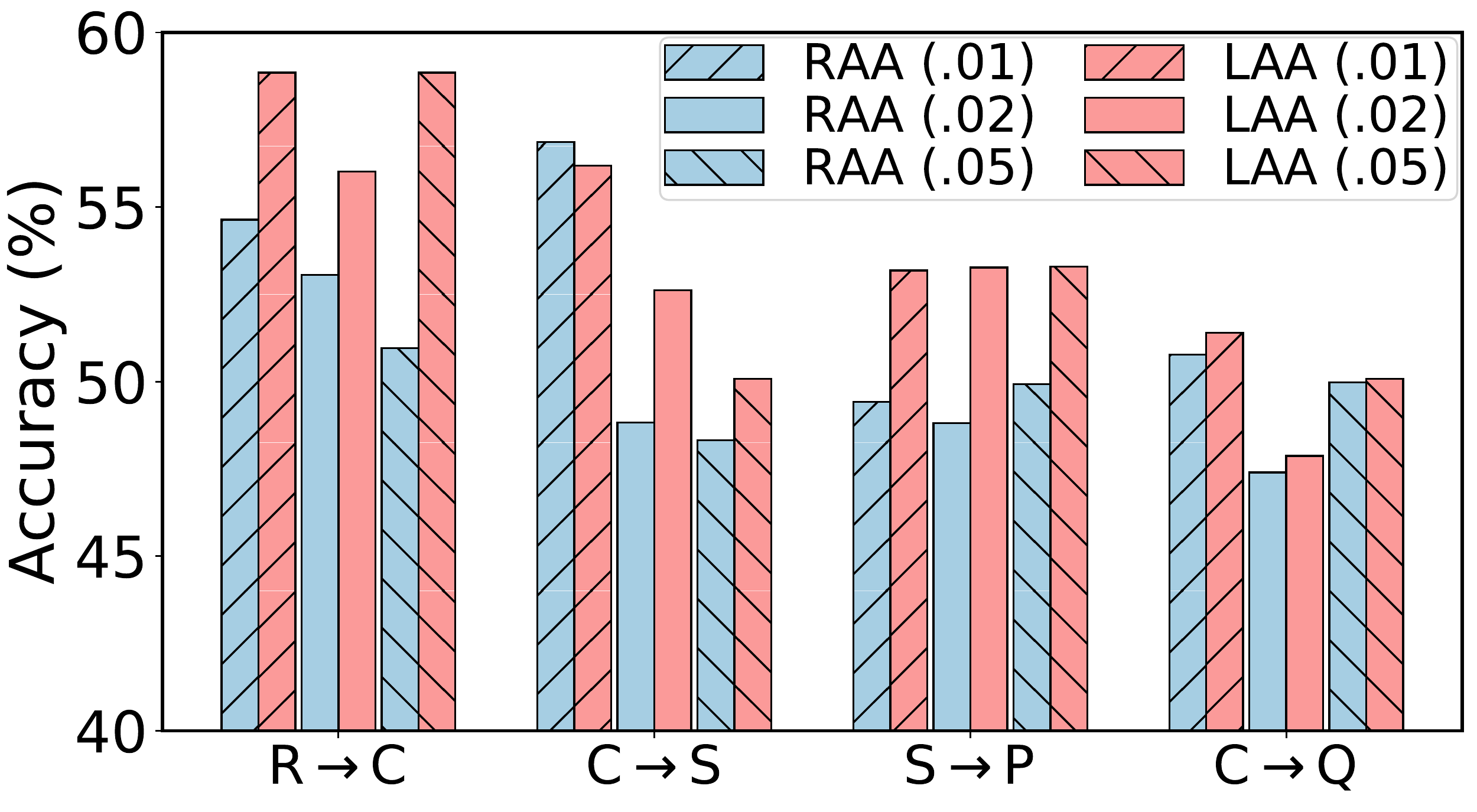}		
	\end{subfigure}		
	\caption{Increasing domain gap by adding a random displacement to source data. (left) DomainNet R$\rightarrow$C; (right) four DomainNet tasks with different $u$.}
       \label{fig:noise_dn}
	\vspace{-3mm}
\end{figure}

\cparagraph{When local context-aware augmentation matters?} Previous results validate the effectiveness of progressive anchor set augmentation. In terms of the two strategies, random (RAA) and local context-aware (LAA), there is a trade-off in exploitation and exploration. RAA can include target samples that are distant to the queried data. However, confident samples may be misclassified due to domain distribution shift. To illustrate the situation when local context matters, we manually increase domain gap by adding noises to the latent features. Given a pretrained ResNet, we first obtain the class prototypes $\{p_c\}$ of source data. Then for each class $c$, we generate a random displacement $\xi_c\triangleq \sum_i r_i(p_i-p_c)$, where $r_i\sim {U}(-u,u)$, and apply it to the entire source data of Class-$c$ before making predictions. Target data are unmodified. It effectively increases the domain gap, while preserving the source domain semantic structure. Fig.~\ref{fig:noise_dn} plots the comparison results. The benefit of local context-aware augmentation is more significant when $u$ is large where source and target domains have a large gap. The performance of RAA drops much faster due to the inclusion of false confident target samples. 

\cparagraph{Ablation studies on components of LADA.} Table~\ref{tab:abl} presents ablation studies on each component of LADA. Improvements in the second row over the first row indicate that it is critical to select a diverse subset in LAS. With anchor set augmentation, the accuracies are increased by +2.6\% on Office-Home and +2.3\% on Office-Home RSUT. This verifies that fine-tuning model on only queried data is ineffective. Using RandAug to create mixed images lead to further gains. To show the effectiveness of Class Balancing rejection, we report per-class average accuracies on Office-Home RSUT, where it is boosted by +1.0\% in the last row.  

\begin{table}[!t]
	\caption{Ablation studies on Office-Home using 5\%-budget and Office-Home RSUT using 10\%-budget. Div-Sel: select a diverse subset in LAS; $\mathcal{A}$-Aug: augment $\mathcal{A}$ with confident neighbors; CBR: class balancing rejection. ($^\dagger$Per-class average accuracy.)}
	\vspace{-3mm}\footnotesize
	\centering
	\scalebox{0.9}{
		\begin{tabular}{p{1.0cm}<{\centering}@{}p{1.4cm}<{\centering}@{}p{1.4cm}<{\centering}@{}p{1.4cm}<{\centering}|p{0.8cm}<{\centering}|p{0.6cm}<{\centering}p{0.6cm}<{\centering}}
			\toprule
			Div-Sel & $\mathcal{A}$-Aug & RandAug & CBR & OH &  \multicolumn{2}{c}{OH-RSUT} \\ 	
			\midrule
			& & &  & 72.6  & 70.6 & 70.3$^\dagger$ \\
			$\checkmark$ & & & & 75.3 & 73.2 & 72.5$^\dagger$\\ 
			$\checkmark$ & $\checkmark$ & &   & 77.9  & 75.5 & 74.2$^\dagger$\\ 
			$\checkmark$ & $\checkmark$ & & $\checkmark$  & 79.5 & 75.6 & 74.5$^\dagger$\\ 
			$\checkmark$ & $\checkmark$ &  $\checkmark$ &  & 79.6 & 76.2 & 74.3$^\dagger$\\
			$\checkmark$ & $\checkmark$ &  $\checkmark$ &   $\checkmark$ & 80.0 & 76.8 & 75.3$^\dagger$\\
			\bottomrule
	\end{tabular} }
	\label{tab:abl}
	\vspace{-3mm}
\end{table}

\vspace{-2mm}
\section{Conclusion}
In this paper, we advocate to utilize the local context of queried data for active domain adaptation. We propose a local context-aware active selection method based on the local inconsistency of model predictions. It consistently selects more informative samples than previous criteria. Then we propose a progressive anchor set augmentation module to mitigate issues from small labeling budgets. It utilizes queried data and their expanded neighbors to refine the model. Extensive experiments validate that our full method, named LADA (Local context-aware Active Domain Adaptation), surpasses state-of-the-art ADA solutions.


\appendix

\renewcommand{\thefigure}{A.\arabic{figure}}
\setcounter{figure}{0}
\renewcommand{\thetable}{A.\arabic{table}}
\setcounter{table}{0}
\renewcommand{\theequation}{A.\arabic{equation}}
\setcounter{figure}{0}

\section{Dataset Details}

\textbf{Office-31}~\cite{saenko2010adapting} contains 31 classes of 4,110 office environment related images. It has three domains: Amazon (A), DSLR (D) and Webcam (W). \textbf{Office-Home} is a similar dataset, containing 15,500 office images from 65 classes, split in four domains: Product (Pr), Clip Art (Cl), Artistic (Ar) and Real-World (Rw). \textbf{Office-Home RSUT}~\cite{tan2019generalized} is a subset of Office-Home created with the protocol of Reverse-unbalanced Source and Unbalanced Target to have a large label distribution shift. The major classes in the `RS' fold become minor classes in the `UT' fold, while the minor classes in the `RS' fold become major classes in the `UT' fold. \textbf{VisDA}~\cite{peng2017visda} is a large-scale Synthetic-to-Real dataset of 12 objects. The training set contains 152,397 synthetic 2D renderings of 3D models and the validation set contains 55,388 real images. We use the training set as the source domain and the validation set as the target domain. \textbf{DomainNet}~\cite{peng2019moment} consists of about 0.6 million images from 345 classes, distributed in six domains. Following~\cite{prabhu2021active,hwang2022combating}, we use five domains: Real (R), Clipart (C), Painting (P), Sketch (S), and Quickdraw (Q) for experiments. 

\section{Implementation Details}
We implement all experiments with PyTorch 1.8. Results are run on servers with NVIDIA A5000/A6000 GPU. Following previous ADA works~\cite{fu2021transferable,xie2022learning,xie2021active}, we use ResNet-50~\cite{he2016deep} pretrained on ImageNet~\cite{ILSVRC15} as the backbone network, a bottleneck layer (\texttt{Linear->BatchNorm1d}), and a classification head of one single \texttt{Linear} layer. The bottleneck feature dimension is 256. Training images are first resized to 256$\times$256, and then randomly cropped to 224$\times$224. Test images use center cropping instead. We adopt Adadelta optimizor with learning rate of 0.1 and a batch size of 32. On Office-Home and Office-31, we first train the models on only source data for 10 epochs, and then train on both source and target data with active domain adaptation for 30 epochs. At the epoch of 10, 12, 14, 16, 18, $B/5$ target data are selected for querying labels, where $B$ is the labeling budget. On VisDA, we conduct source-only training for 1 epoch and ADA for 10 epochs. On DomainNet, we conduct source-only training for 10 epochs and ADA for another 10 epochs. Mean accuracies of 3 repeated experiments are reported.

\begin{figure}[!h]
	\begin{center}	\includegraphics[width=0.9\linewidth]{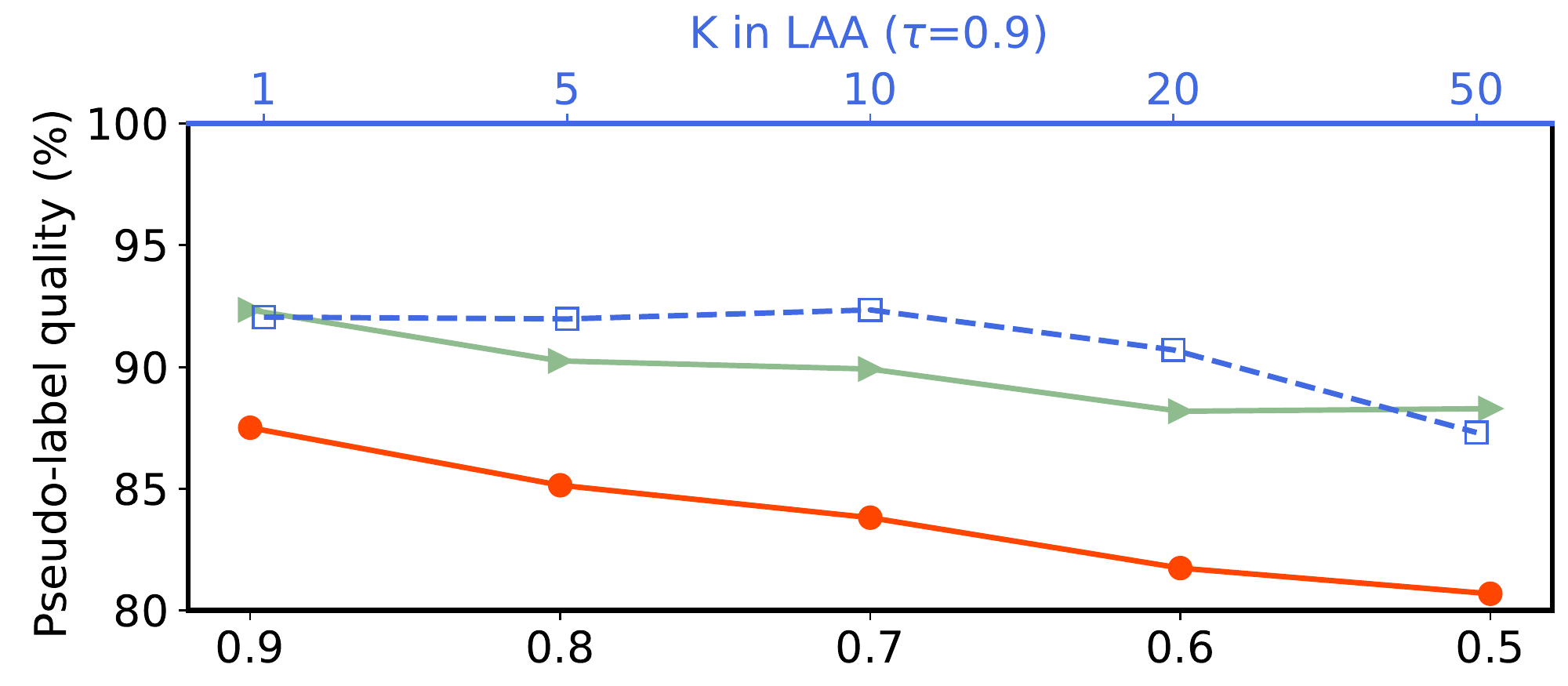}
    \includegraphics[width=0.9\linewidth]{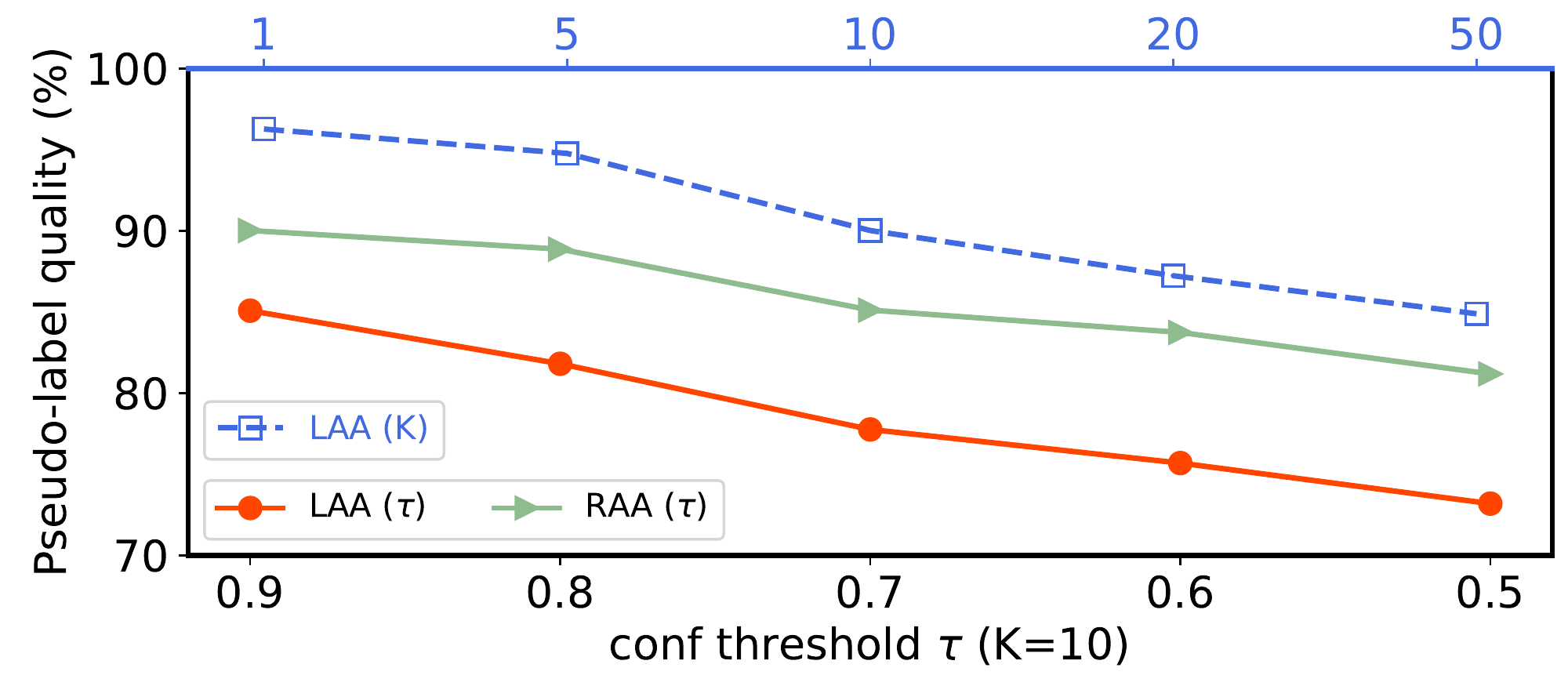}
	\end{center}
 \caption{Pseudo-label quality averaged over {Office-Home Cl$\shortrightarrow$\{Ar, Pr, Rw\}} (upper) and {Office-Home RSUT C$\shortrightarrow$\{P, R\}} (lower) using 10\%-budget.}
 \label{fig:supp:pl_quality}
\end{figure}

\section{Additional Results and Analyses}
\cparagraph{Running time.} Table~\ref{tab:supp:time} reports running time in seconds with one A6000 GPU, including active sampling (AL) time averaged over 5 rounds (10\%-budget) and model update (DA) time averaged over all training epochs. LAS consumes much less time than CLUE and slightly more than other AL methods. RAA/LAA is comparable to MCC and faster than CDAC.

\cparagraph{Pseudo-label quality of LAA/RAA.} We conduct experiments with different confidence thresholds $\tau$ and report the percentage of target samples in the anchor set with correct pseudo-labels. Shown in Fig.~\ref{fig:supp:pl_quality}, as $\tau$ reduces, the pseudo-label quality decreases. LAA has better pseudo-label quality than RAA. We also report results by fixing $\tau=0.9$ and varying neighborhood size $K$ in LAA (dashed lines). Overall, $\tau=0.9$ used in the paper leads to a decent pseudo-label quality. 

\begin{figure*}[!h]
	\centering	
	\includegraphics[width=0.36\textwidth]{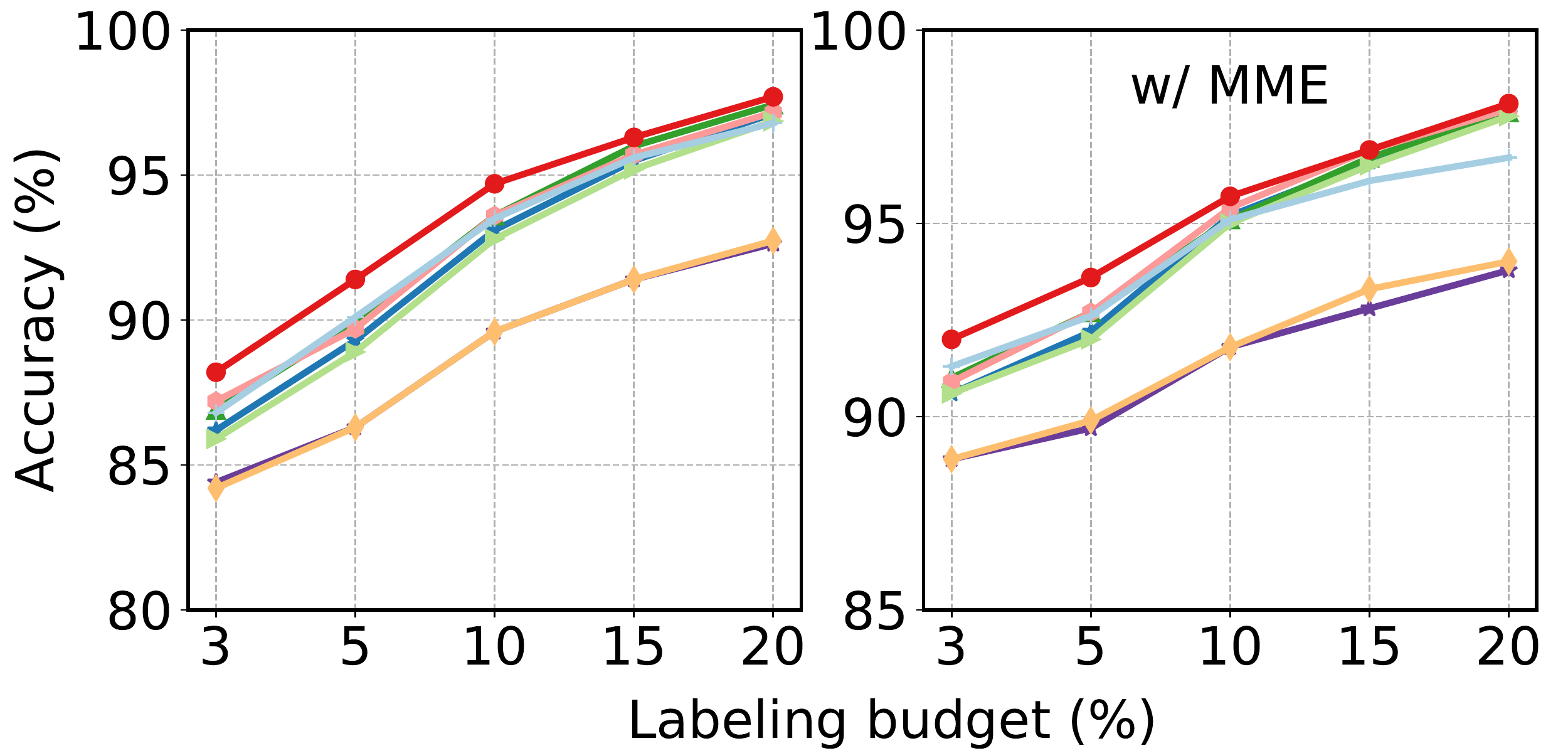}	
	\hfil
	\includegraphics[width=0.28\textwidth]{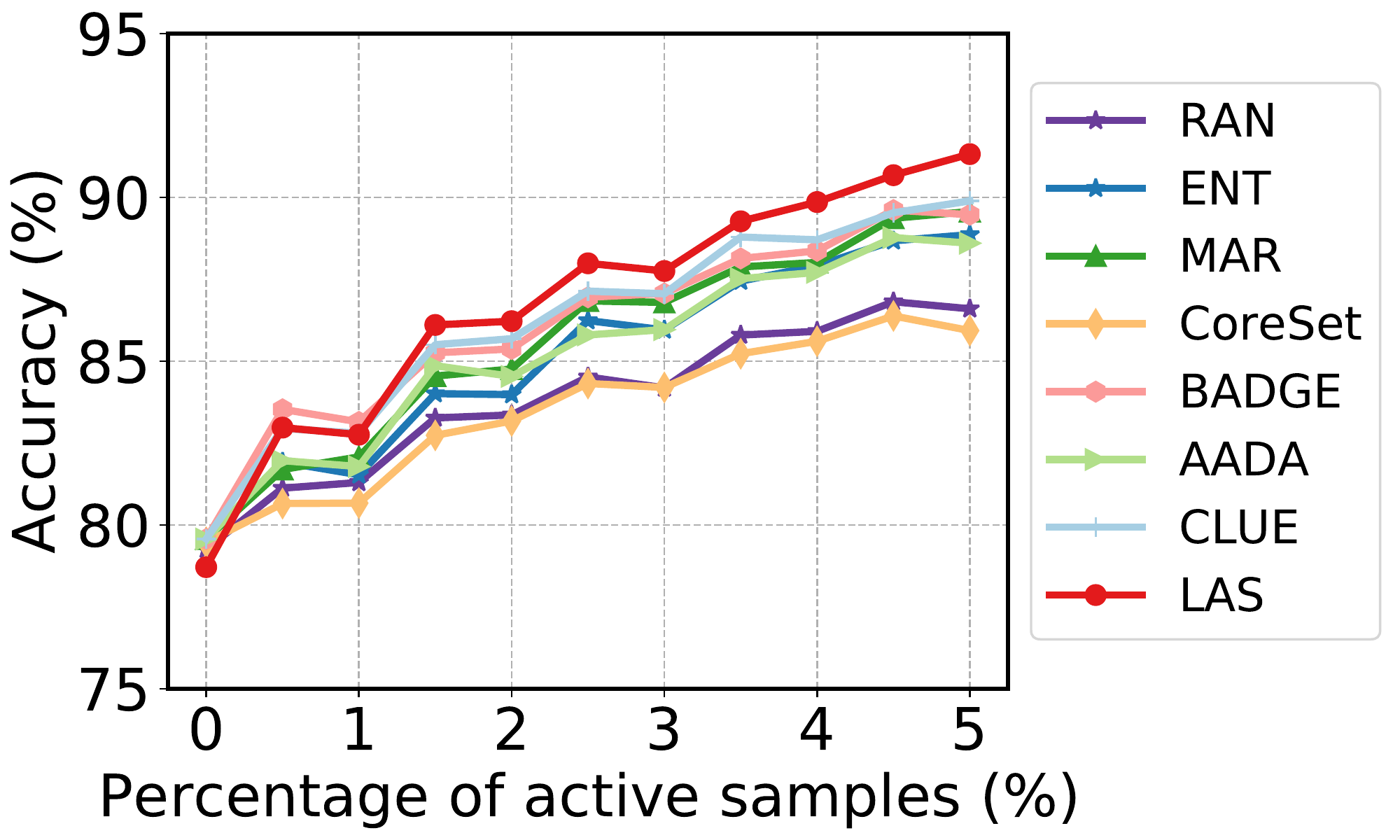}	
	\hfil
	\includegraphics[width=0.3\textwidth]{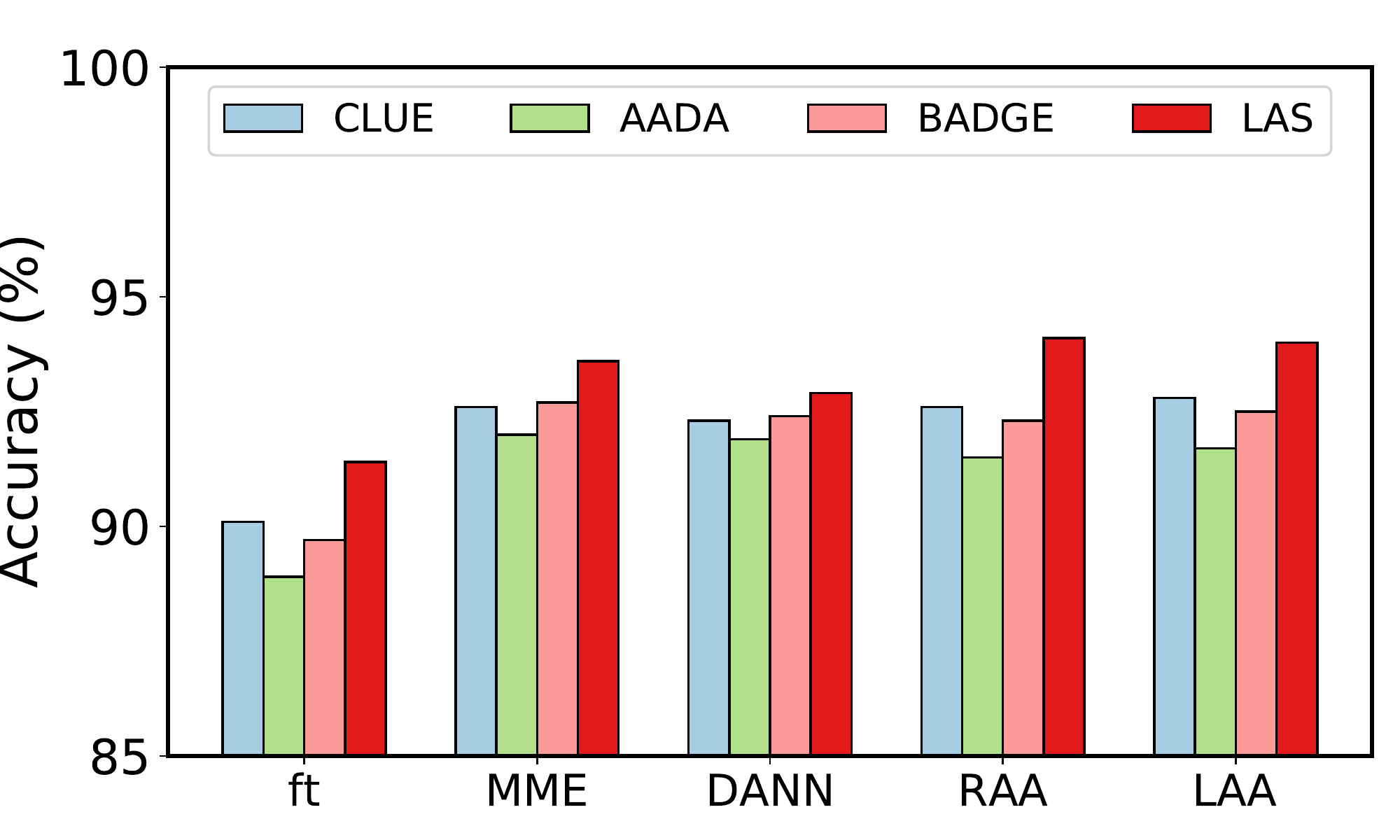}	
	\caption{Analysis on Office-31. (Left) varying labeling budget; (Center) accuracy curves with 5\%-budget; (Right) combining AL criteria with different DA strategies.}
	\label{fig:sup:Office31_budgets_DA}
\end{figure*}

\cparagraph{Comparison with other criteria on Office-31}
Figure~\ref{fig:sup:Office31_budgets_DA} presents analyses on Office-31 similar to that on Office-Home in the Fig.~\ref{fig:tsne} of the main paper. In the left figure, our LAS outperforms other active leaning criteria for labeling budgets ranging from 3\% to 20\%. When the labeling budget is small (\eg, 3\% or 5\%), LAS boosts the accuracy by a large margin. Since in ADA the situation with a small labeling budget is more important, it shows the effectiveness of LAS. In the center figure with 5\%-budget, the curve of LAS lies above others after 1\% samples are selected. In the right figure, when combined with five different domain adaptation strategies, LAS consistently achieves the highest accuracies than three other active learning criteria that are previous arts.

\begin{figure*}[!t]
	\centering	
	\begin{subfigure}[]{0.175\textwidth}
		\includegraphics[width=\textwidth]{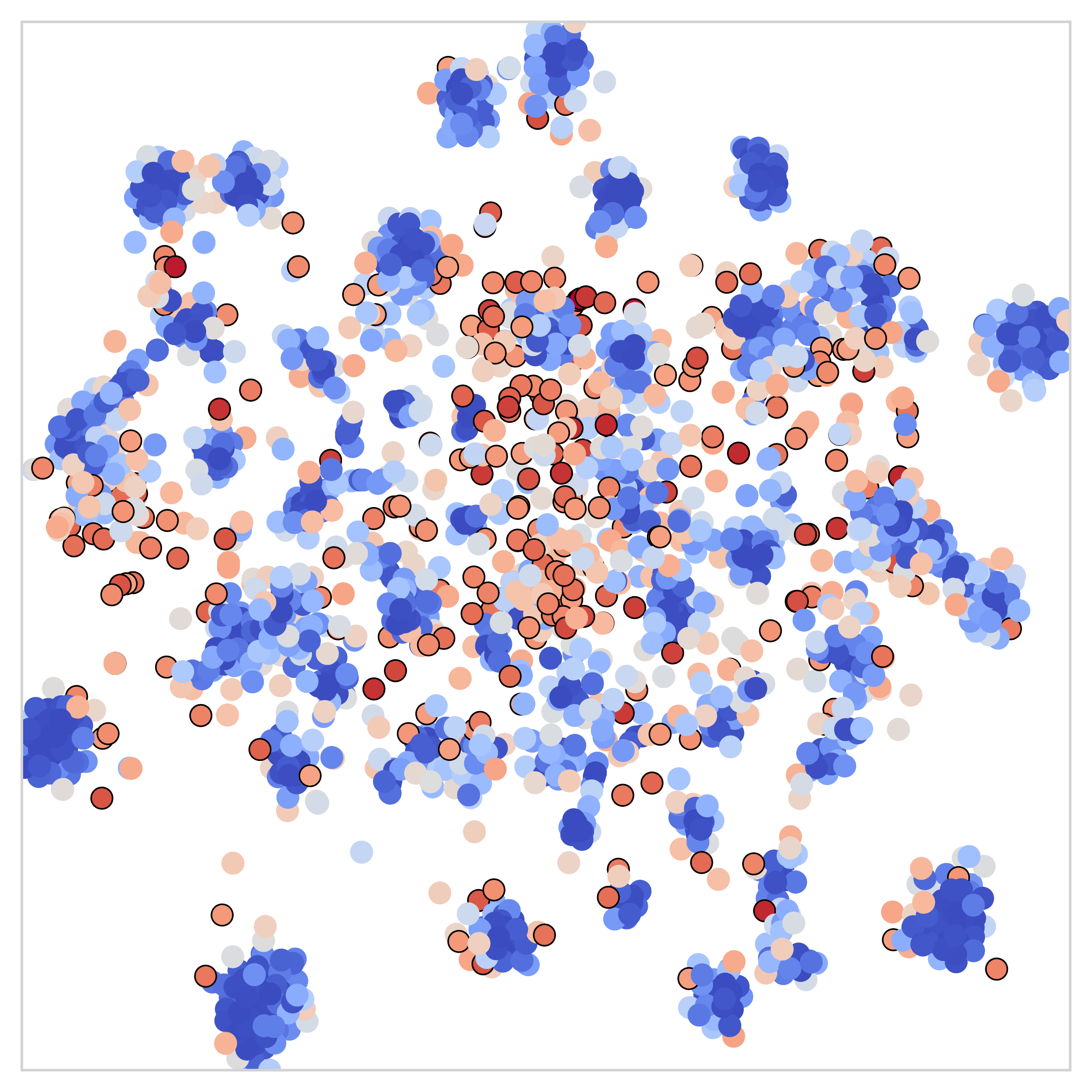}
		\caption{entropy}
		\label{fig:sup:tsne:entropy}
	\end{subfigure}	
	\begin{subfigure}[]{0.175\textwidth}
		\includegraphics[width=\textwidth]{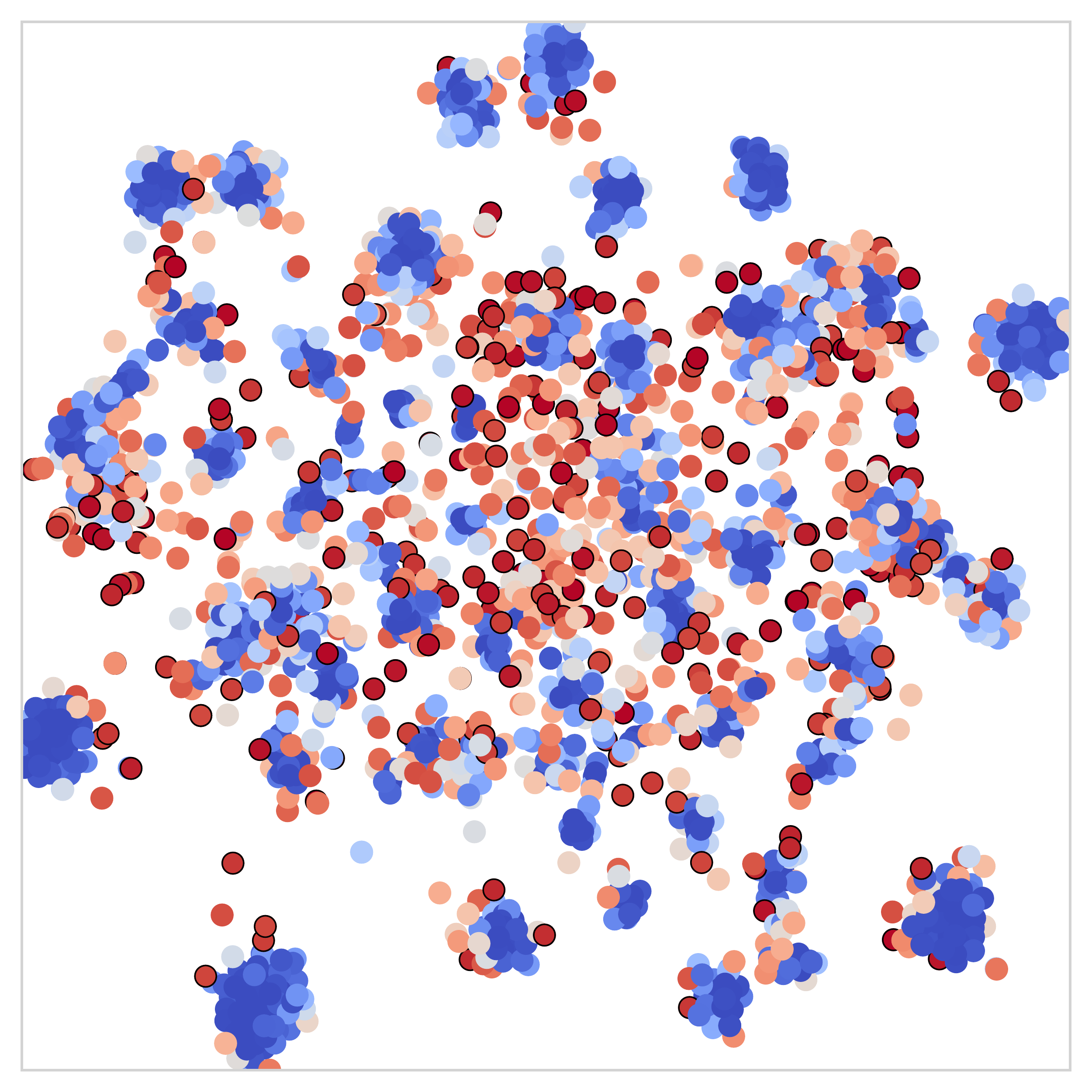}
		\caption{margin}
		\label{fig:sup:tsne:margin}
	\end{subfigure}	
	\begin{subfigure}[]{0.175\textwidth}
		\includegraphics[width=\textwidth]{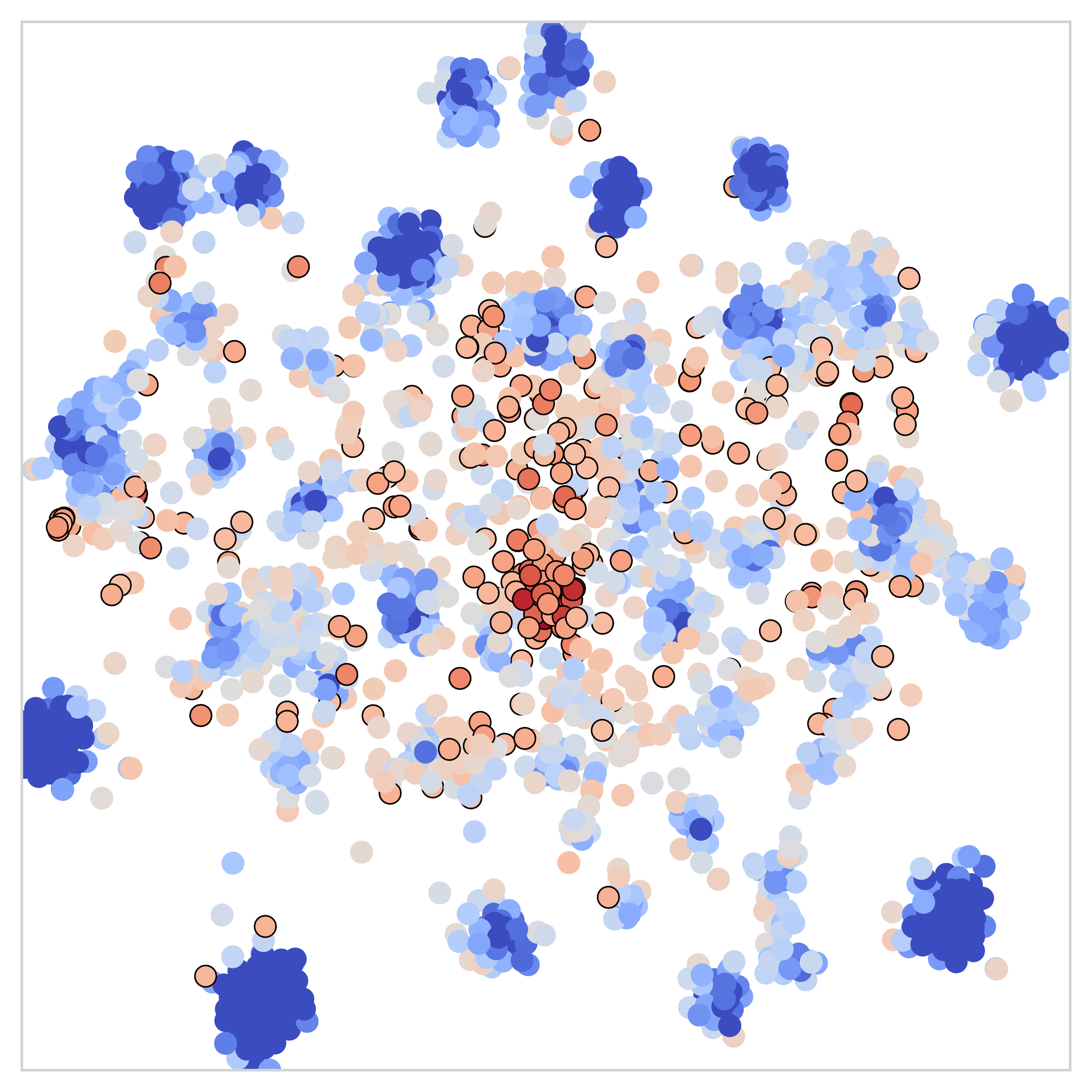}
		\caption{NAU}
		\label{fig:sup:tsne:NAU}
	\end{subfigure}	
	\begin{subfigure}[]{0.175\textwidth}
		\includegraphics[width=\textwidth]{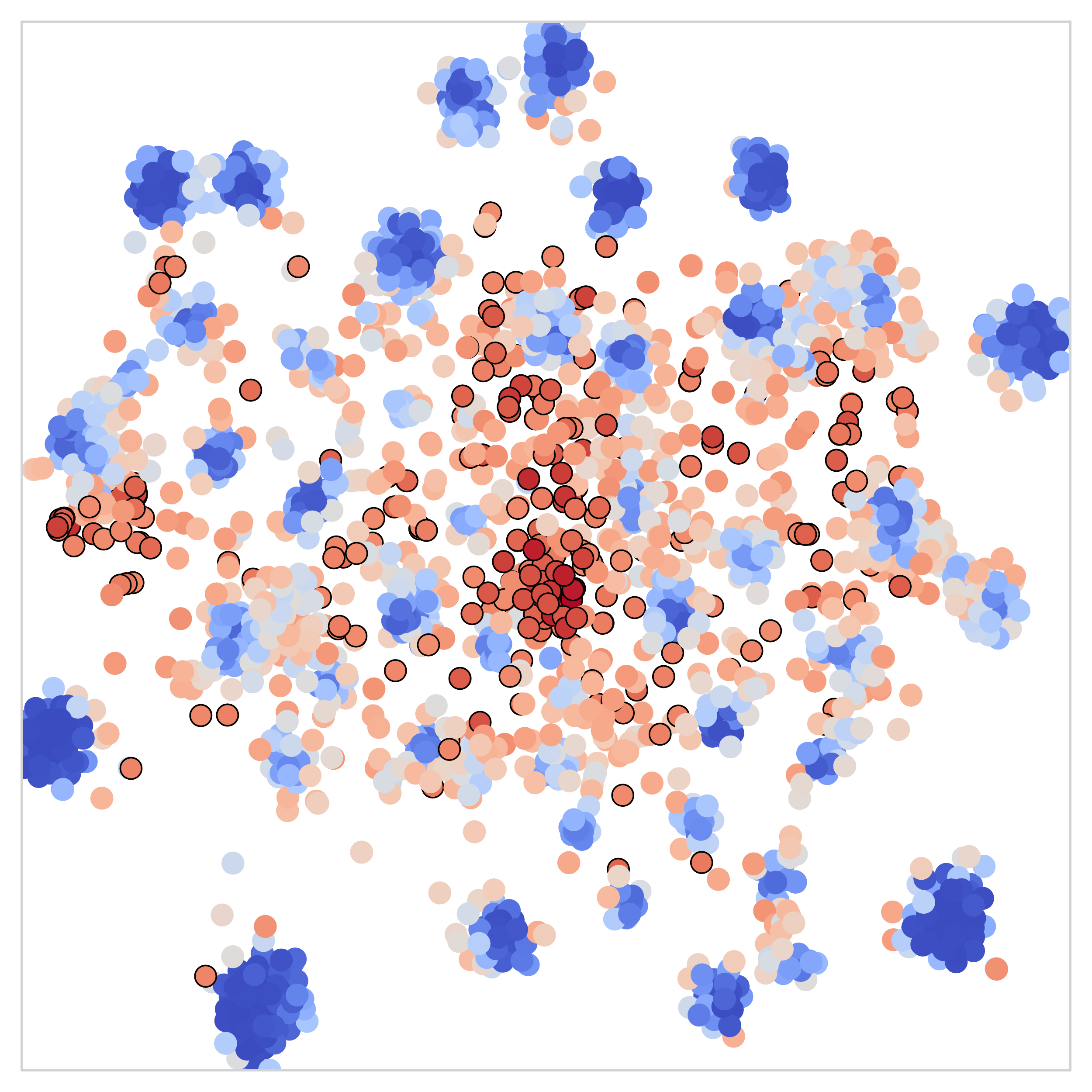}
		\caption{LI}
		\label{fig:sup:tsne:LI}
	\end{subfigure}	
    \begin{subfigure}[]{0.028\textwidth}
    \includegraphics[width=\textwidth]{tsne/colorbar.pdf}
    \end{subfigure}	
	\begin{subfigure}[]{0.24\textwidth}
		\includegraphics[width=\textwidth]{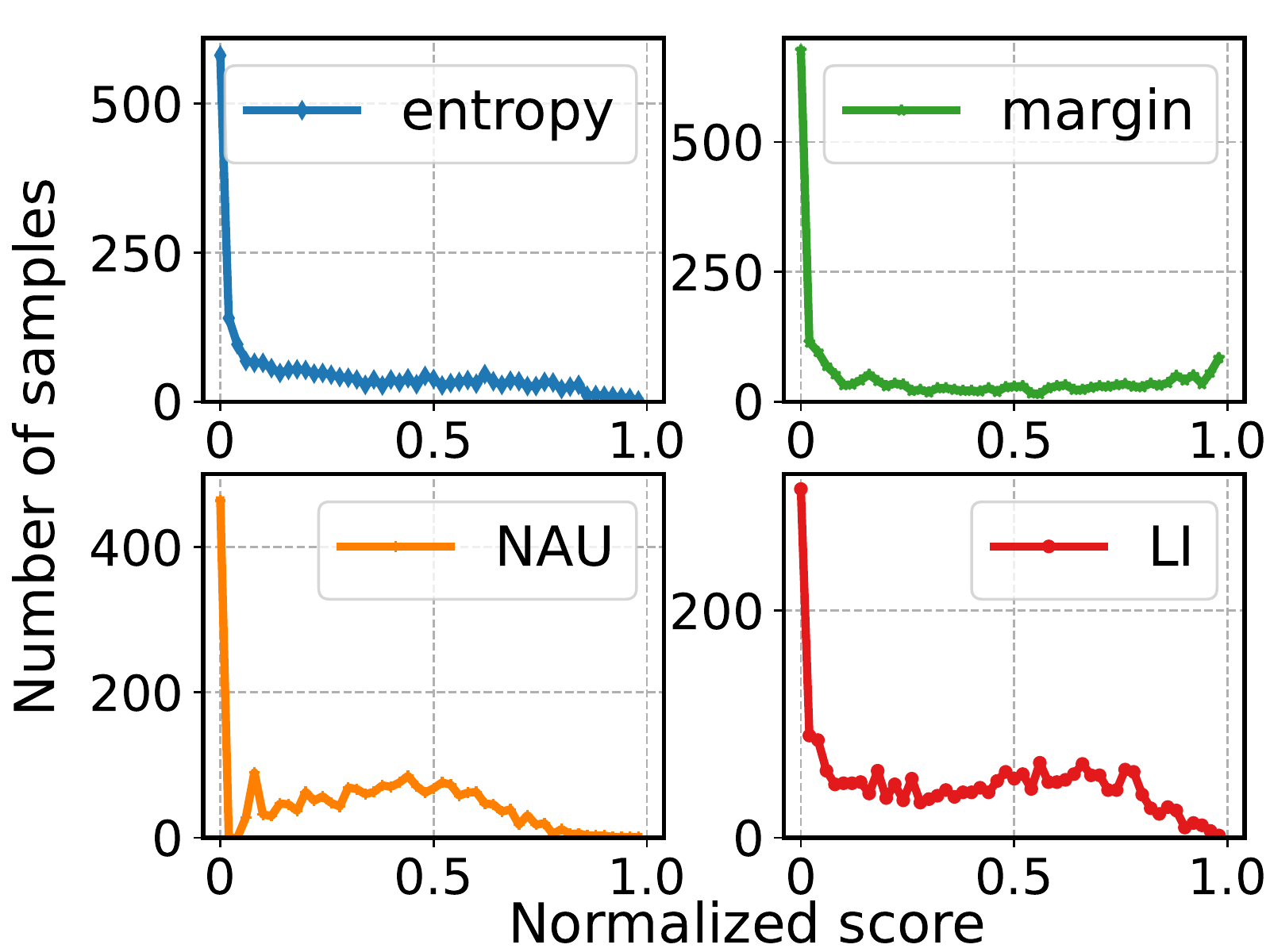}
		\caption{histogram}
		\label{fig:sup:tsne:histogram}
	\end{subfigure}	
	\caption{(a-d) $t$-SNE visualization of target features on Office-Home Rw$\rightarrow$Ar. Samples are colored according to their normalized uncertainty scores, where red indicates large values and blue indicates small values. The top 10\% samples with highest scores are marked with black boarders. (e) Histogram of target samples by normalized scores. }
	\label{fig:sup:tsne}
\end{figure*}

\begin{table*}[!h]
\caption{Running time on \textbf{Office-Home Ar$\shortrightarrow$Rw} and \textbf{VisDA} in seconds.}
\label{tab:supp:time}
    \centering
    \scriptsize
    \begin{tabular}{p{1.5cm}<{\centering}|p{1.1cm}<{\centering}|p{1.1cm}<{\centering}|p{1.1cm}<{\centering}|p{1.1cm}<{\centering}|p{1.1cm}<{\centering}|p{1.1cm}<{\centering}|p{1.1cm}<{\centering}|p{1.1cm}<{\centering}|p{1.1cm}<{\centering}|p{1.1cm}<{\centering}}
    \toprule
    & \multicolumn{5}{c|}{AL} & \multicolumn{5}{c}{DA} \\
    \cmidrule{2-11}
     & BADGE & AADA & CLUE & MHPL & LAS &  ft & MCC & CDAC & RAA & LAA\\
     \midrule
     Ar$\rightarrow$Rw   &  17.6\tiny{$\pm$0.5} & 18.4\tiny{$\pm$0.8} & \textbf{24.0}\tiny{$\pm$0.9} & 17.8\tiny{$\pm$0.4} & 18.2\tiny{$\pm$0.7} & 30.2\tiny{$\pm$8.7} & 39.6\tiny{$\pm$8.6} & \textbf{67.4}\tiny{$\pm$16.5} & 63.0\tiny{$\pm$18.2} & 62.0\tiny{$\pm$10.8} \\
     VisDA & 80.4\tiny{$\pm$2.4} & 57.0\tiny{$\pm$1.7} & \textbf{733.2}\tiny{$\pm$56.6} & 70.0\tiny{$\pm$2.6} & 105.6\tiny{$\pm$17.9} & 1136.6\tiny{$\pm$45.3} & 1623.4\tiny{$\pm$18.4} & \textbf{2662.0}\tiny{$\pm$17.2} & 1449.8\tiny{$\pm$15.8} & 1624.2\tiny{$\pm$245.2} \\
    \midrule
    \end{tabular}
\end{table*}

\cparagraph{Remarks on SSDA.}
Semi-supervised DA (SSDA) is closely related to Active DA. In both task, a few labeled target data and many unlabeled target data are available. Yet there are some differences. In SSDA, all labeled target data are provided for once at the beginning of training and fixed afterwards. While in ADA, labeled target data are actively selected. The active querying and model update interleave for several rounds during the training of ADA.

Existing SSDA methods can be directly applied in ADA. In the paper, we have compared different active query methods when using MME~\cite{saito2019semi} and CDAC~\cite{li2021cross} as the model adaptation methods. Additional results on Office-Home with CDAC is provided in Tab.~\ref{tab:sup:officehome_ftj}. Generally, the proposed LAS can select more informative samples than other active selection criteria.  

Nevertheless, it may be sub-optimal to simply combine active query with existing SSDA methods. A unified ADA solution that considers both active query and model adaptation would be better effective. Tables~\ref{tab:sup:officehome_ssda}, \ref{tab:sup:officehome_rsut_ssda} present comparison results with two state-of-the-art SSDA methods, ECACL~\cite{li2021ecacl} and CDAC~\cite{li2021cross}. The comparison results are taken from~\cite{hwang2022combating}. It should be noted that although ADA can select more informative labeled samples, the performance is also affected by the way to utilize unlabeled data. From the tables, ECACL and CDAC surpass three early ADA methods. The state-of-the-art LAMDA method selects target data to approximate the entire target distribution, and addresses the issue of label distribution mismatch between source and target domains. It obtains better performances than SSDA arts. Our proposed LADA (LAS w/ LAA) selects locally-representative samples, and progressively expand the labeled data with confident samples in a class-balanced manner. LADA outperforms LAMDA by +2.3\% on Office-Home and +1.0\% on Office-Home RSUT. When replacing LAA with CDAC in LADA, the performance drops, as we show in the paper.

\begin{table*}[!t]
	\caption{Accuracies (\%) on \textbf{Office-Home} with 5\% labeled target samples. ($^\dagger$Training mini-batches are sampled from a joint labeled set.)}
	\footnotesize
	\centering
	\scalebox{0.9}{
		\begin{tabular}{p{2cm}@{}p{2.5cm}@{}p{1.05cm}<{\centering}@{}p{1.05cm}<{\centering}@{}p{1.05cm}<{\centering}@{}p{1.05cm}<{\centering}@{}p{1.05cm}<{\centering}@{}p{1.05cm}<{\centering}@{}p{1.05cm}<{\centering}@{}p{1.05cm}<{\centering}@{}p{1.05cm}<{\centering}@{}p{1.05cm}<{\centering}@{}p{1.05cm}<{\centering}@{}p{1.05cm}<{\centering}@{}p{1.05cm}<{\centering}}
			\toprule
			AL method & DA method & Ar$\shortrightarrow$Cl & Ar$\shortrightarrow$Pr & Ar$\shortrightarrow$Rw &  Cl$\shortrightarrow$Ar & Cl$\shortrightarrow$Pr & Cl$\shortrightarrow$Rw &  Pr$\shortrightarrow$Ar & Pr$\shortrightarrow$Cl & Pr$\shortrightarrow$Rw &      
			Rw$\shortrightarrow$Ar & Rw$\shortrightarrow$Cl & Rw$\shortrightarrow$Pr & Avg.   \\ 	
			\midrule
        RAN & \multirow{10}{*}{ft w/ CE loss$^\dagger$} & 60.5 & 78.4 & 79.0 & 60.2 & 74.4 & 72.7 & 61.5 & 56.2 & 77.6 & 69.9 & 59.6 & 82.8 & 69.4 \\ 
        ENT & & 62.8 & 81.5 & 82.7 & 64.1 & 78.1 & 75.7 & 63.4 & 57.5 & 81.2 & 72.8 & 62.5 & 87.3 & 72.5 \\ 
        MAR & & 64.0 & 81.8 & 82.7 & 64.1 & 79.0 & 74.9 & 64.6 & 59.9 & 80.7 & 73.2 & 64.6 & 87.8 & 73.1 \\ 
        CoreSet & & 58.5 & 77.1 & 79.3 & 60.8 & 72.4 & 71.8 & 60.9 & 54.9 & 77.3 & 70.9 & 58.8 & 81.3 & 68.7 \\ 
        BADGE & & 65.5 & 83.6 & 82.1 & 63.1 & 79.8 & 75.3 & 64.9 & 61.0 & 80.8 & 73.1 & 65.1 & 87.1 & 73.5 \\ 
        AADA & & 61.8 & 82.0 & 82.1 & 62.3 & 77.7 & 76.0 & 63.1 & 59.4 & \underline{81.8} & 72.9 & 62.4 & 87.2 & 72.4 \\ 
        CLUE & & 65.3 & 81.8 & 81.7 & 62.6 & 78.5 & 74.8 & 63.9 & 61.4 & 79.9 & 72.9 & 63.1 & 87.6 & 72.8 \\ 
        CONF & & 63.4 & 81.9 & 82.9 & 63.8 & 78.2 & 75.8 & 64.2 & 60.2 & 81.6 & 73.3 & 63.2 & 87.4 & 73.0 \\ 
        MHPL & & 65.6 & 82.1 & 82.9 & \underline{65.3} & 79.1 & 74.6 & 64.7 & 61.4 & 81.6 & 73.3 & 63.7 & 88.1 & 73.5 \\ 
        \rc
        LAS & & \underline{67.2} & \underline{84.3} & \underline{83.1} & 65.1 & \underline{80.9} & \underline{77.0} & \underline{65.3} & \underline{62.5} & 81.4 & \underline{73.8} & \underline{66.7} & \underline{89.0} & \underline{74.7} \\ 
        \midrule
        RAN & \multirow{10}{*}{CDAC} & 61.6 & 78.8 & 80.1 & 67.7 & 80.2 & 77.6 & 68.7 & 61.9 & 79.7 & 74.1 & 63.0 & 85.2 & 73.2 \\ 
        ENT & & 62.9 & 81.9 & 83.4 & 69.0 & 82.0 & 80.0 & 70.3 & 63.3 & 84.2 & 75.6 & 67.7 & 87.1 & 75.6 \\ 
        MAR & & 65.6 & 83.8 & 83.3 & 69.0 & 83.7 & 81.0 & 70.2 & 65.7 & 84.6 & 75.9 & 67.0 & 88.1 & 76.5 \\ 
        CoreSet & & 58.9 & 77.7 & 79.6 & 67.1 & 77.9 & 77.2 & 67.4 & 58.6 & 81.6 & 73.6 & 63.4 & 83.4 & 72.2 \\ 
        BADGE & & 63.3 & 80.4 & 81.3 & 69.6 & 83.0 & 78.8 & 70.4 & 62.7 & 83.6 & 76.1 & 67.4 & 88.0 & 75.4 \\ 
        AADA & & 61.8 & 81.8 & 82.8 & 69.6 & 83.2 & 80.4 & 70.7 & 63.5 & 84.3 & 76.2 & 66.2 & 87.2 & 75.6 \\ 
        CLUE & & 65.2 & 83.6 & 82.3 & 68.8 & 84.4 & 79.8 & 69.7 & 64.9 & 83.6 & 75.2 & 68.0 & 87.5 & 76.1 \\ 
        CONF & & 62.6 & 82.9 & \underline{83.8} & 70.6 & 83.6 & 79.7 & 70.0 & 64.6 & 84.3 & 76.4 & 66.8 & 88.1 & 76.1 \\ 
        MHPL & & 65.5 & 82.4 & 82.7 & 70.8 & 84.1 & \underline{81.7} & 70.5 & 66.2 & 84.3 & 76.9 & 68.7 & 87.8 & 76.8 \\ 
        \rc
        LAS & & \underline{67.4} & \underline{85.4} & 83.1 & \underline{71.0} & \underline{85.0} & \underline{81.7} & \underline{72.1} & \underline{67.8} & \underline{85.1} & \underline{77.4} & \underline{70.4} & \underline{89.5} & \underline{78.0} \\ 
        \midrule
        \rc
	LAS & RAA & \textbf{71.2} &	\textbf{88.1} &	\textbf{85.3} &	\textbf{73.2} &	\textbf{87.8} &	\textbf{83.8} &	\textbf{72.6} &	\textbf{72.2} &	\textbf{86.6} &	79.2 &	74.4 &	\textbf{91.7} &	\textbf{80.5} \\
			\rc	 
			LAS & LAA & \textbf{71.2} &	87.4 &	84.6 &	72.1 &	87.0 &	83.6 &	71.5 &	71.6 &	85.3 &	\textbf{79.3} &	\textbf{75.5} &	90.4 &	80.0 \\
	\bottomrule
	\end{tabular} }
	\label{tab:sup:officehome_ftj}

\end{table*}

\begin{table*}[!t]
    \caption{Comparison with Semi-Supervised Domain Adaptation methods on \textbf{Office-Home} using 10\%-budget.}
    \footnotesize
    \centering
    \scalebox{0.85}{
        \begin{tabular}{p{2.2cm}p{2.2cm}@{}p{1.05cm}<{\centering}@{}p{1.05cm}<{\centering}@{}p{1.05cm}<{\centering}@{}p{1.05cm}<{\centering}@{}p{1.05cm}<{\centering}@{}p{1.05cm}<{\centering}@{}p{1.05cm}<{\centering}@{}p{1.05cm}<{\centering}@{}p{1.05cm}<{\centering}@{}p{1.05cm}<{\centering}@{}p{1.05cm}<{\centering}@{}p{1.05cm}<{\centering}@{}p{0.8cm}<{\centering}}
            \toprule
            Task & Method & Ar$\shortrightarrow$Cl & Ar$\shortrightarrow$Pr & Ar$\shortrightarrow$Rw &  Cl$\shortrightarrow$Ar & Cl$\shortrightarrow$Pr & Cl$\shortrightarrow$Rw &  Pr$\shortrightarrow$Ar & Pr$\shortrightarrow$Cl & Pr$\shortrightarrow$Rw &      
            Rw$\shortrightarrow$Ar & Rw$\shortrightarrow$Cl & Rw$\shortrightarrow$Pr & Avg.   \\ 	
            \midrule			
            \multirow{2}{*}{SSDA}  & ECACL & 72.2 & 86.7 & 82.8 & 70.5 & 85.0 & 82.6 & 70.9 & 71.5 & 82.9 & 76.0 & 74.0 & 88.9 & 78.7 \\
            & CDAC & 69.5 & 83.2 & 80.2 & 66.9 & 82.4 & 78.7 & 66.1 & 70.6 & 80.9 & 72.3 & 70.5 & 87.2 & 75.7 \\ 
            \midrule
            \multirow{5}{*}{ADA}  & TQS & 64.3 & 84.8 & 83.5 & 66.1 & 81.0 & 76.7 & 66.5 & 61.4 & 82.0 & 73.7 & 65.9 & 88.5 & 74.5 \\ 
            & CLUE & 62.1 & 80.6 & 73.9 & 55.2 & 76.4 & 75.4 & 53.9 & 62.1 & 80.7 & 67.5 & 63.0 & 88.1 & 69.9 \\
            & S$^3$VAADA & 67.8 & 83.9 & 82.9 & 67.0 & 81.4 & 79.5 & 65.8 & 65.9 & 82.4 & 74.8 & 68.6 & 87.9 & 75.7 \\ 
            & LAMDA & 74.8 & 88.5 & 86.9 & 73.8 & 88.2 & 83.3 & 74.6 & 75.5 & 86.9 & 80.8 & 77.8 & 91.7 & 81.9 \\
            \rc
            & LADA &  \textbf{77.2} & \textbf{91.9} &	\textbf{88.1} &	\textbf{76.9} & \textbf{91.1} &	\textbf{86.8} &	\textbf{76.6} &	\textbf{78.1} &	\textbf{88.3} &	\textbf{82.0} &	\textbf{79.0} &	\textbf{93.8} &	\textbf{84.2} \\	 
            \bottomrule
    \end{tabular} }

    \label{tab:sup:officehome_ssda}
\end{table*}
	
\begin{table}[!t]
	\caption{Comparison with SSDA methods on \textbf{Office-Home RSUT} using 10\%-budget.} 
	\centering
	\footnotesize
	\scalebox{0.85}{
		\begin{tabular}{p{1.4cm}p{1.4cm}p{0.4cm}p{0.4cm}p{0.4cm}p{0.4cm}p{0.4cm}p{0.4cm}p{0.4cm}<{\centering} }
			\toprule
			Task & Method & C$\shortrightarrow$P & C$\shortrightarrow$R & P$\shortrightarrow$C & P$\shortrightarrow$R & R$\shortrightarrow$C & R$\shortrightarrow$P & Avg.\\
			\midrule
			\multirow{2}{*}{SSDA}  & ECACL & 78.6 & 68.6 & 59.5 & 77.1 & 61.9 & 82.0 & 71.3 \\
			& CDAC & 73.0 & 58.7 & 55.8 & 73.3 & 50.3 & 77.3 & 64.7 \\
			\midrule
			\multirow{5}{*}{ADA}  & TQS & 69.4 & 65.7 & 53.0 & 76.3 & 53.1 & 81.1 & 66.4 \\ 
			& CLUE & 69.7 & 65.9 & 57.1 & 73.4 & 59.5 & 82.7 & 68.1 \\ 
			&  S$^3$VAADA & 73.0 & 63.0 & 50.7 & 69.6 & 52.6 & 78.3 & 64.5 \\
			& LAMDA & 81.2 & 75.7 & \textbf{64.1} & 81.6 & 65.1 & 87.2 & 75.8 \\ 
			\rc
			 & LADA & \textbf{83.2} &	\textbf{77.2} &	63.8 &	\textbf{83.0} &	\textbf{65.4} &	\textbf{88.1} &	\textbf{76.8} \\		
			\bottomrule
	
		\end{tabular}
	}
	\label{tab:sup:officehome_rsut_ssda}

\end{table}

\cparagraph{Uncertainty measures in LAS.} Figure~\ref{fig:sup:tsne} visualizes the target features on Office-Home Rw$\rightarrow$Ar. Similar to the plots in Fig.~3 of the paper, entropy and margin include some outliers in the top 10\% samples (see circles with black boarders in the bottom part of Figs.~\ref{fig:sup:tsne:entropy}, \ref{fig:sup:tsne:margin}). For NAU in Fig.~\ref{fig:sup:tsne:NAU}, target data have small normalized scores. Target data with high normalized LI scores form several small clusters in Fig.~\ref{fig:sup:tsne:LI}. From the histogram in  Fig.~\ref{fig:sup:tsne:histogram}, a maximal can be observed around 0.6-0.7 for LI. These phenomenons are similar to Office-31 W$\rightarrow$A in the paper.

\cparagraph{Training with a joint labeled set.}
In the implementation of some early ADA works~\cite{fu2021transferable,xie2022learning}, the queried labeled data are added to the source labeled data, and training mini-batches are sampled from this joint labeled set. The objective is
\begin{equation}\label{eq:sup:lossoj}
\mathcal{L}=\mathbb{E}_{(x,y)\sim \mathcal{D}_s\cup \mathcal{D}_{tl}}\ell_{\rm ce}(h(x),y)
\end{equation}
where $\ell_{\rm ce}$ is the cross entropy loss. Differently, recent works~\cite{xie2021active,hwang2022combating} and ours adopt
\begin{equation}\label{eq:sup:losso}
\!\!\mathcal{L}=\mathbb{E}_{(x,y)\sim \mathcal{D}_s}\ell_{\rm ce}(h(x),y)+\mathbb{E}_{(x,y)\sim \mathcal{D}_{tl}}\ell_{\rm ce}(h(x),y)
\end{equation}
Comparing Eq.~\ref{eq:sup:lossoj} with Eq.~\ref{eq:sup:losso}, the advantage of training with a joint labeled set is that it only needs to back-propagate through one batch of data, thus reducing the memory and computation usage. The disadvantage is that the labeled data set is dominated by the source data. When there is a large domain gap (\eg when label distribution shift exists), the performance may be hurt. 

Nevertheless, to better demonstrate the effectiveness of LAS, Table~\ref{tab:sup:officehome_ftj} lists the results using fine-tuning with a joint labeled set. Accuracies are slightly lower than their counterparts in Table~1 of the main paper. LAS still achieves the best scores among all AL methods.

\begin{figure}[t]
 	\centering
 	\includegraphics[width=\linewidth]{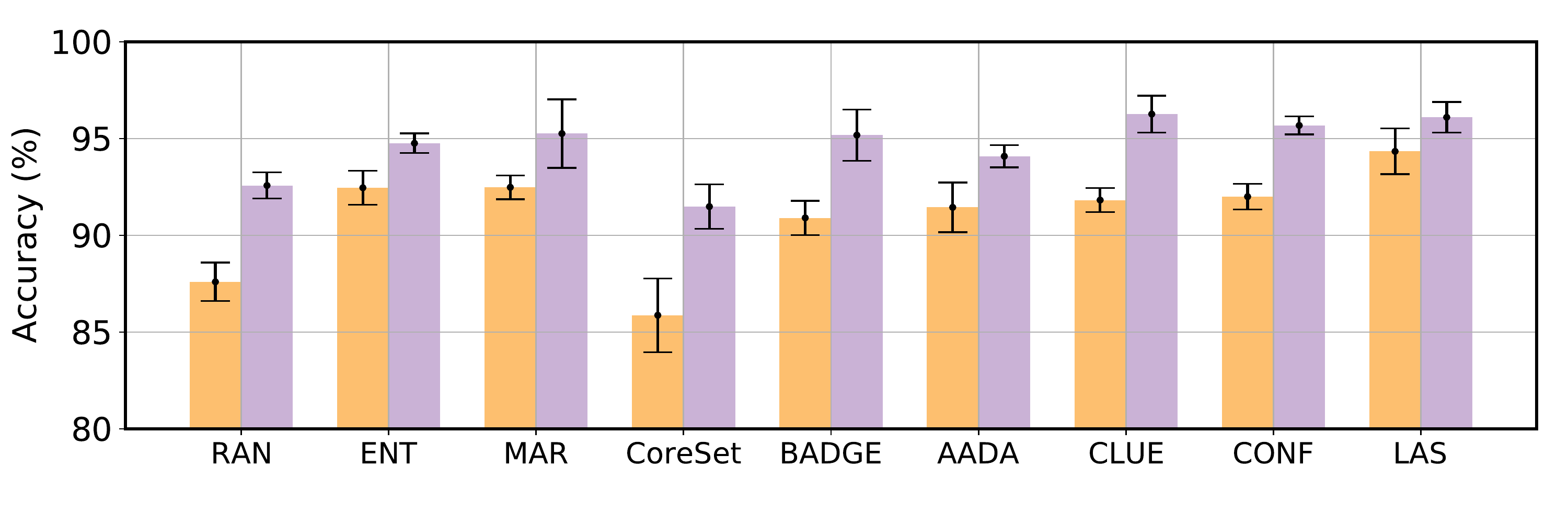}
 	\includegraphics[width=\linewidth]{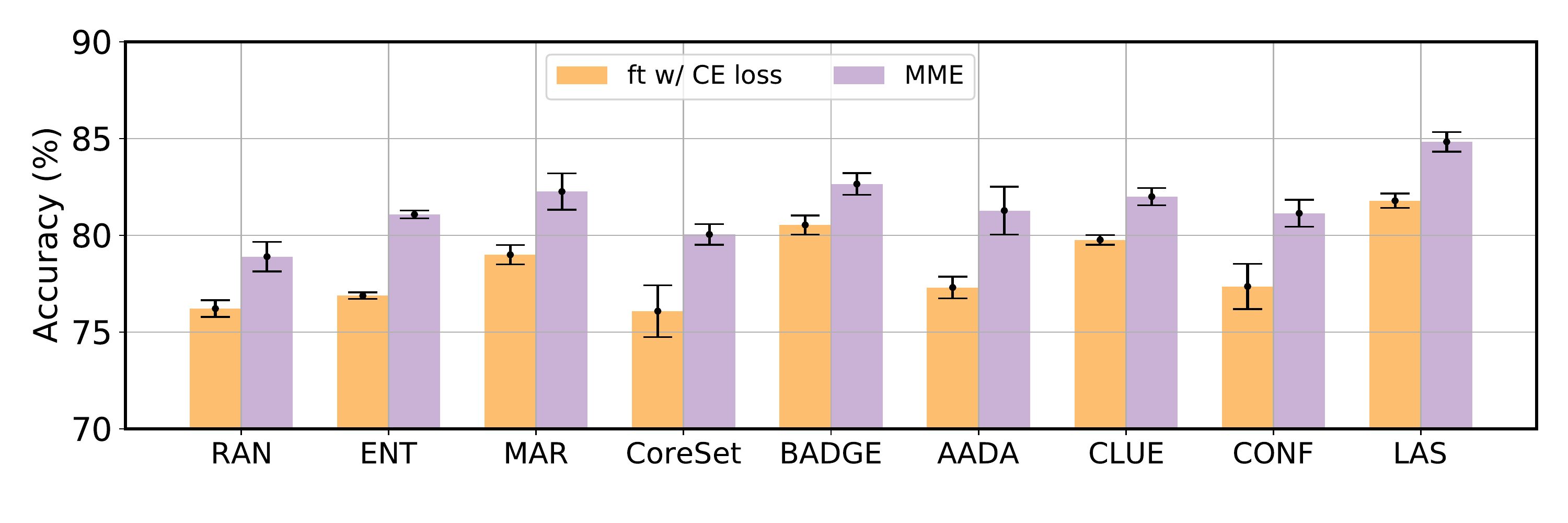}
 	\caption{Visualization of standard deviations on Office-31 A$\rightarrow$W (upper) and  W$\rightarrow$A (lower) using 5\%-budget.}
 	\label{fig:sup:std}

 \end{figure}

\cparagraph{Visualization of standard deviations.} Figure~\ref{fig:sup:std} plots the standard deviations on two Office-31 tasks over 3 repeated experiments. Performances are relatively stable to different random initializations. Of all active selection methods, our proposed LAS obtains the highest average accuracies. 

\begin{figure*}[!t]
\centering	
\begin{minipage}{0.99\linewidth}
    \hspace{1.5mm}	
    \includegraphics[width=0.5\linewidth]{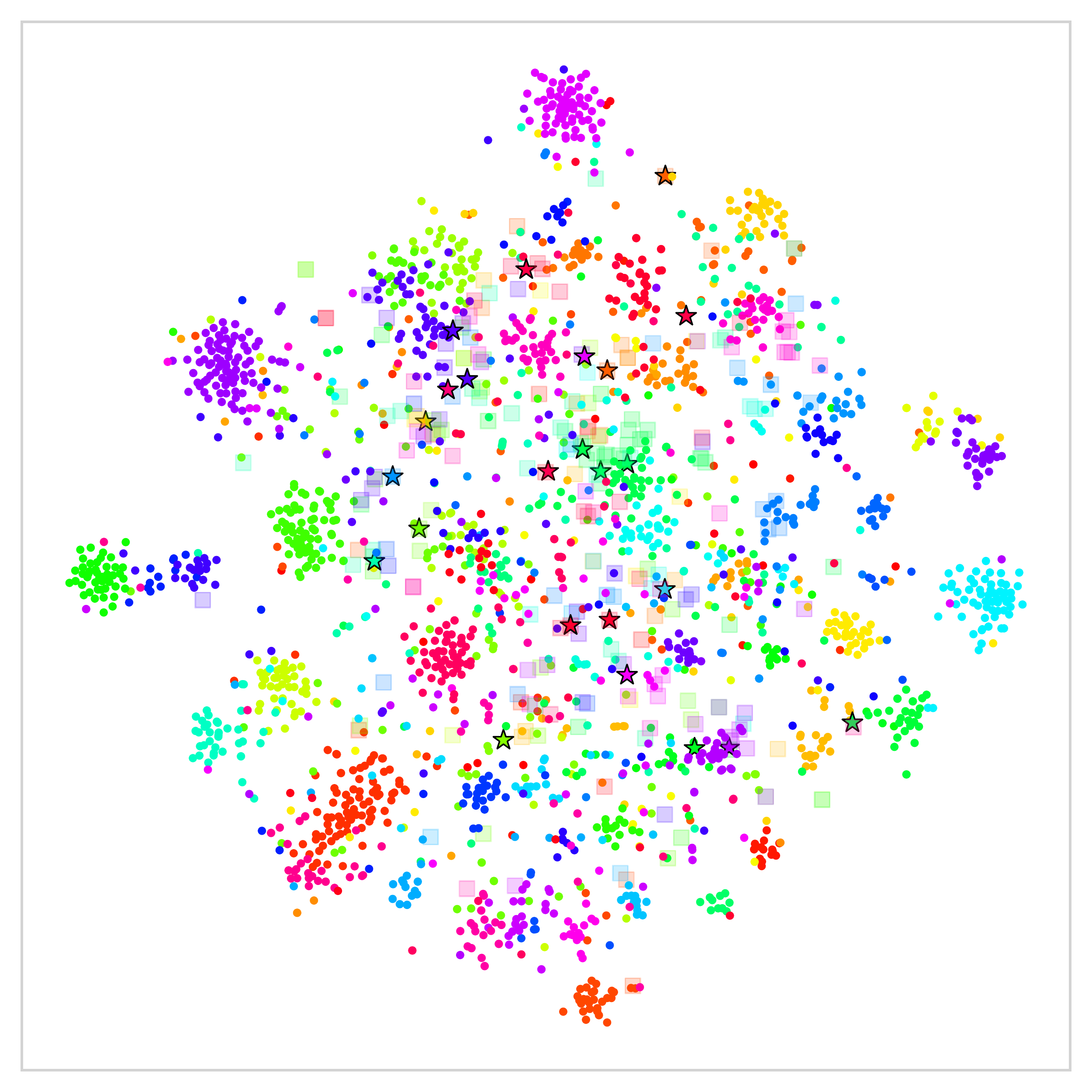}	
    \includegraphics[width=0.5\linewidth]{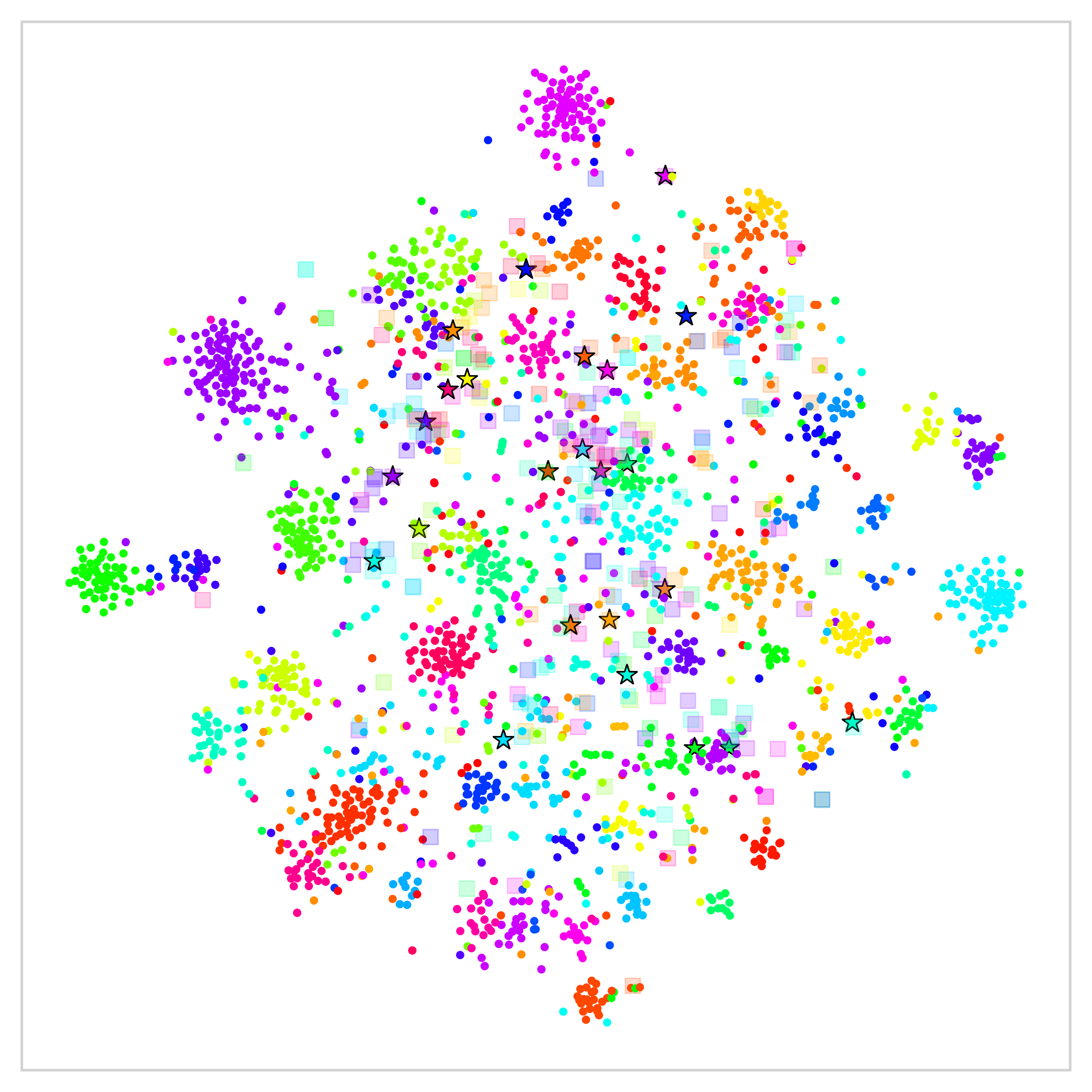}	
\end{minipage}

\begin{minipage}{\linewidth}
    \begin{minipage}{0.015\linewidth}
        \centering
        \rotatebox{90}{
            \footnotesize
            Candidates}
    \end{minipage} 
    \hfill   
    \begin{boxedminipage}{0.99\linewidth}
    \includegraphics[width=0.095\linewidth,height=0.1\linewidth]{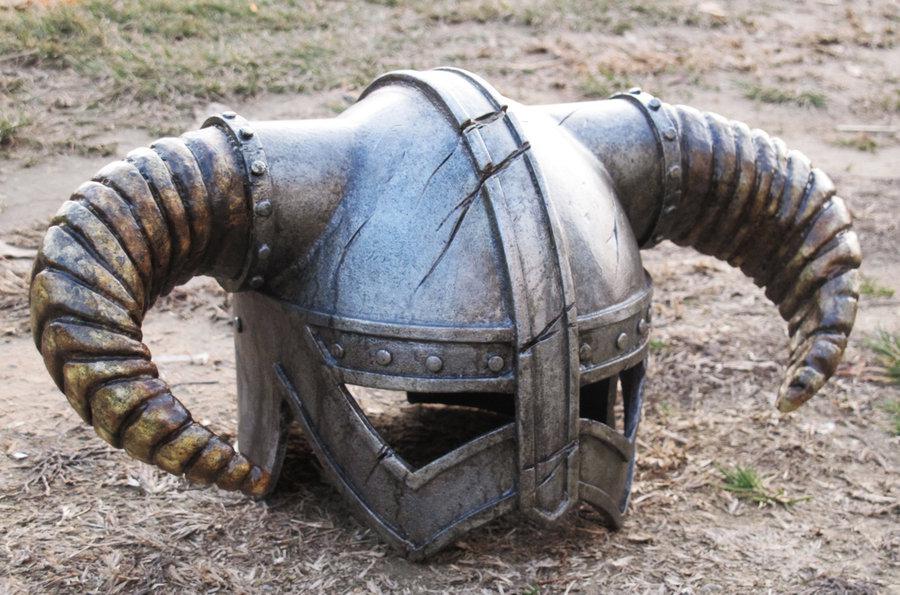}
    \includegraphics[width=0.095\linewidth,height=0.1\linewidth]{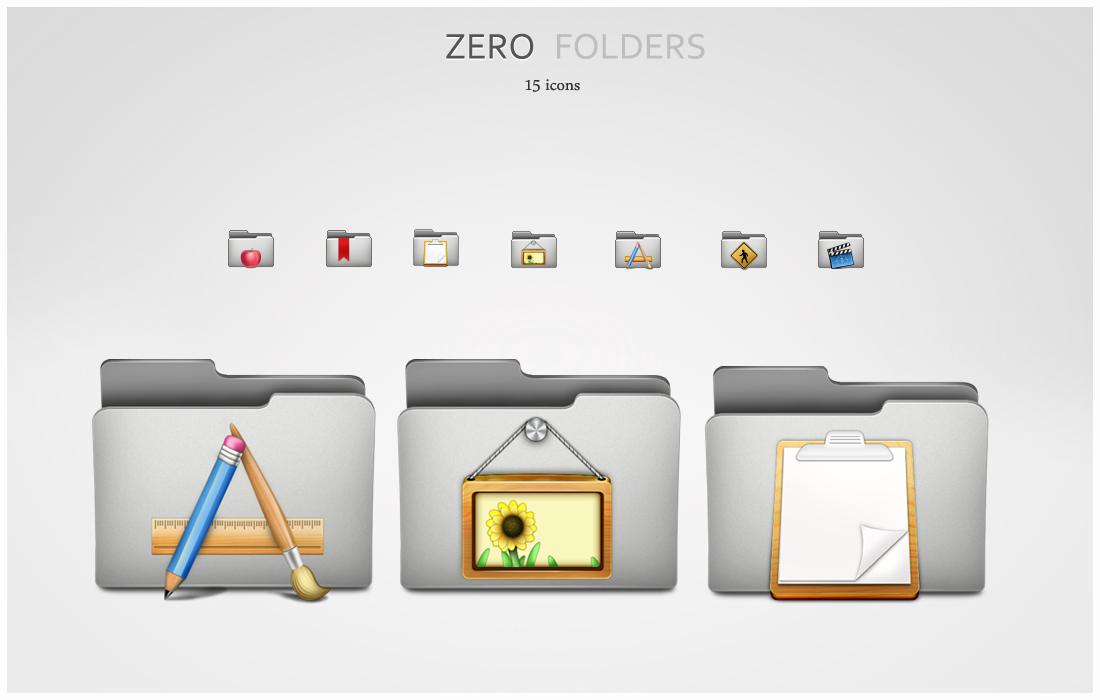}
    \includegraphics[width=0.095\linewidth,height=0.1\linewidth]{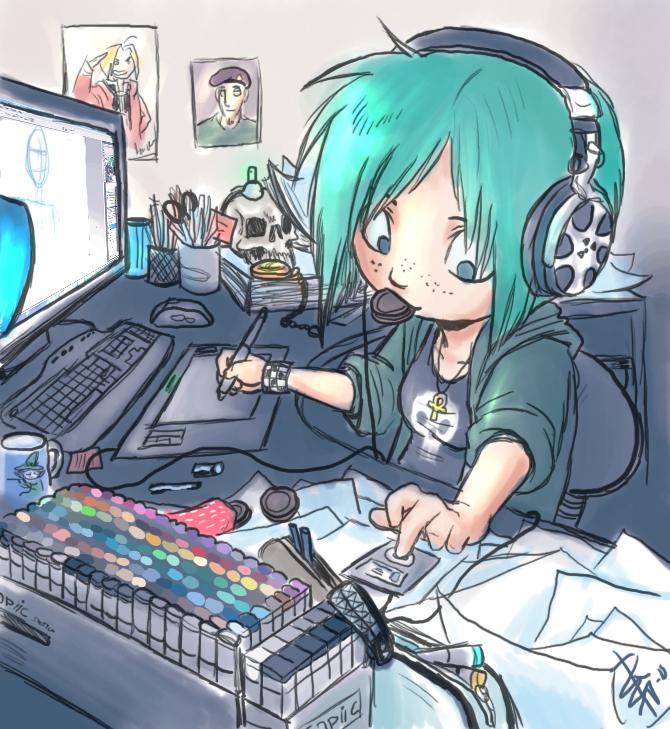}
    \includegraphics[width=0.095\linewidth,height=0.1\linewidth]{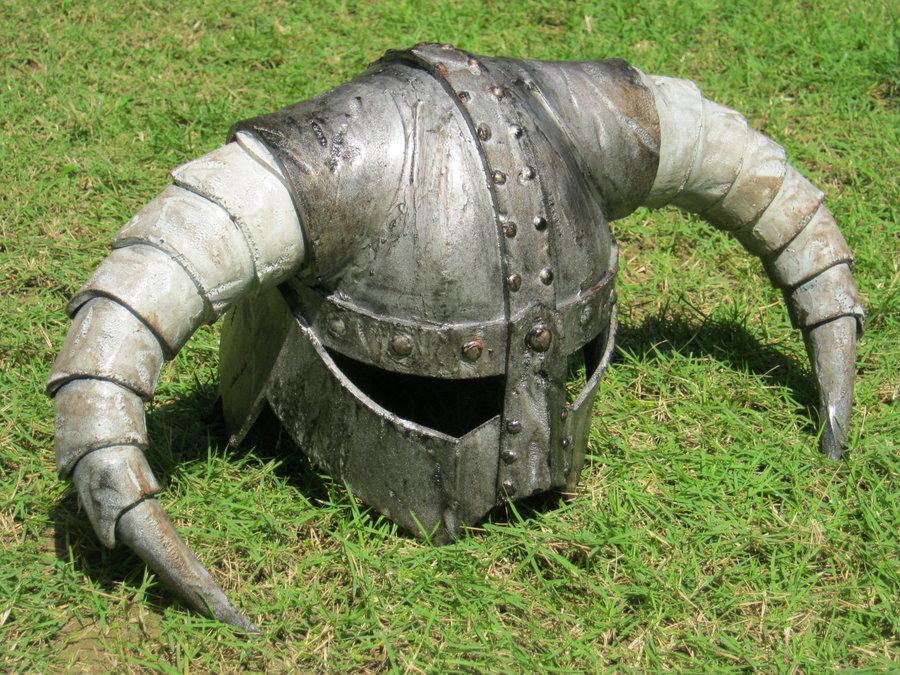}
    \includegraphics[width=0.095\linewidth,height=0.1\linewidth]{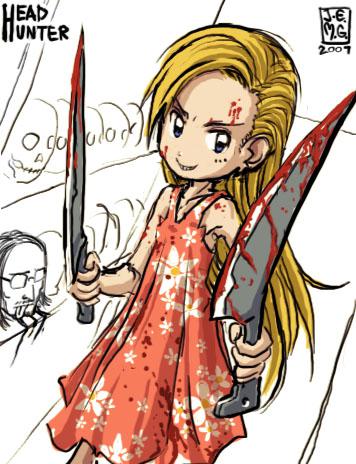}
    \includegraphics[width=0.095\linewidth,height=0.1\linewidth]{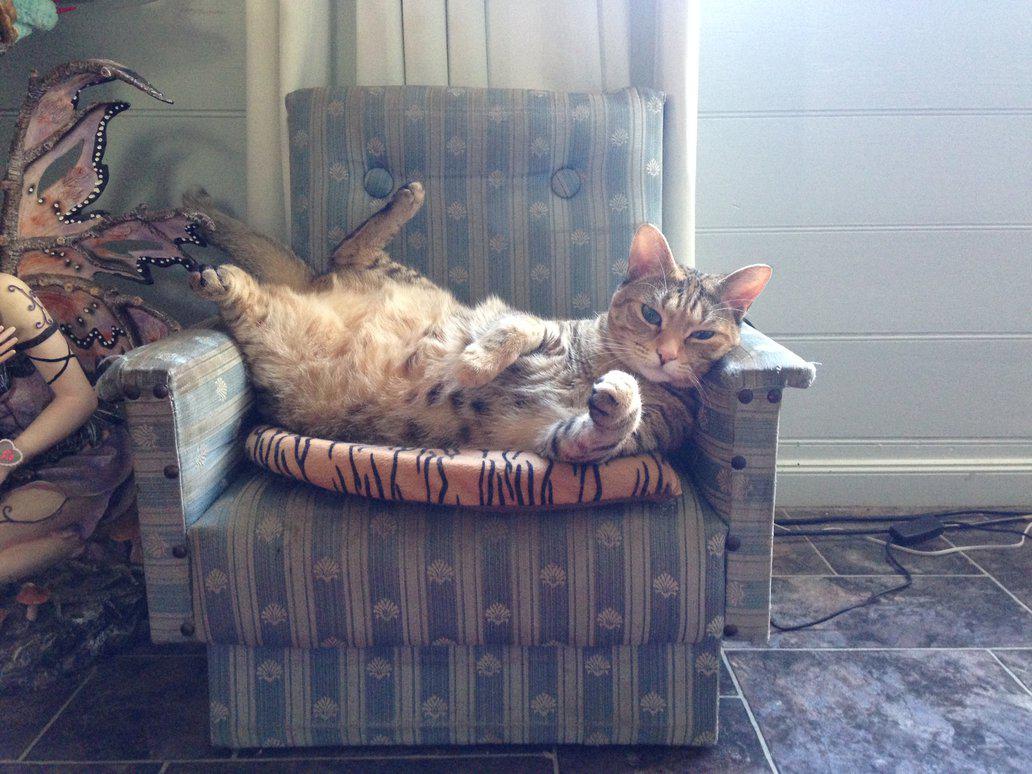}
    \includegraphics[width=0.095\linewidth,height=0.1\linewidth]{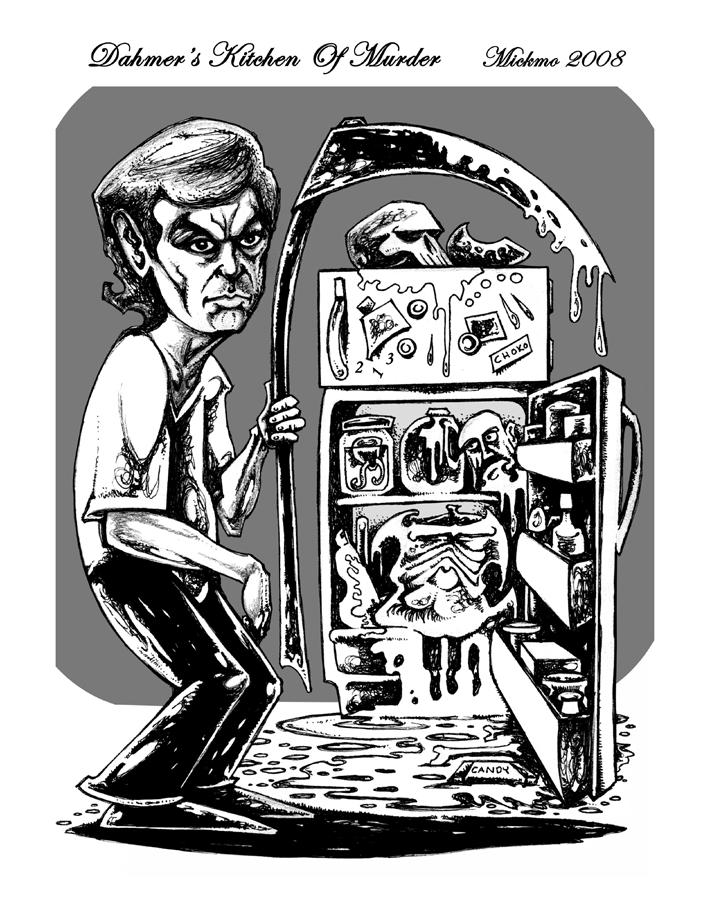}
    \includegraphics[width=0.095\linewidth,height=0.1\linewidth]{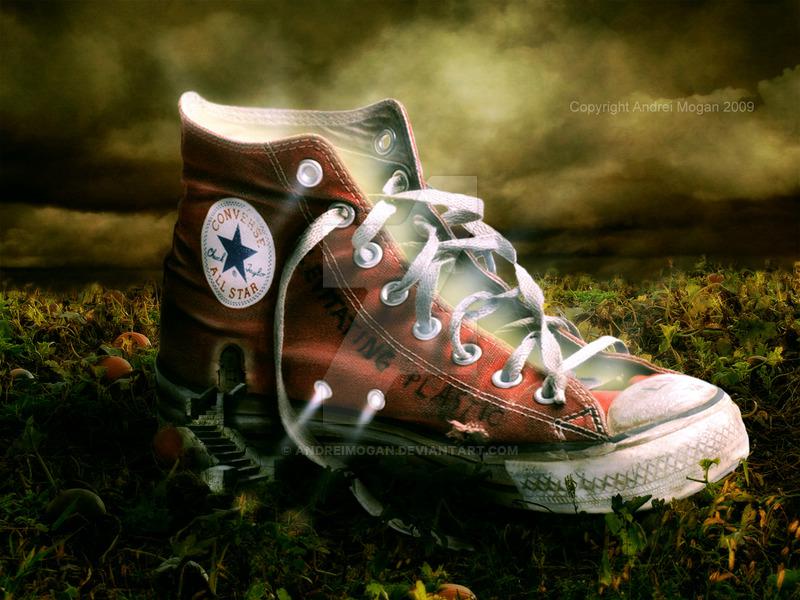}
    \includegraphics[width=0.095\linewidth,height=0.1\linewidth]{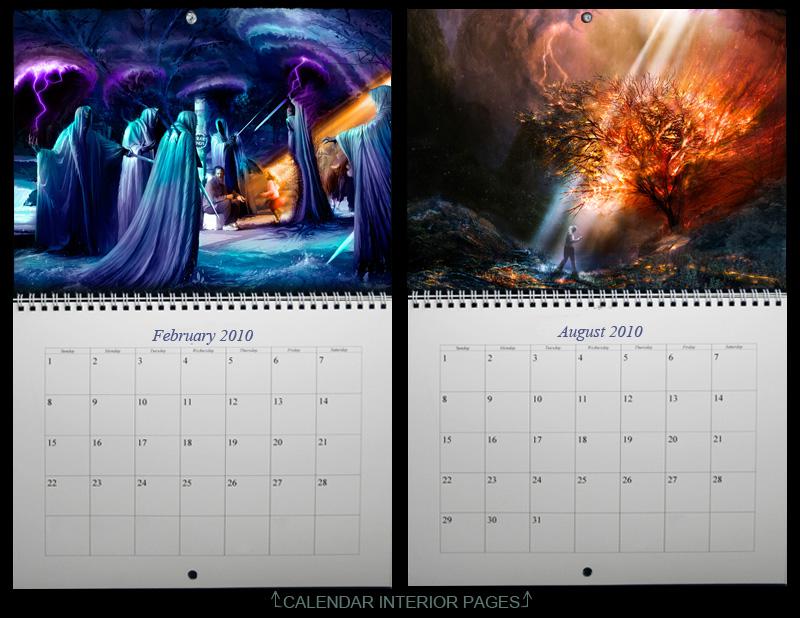}
    \includegraphics[width=0.095\linewidth,height=0.1\linewidth]{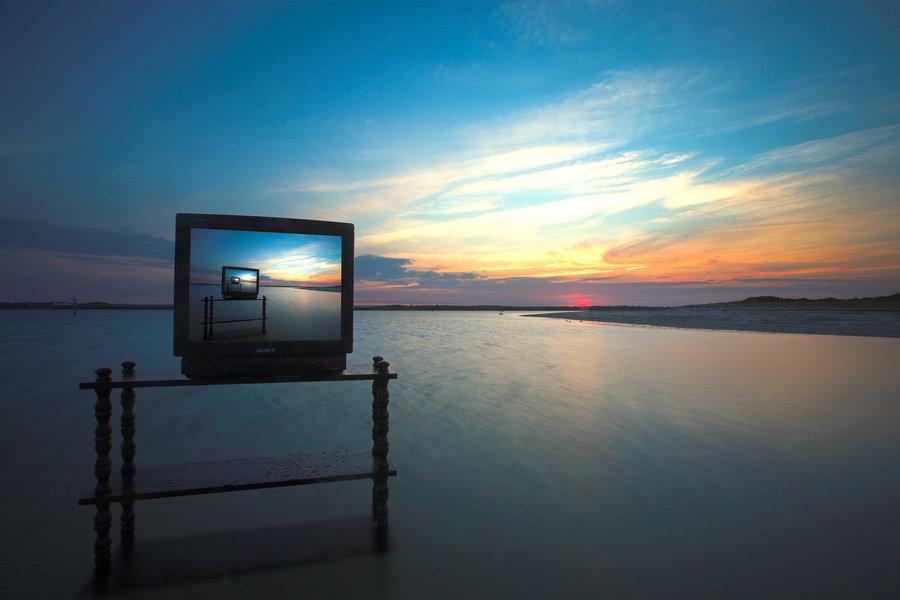}	
    
    \includegraphics[width=0.095\linewidth,height=0.1\linewidth]{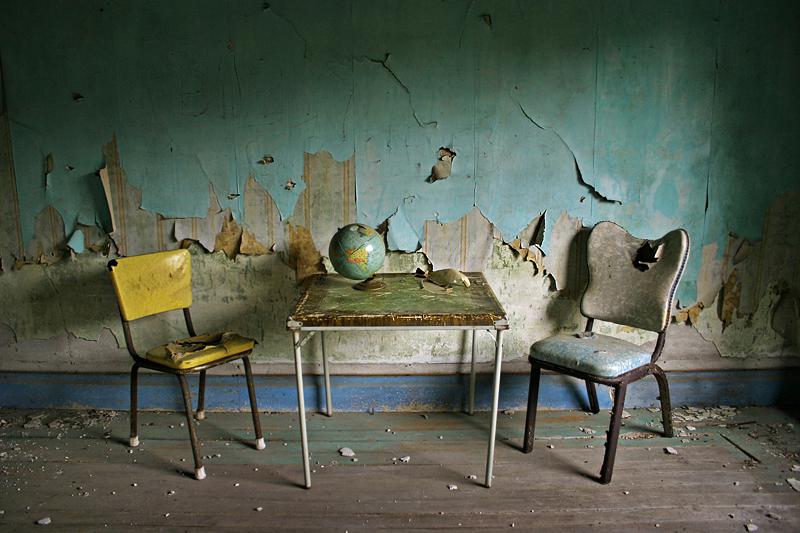}
    \includegraphics[width=0.095\linewidth,height=0.1\linewidth]{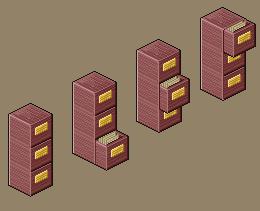}
    \includegraphics[width=0.095\linewidth,height=0.1\linewidth]{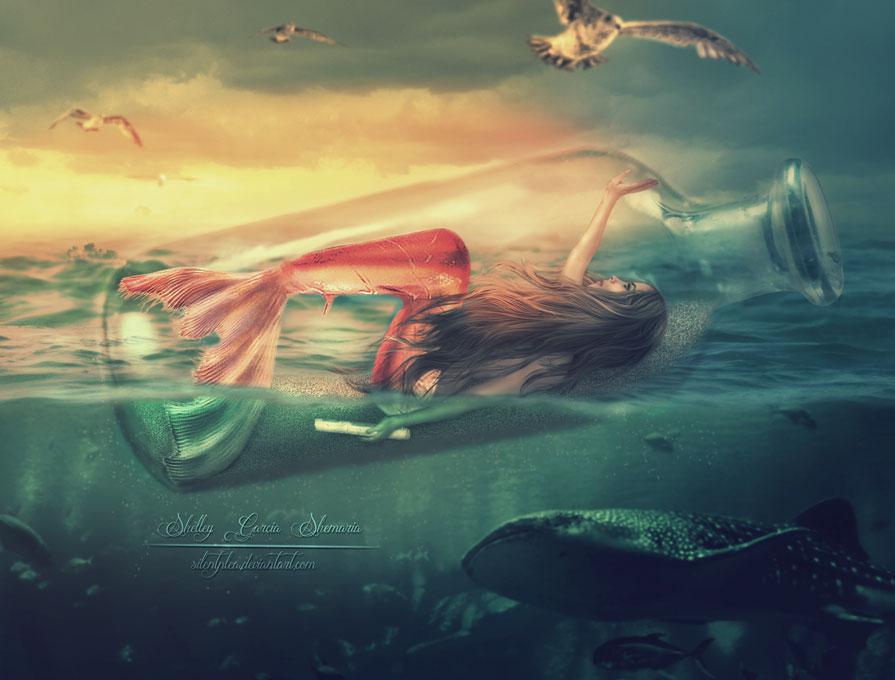}
    \includegraphics[width=0.095\linewidth,height=0.1\linewidth]{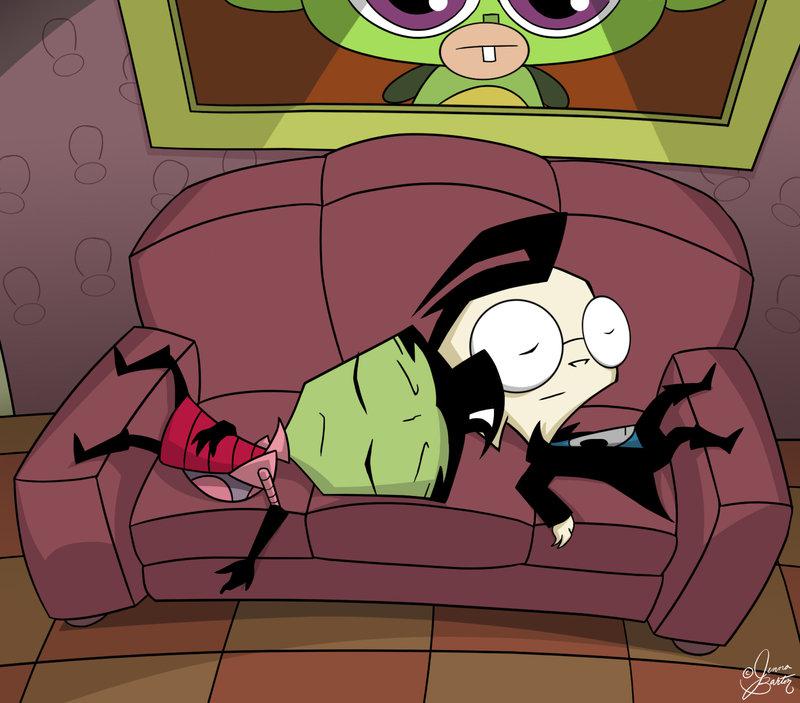}
    \includegraphics[width=0.095\linewidth,height=0.1\linewidth]{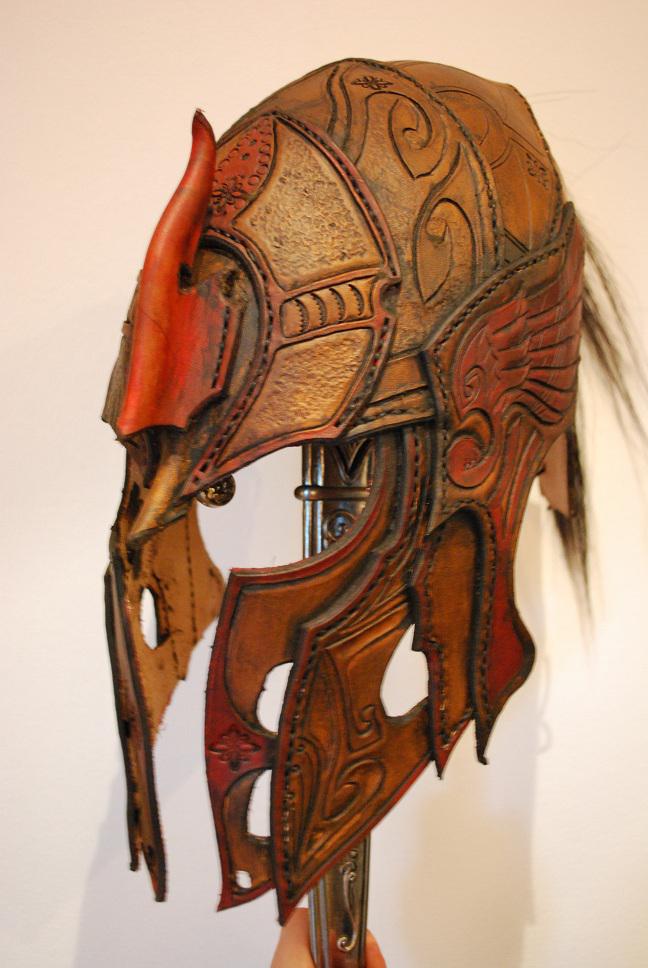}
    \includegraphics[width=0.095\linewidth,height=0.1\linewidth]{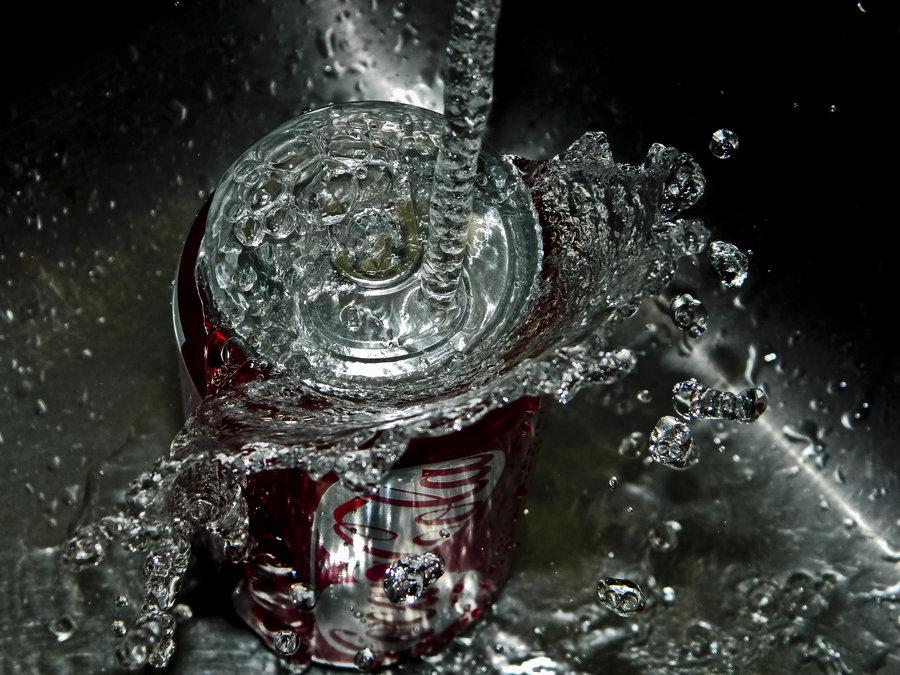}
    \includegraphics[width=0.095\linewidth,height=0.1\linewidth]{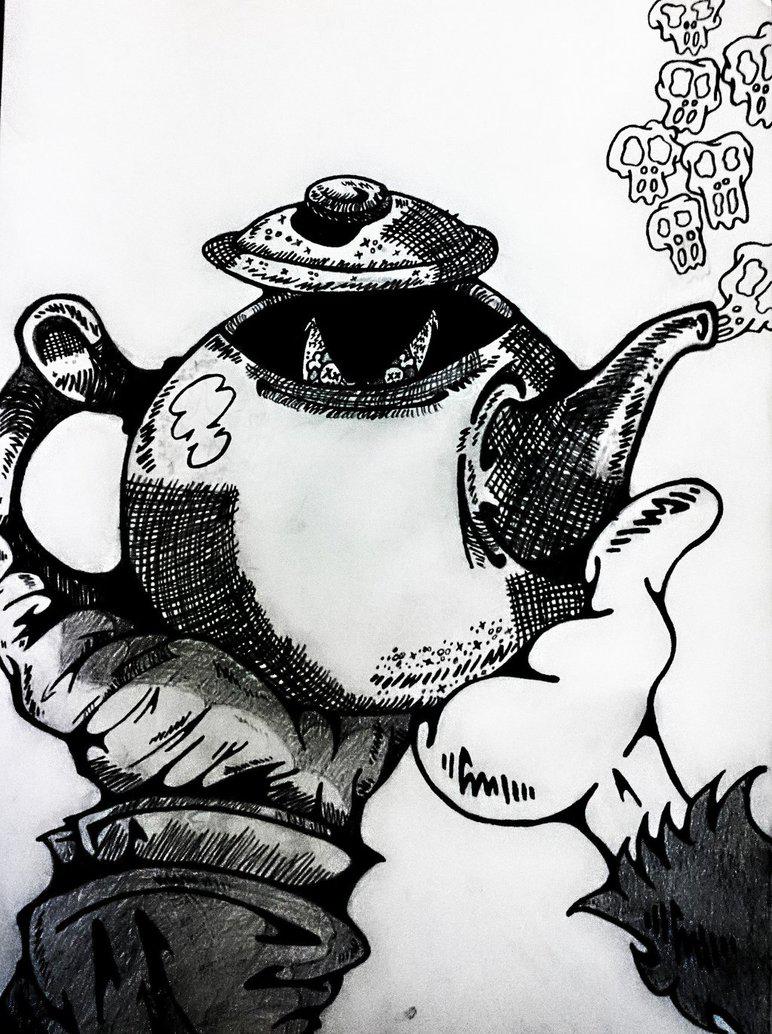}
    \includegraphics[width=0.095\linewidth,height=0.1\linewidth]{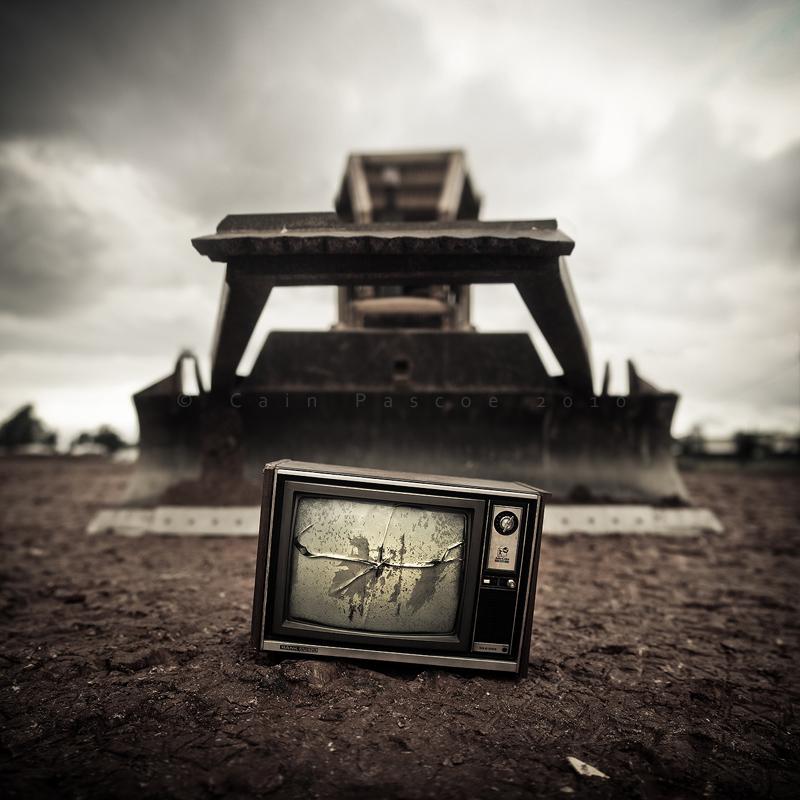}
    \includegraphics[width=0.095\linewidth,height=0.1\linewidth]{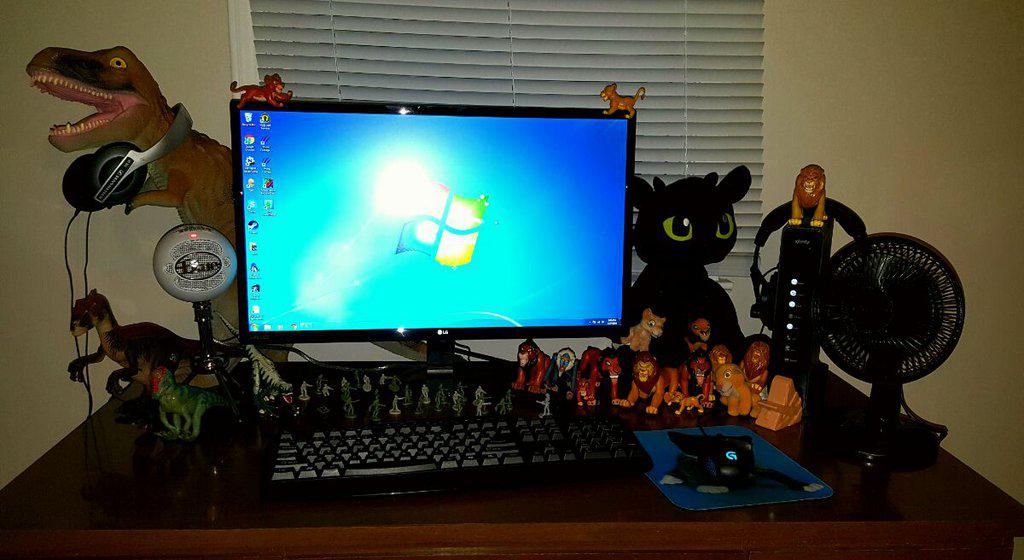}
    \includegraphics[width=0.095\linewidth,height=0.1\linewidth]{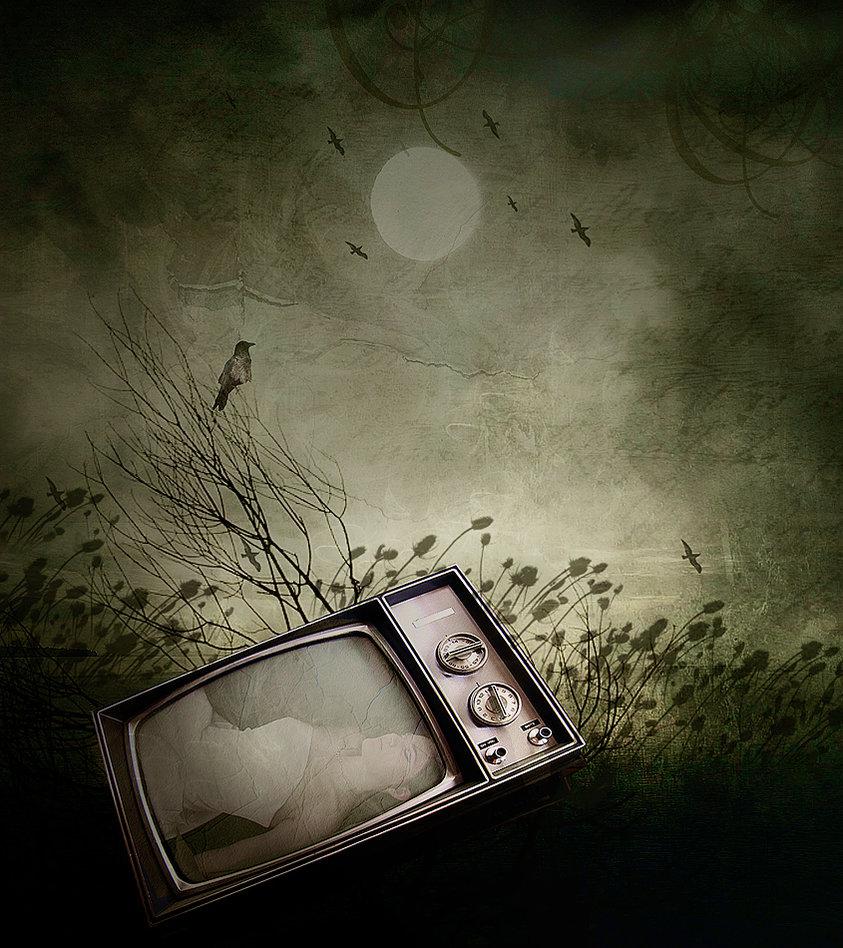}	
    \end{boxedminipage}
\end{minipage}

\begin{minipage}{\linewidth}
    \begin{minipage}{0.015\linewidth}
        \centering
        \rotatebox{90}{
            \footnotesize
            Queried}
    \end{minipage} 
    \hfill   
    \begin{boxedminipage}{0.99\linewidth}
        \includegraphics[width=0.095\linewidth,height=0.1\linewidth]{fig/Art/Table_00015.jpg}
        \includegraphics[width=0.095\linewidth,height=0.1\linewidth]{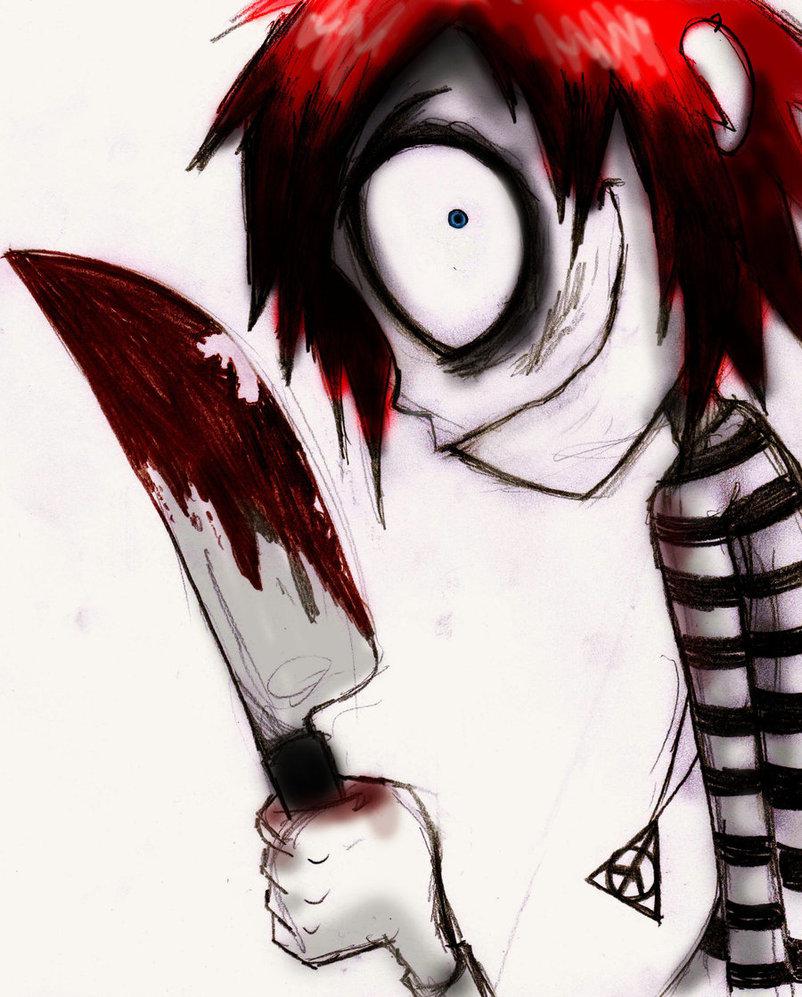}
        \includegraphics[width=0.095\linewidth,height=0.1\linewidth]{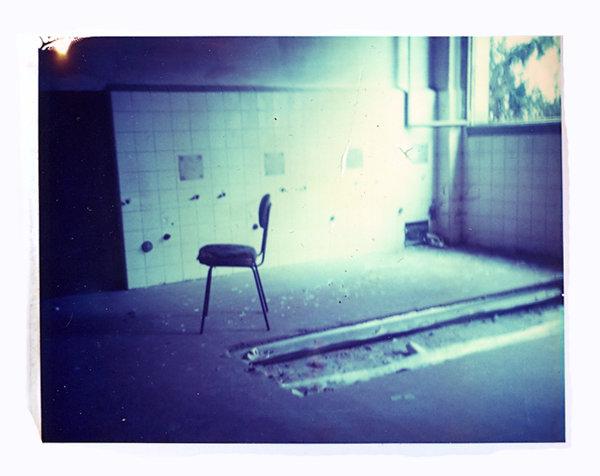}
        \includegraphics[width=0.095\linewidth,height=0.1\linewidth]{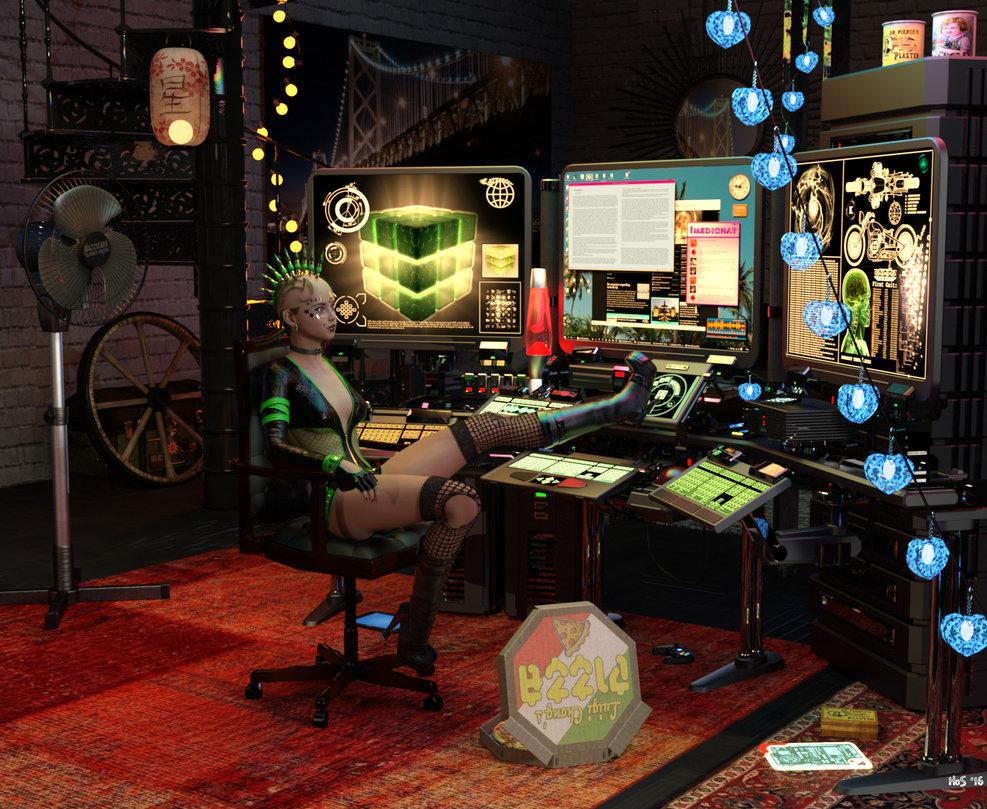}
        \includegraphics[width=0.095\linewidth,height=0.1\linewidth]{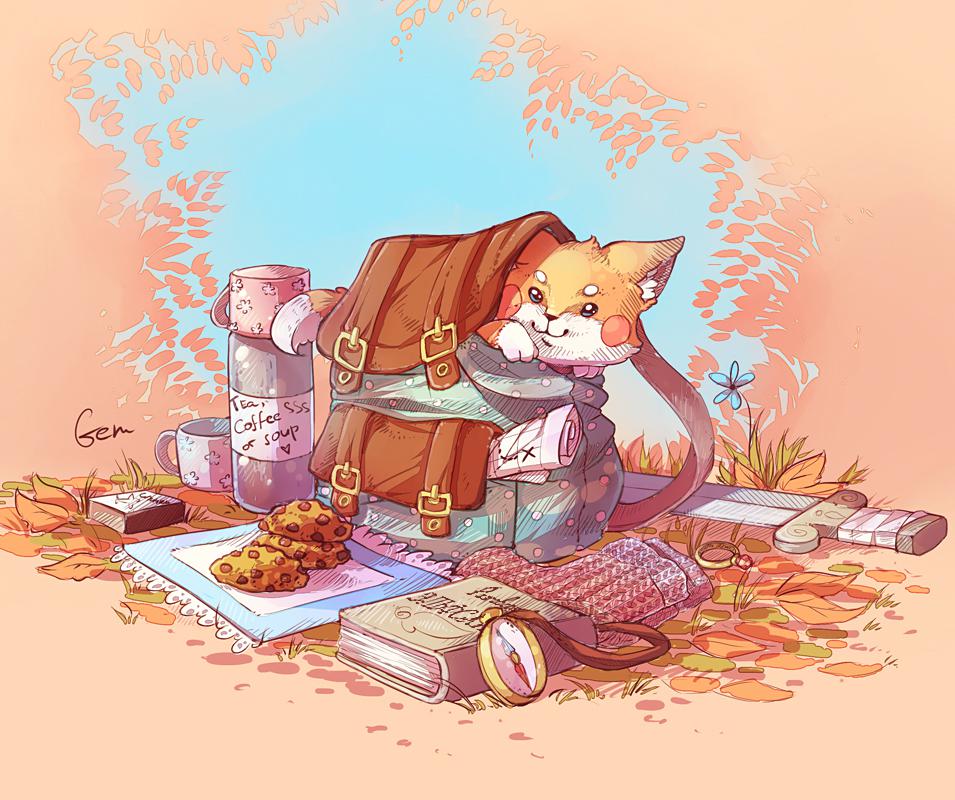}
        \includegraphics[width=0.095\linewidth,height=0.1\linewidth]{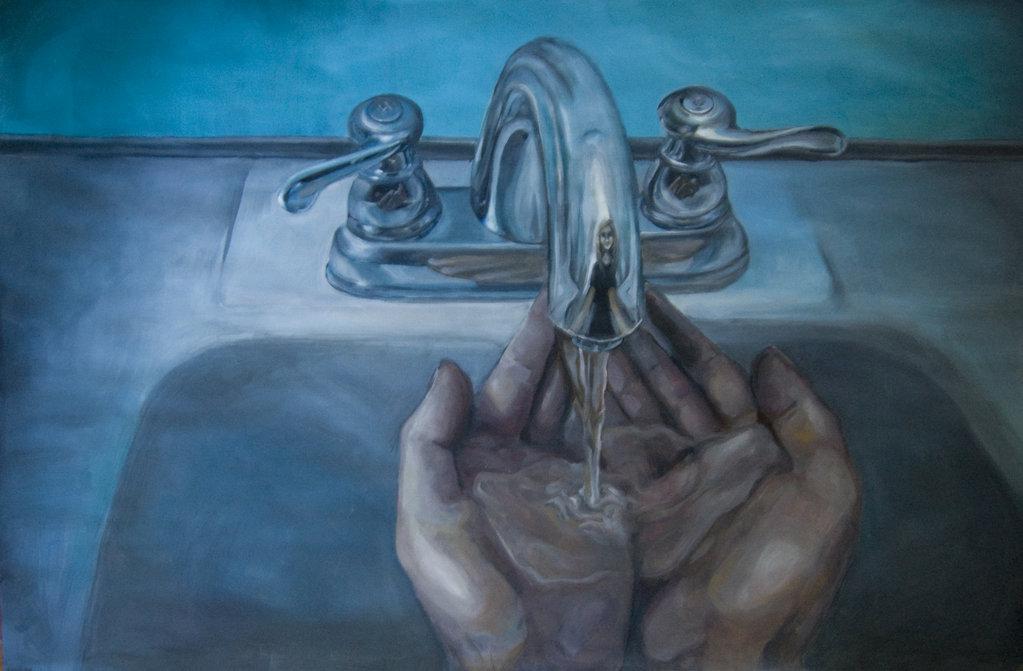}
        \includegraphics[width=0.095\linewidth,height=0.1\linewidth]{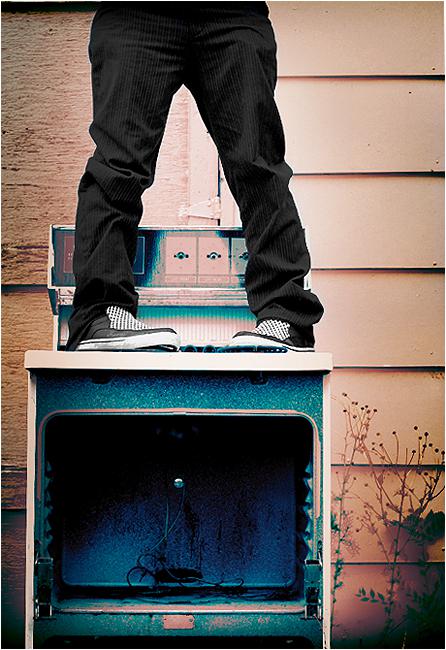}
        \includegraphics[width=0.095\linewidth,height=0.1\linewidth]{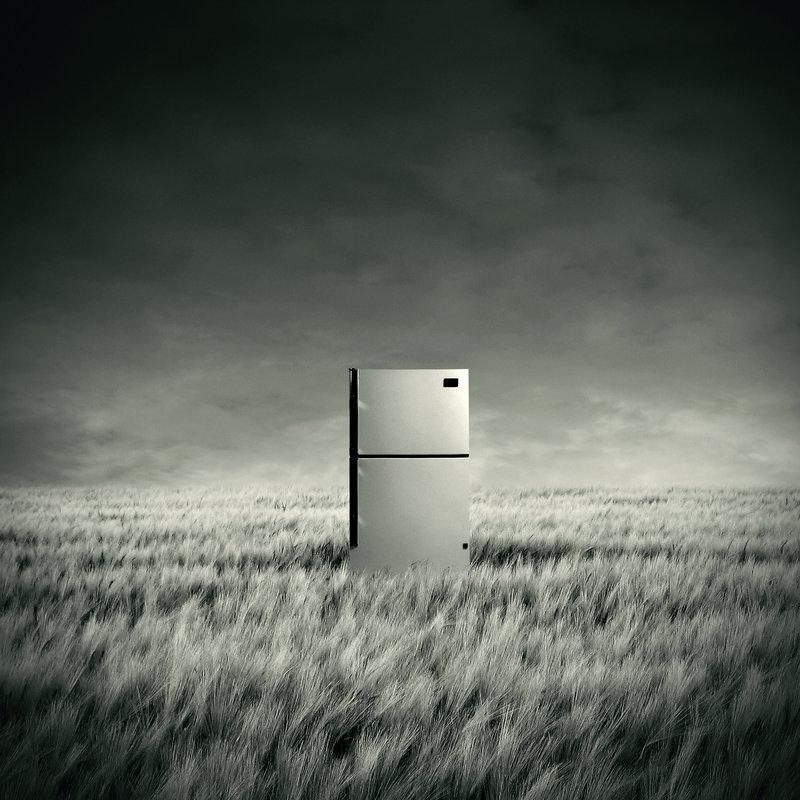}
        \includegraphics[width=0.095\linewidth,height=0.1\linewidth]{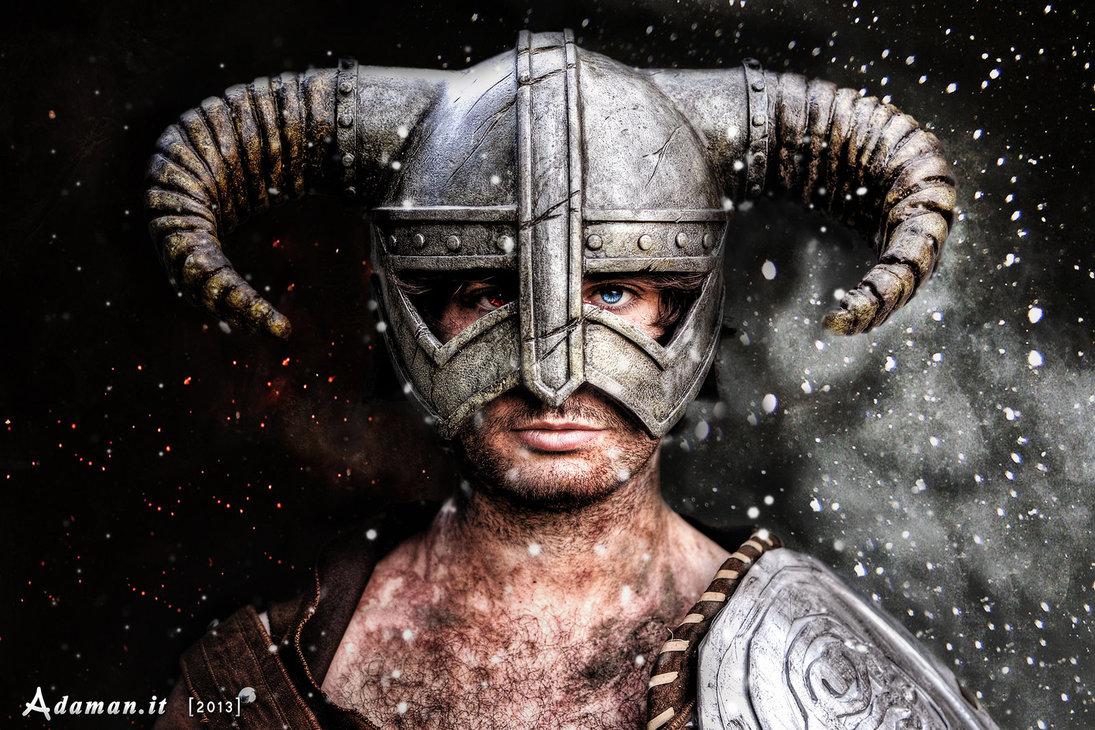}
        \includegraphics[width=0.095\linewidth,height=0.1\linewidth]{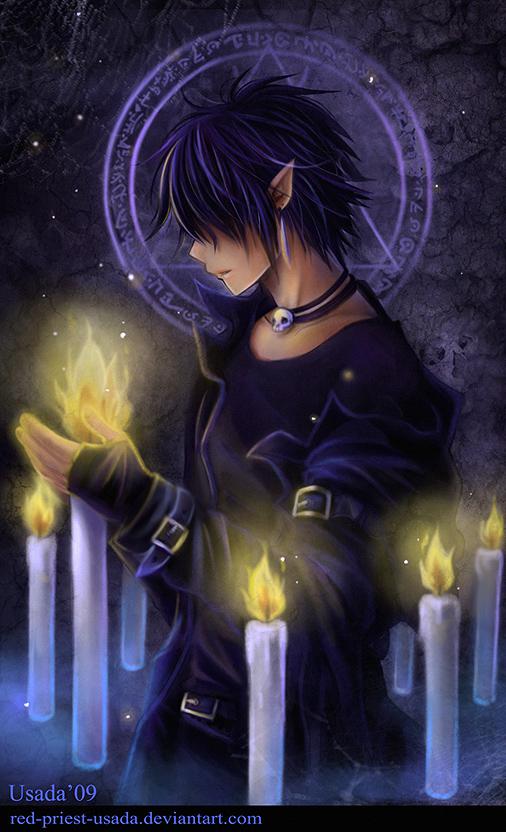}
        
        \includegraphics[width=0.095\linewidth,height=0.1\linewidth]{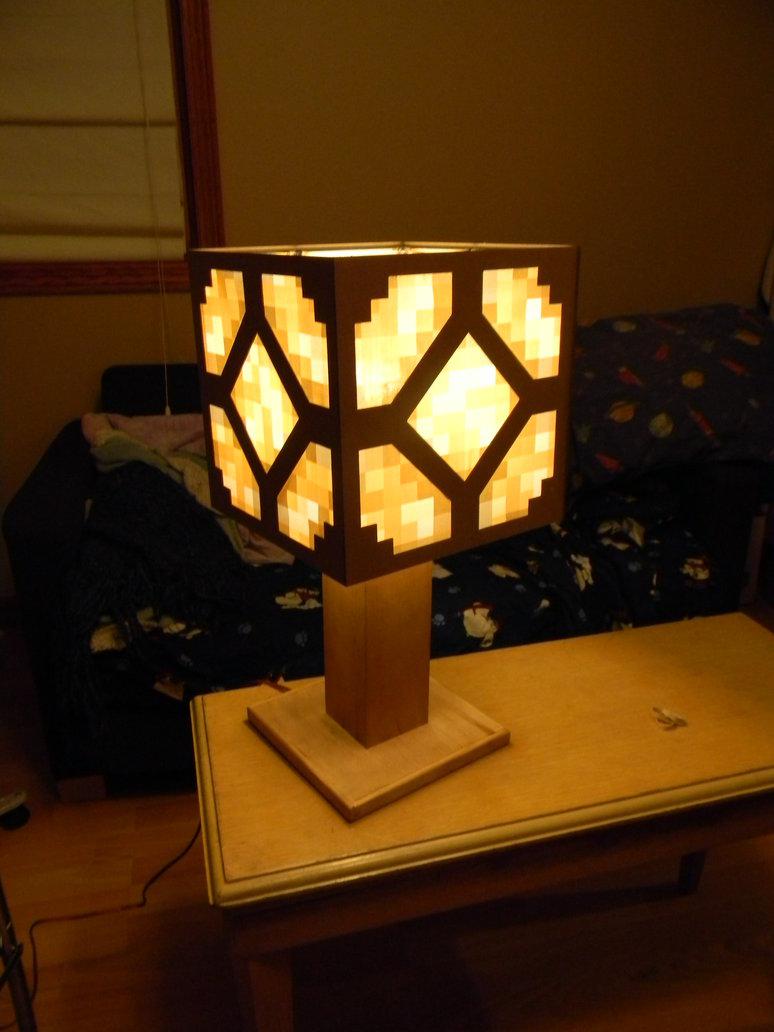}
        \includegraphics[width=0.095\linewidth,height=0.1\linewidth]{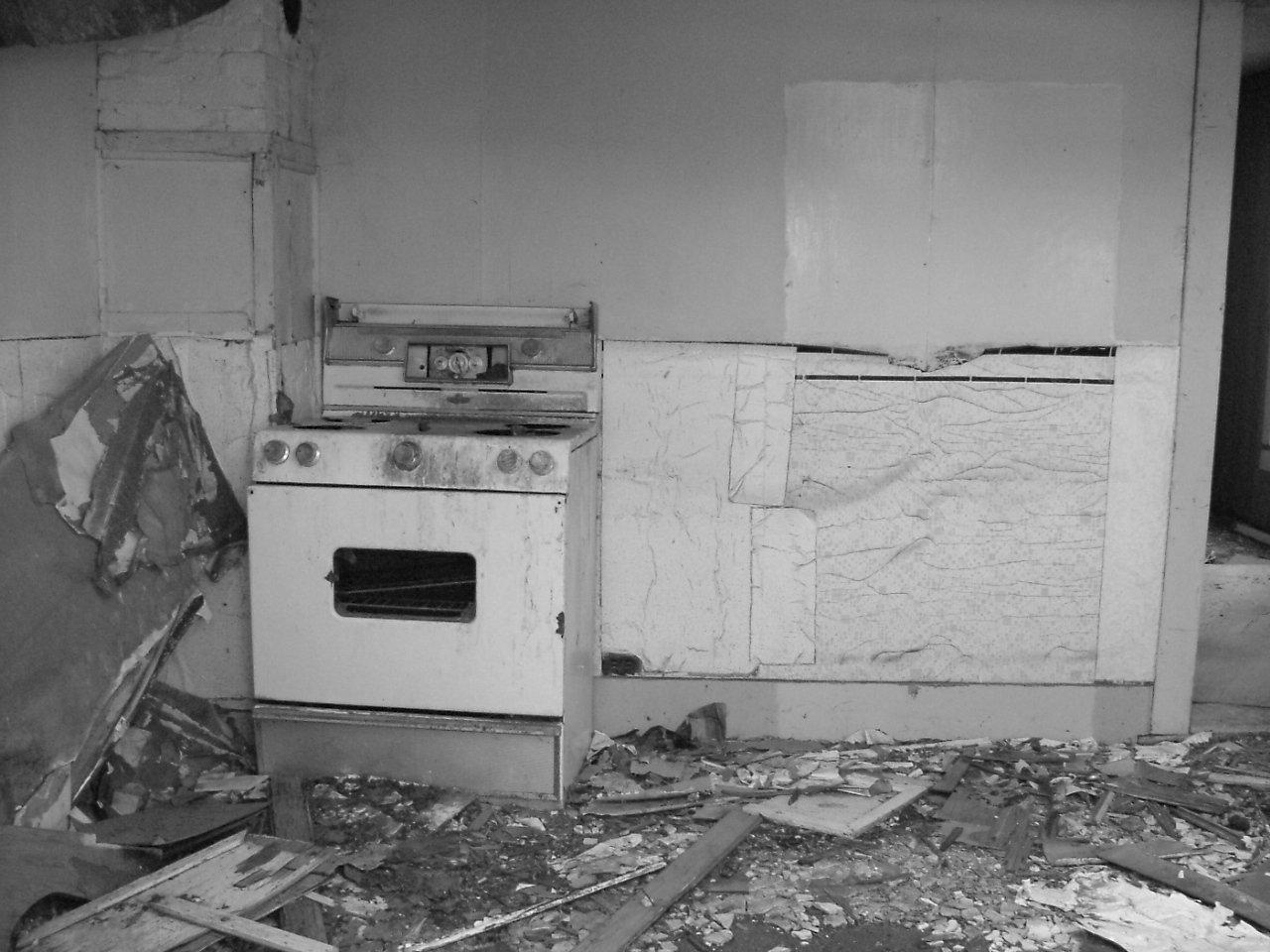}
        \includegraphics[width=0.095\linewidth,height=0.1\linewidth]{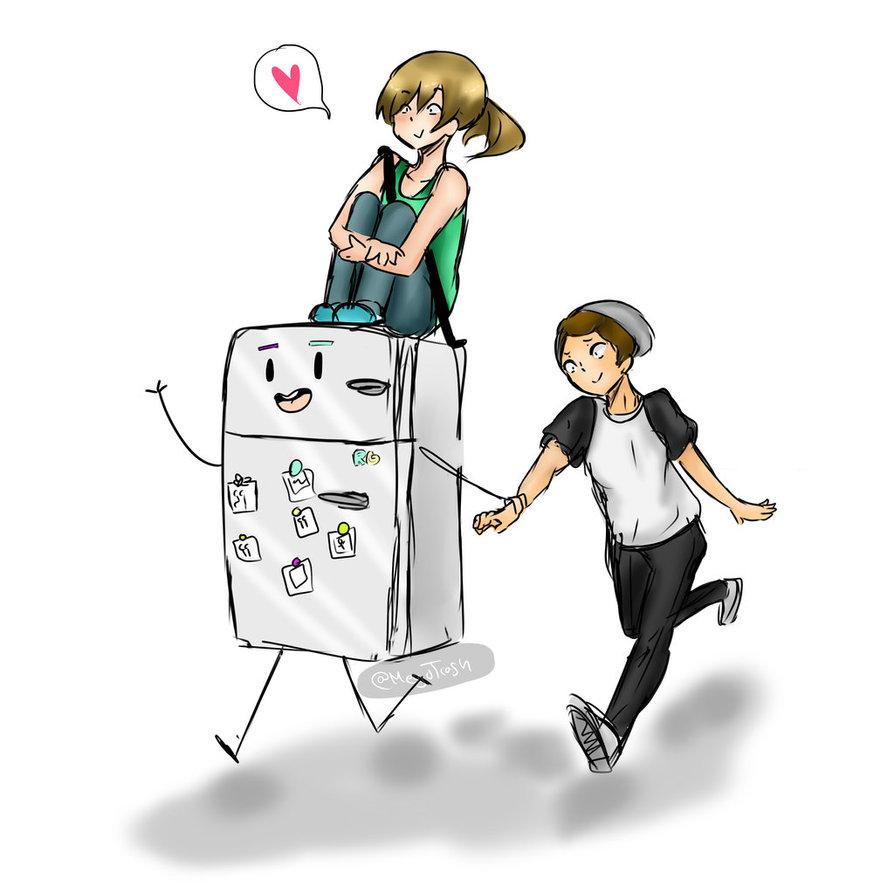}
        \includegraphics[width=0.095\linewidth,height=0.1\linewidth]{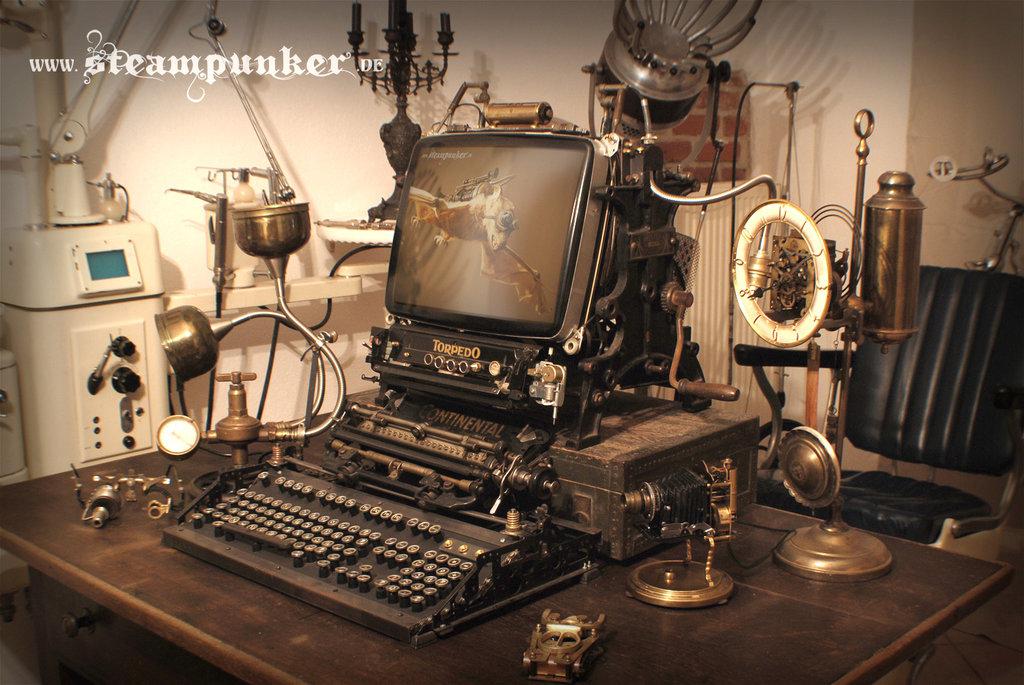}
        \includegraphics[width=0.095\linewidth,height=0.1\linewidth]{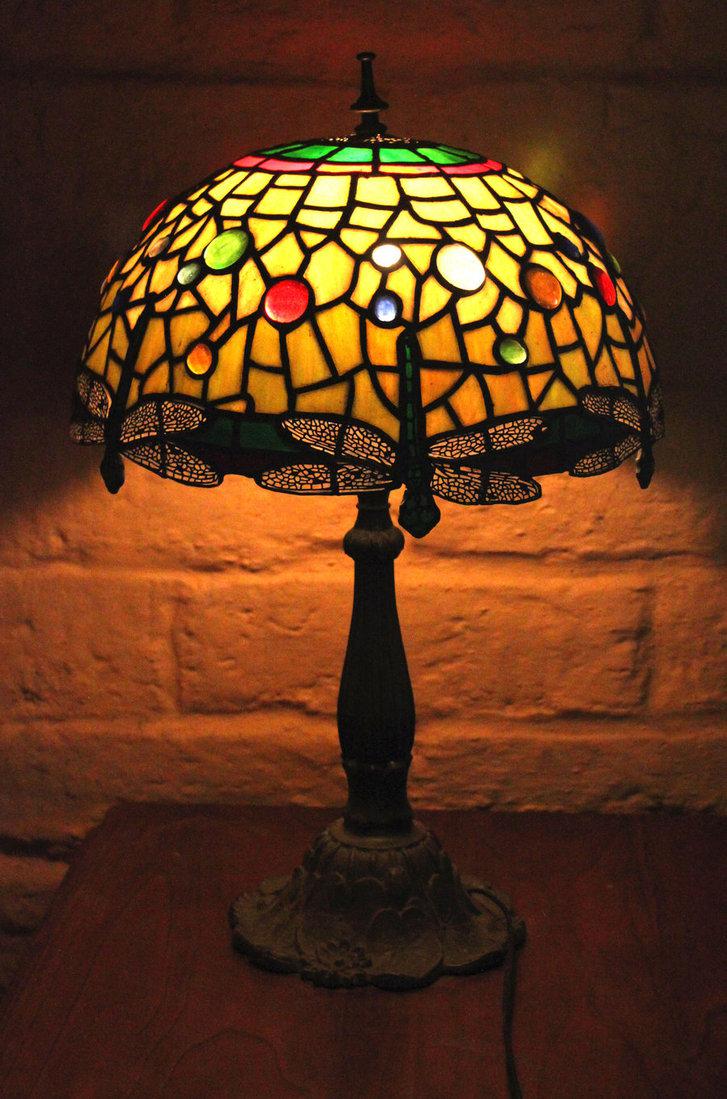}
        \includegraphics[width=0.095\linewidth,height=0.1\linewidth]{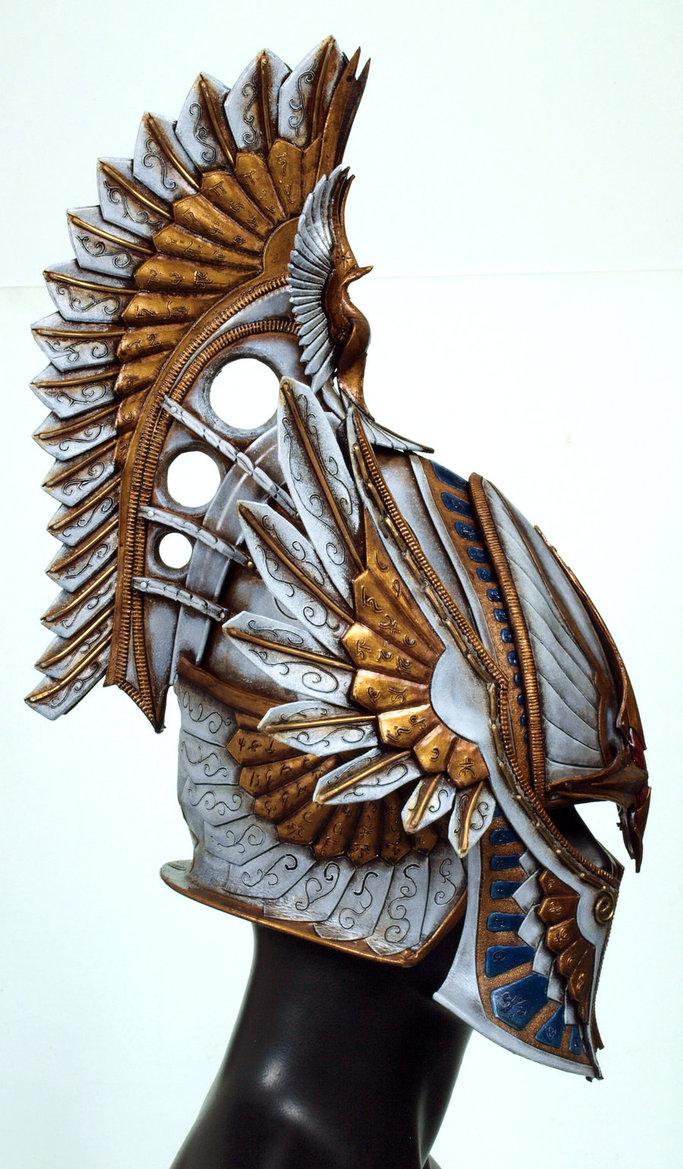}
        \includegraphics[width=0.095\linewidth,height=0.1\linewidth]{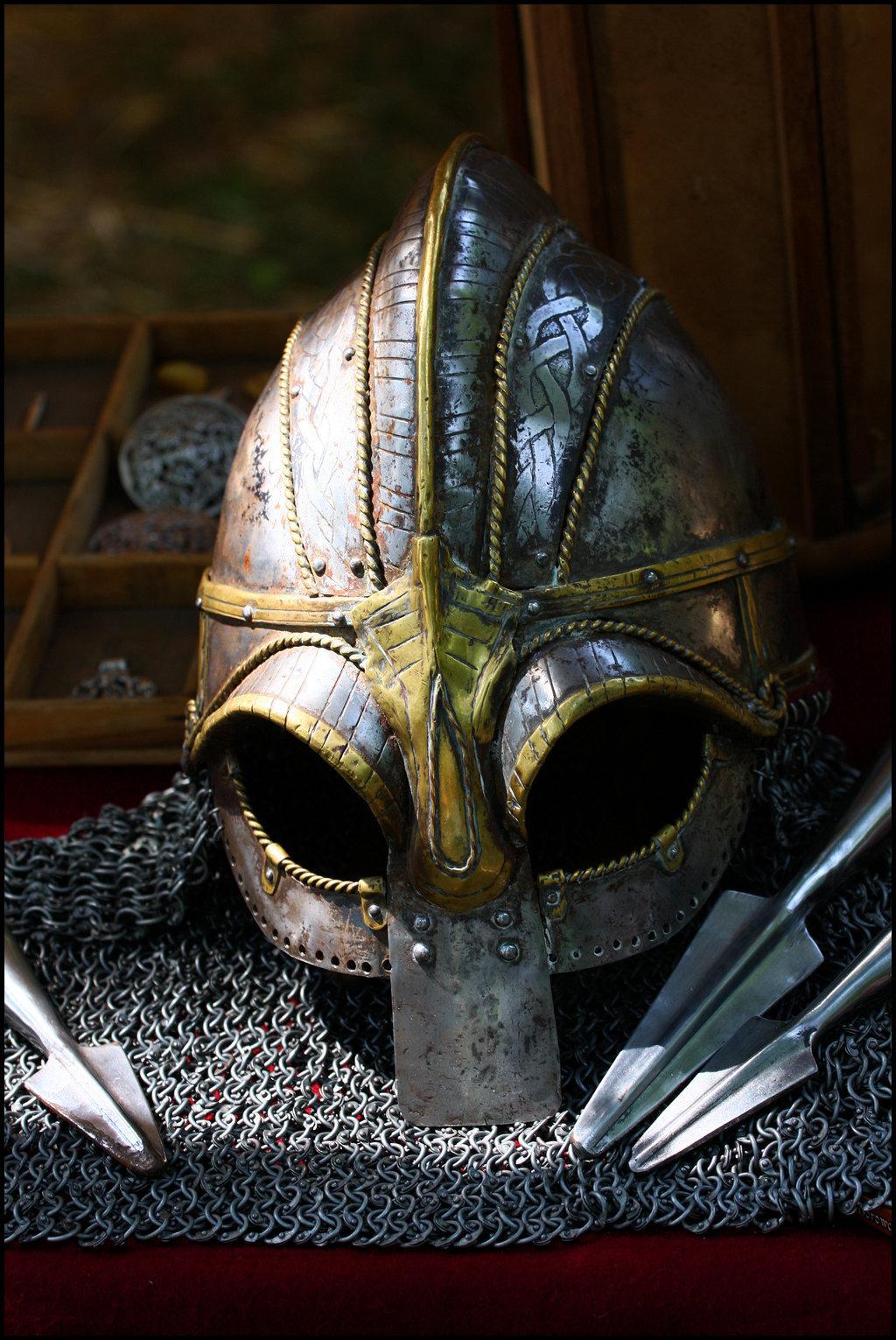}
        \includegraphics[width=0.095\linewidth,height=0.1\linewidth]{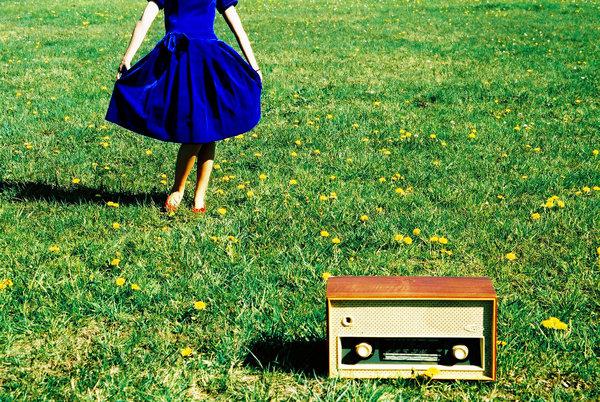}
        \includegraphics[width=0.095\linewidth,height=0.1\linewidth]{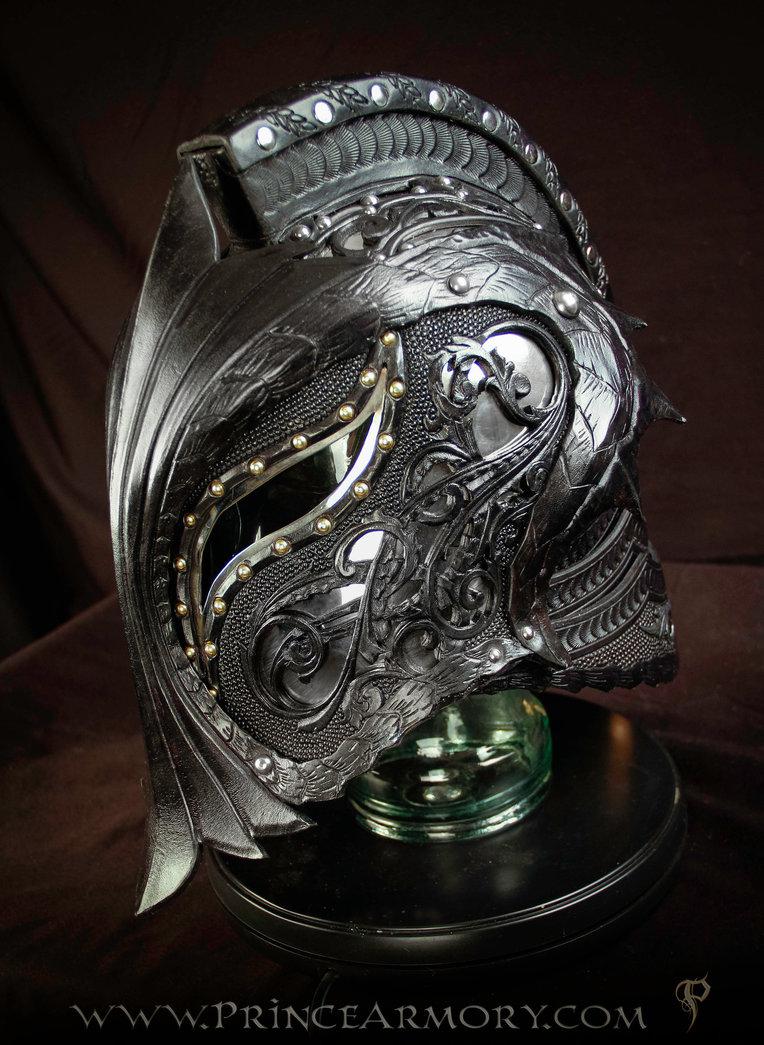}
        \includegraphics[width=0.095\linewidth,height=0.1\linewidth]{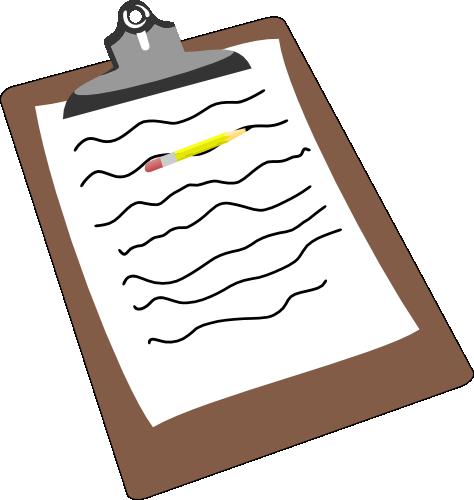}
    \end{boxedminipage}
\end{minipage}

\caption{Visualization of LAS sampling on Office-Home Pr$\rightarrow$Ar. The first row presents $t$-SNE plots. Squares denote candidate target samples based on LI-scores; stars denote selected target samples for querying labels; and points denote the rest target samples. Each marker is colored according to its (left) ground-truth label and (right) pseudo label from the current model. The last two rows plot top 20 candidate samples and queried samples with largest LI-scores, respectively.}
\label{fig:sup:pr2ar}
 \vspace{40mm}
\end{figure*}

\begin{figure*}[!t]
	\centering
	\begin{minipage}{0.99\linewidth}
	\hspace{1.5mm}	
	\includegraphics[width=0.5\linewidth]{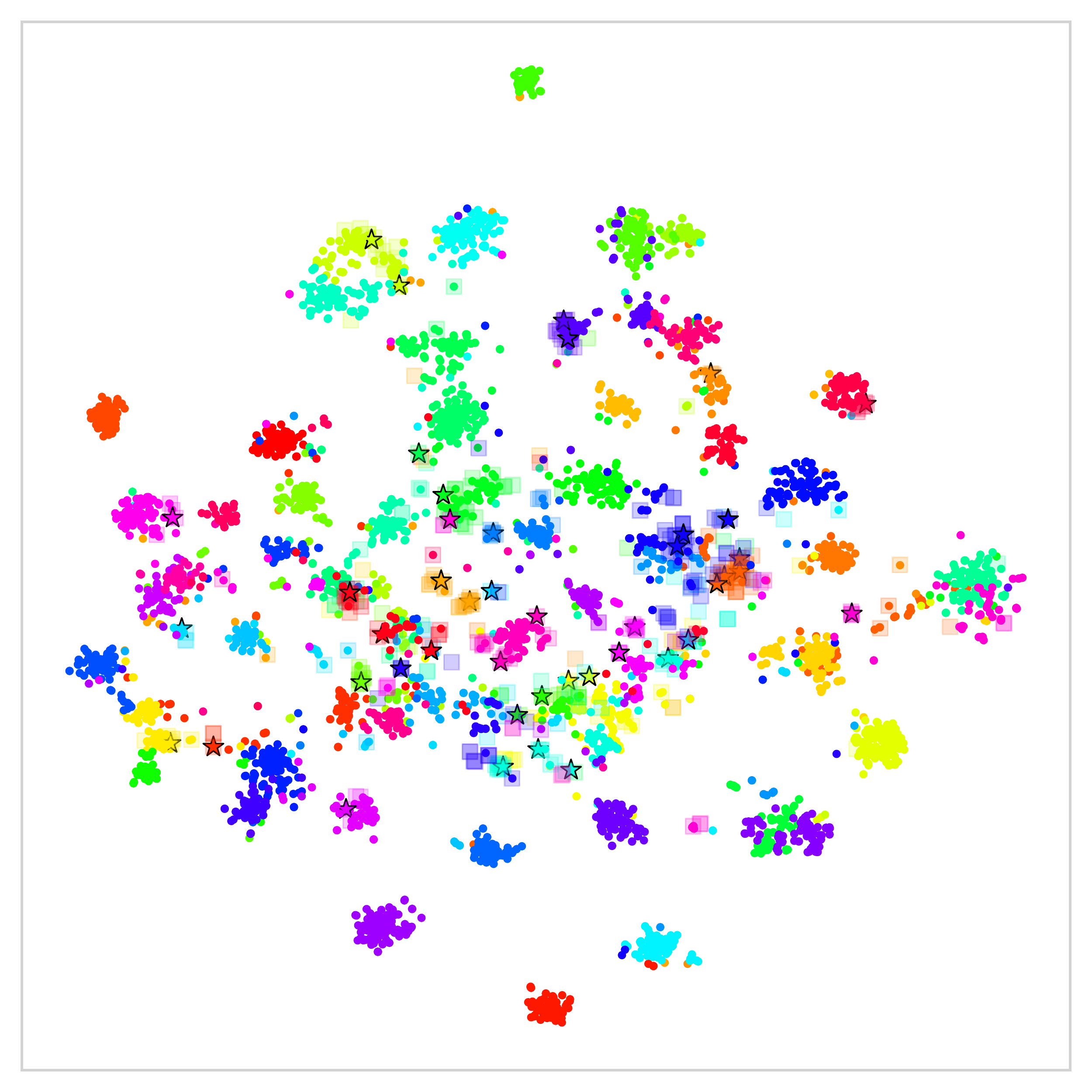}	
	\includegraphics[width=0.5\linewidth]{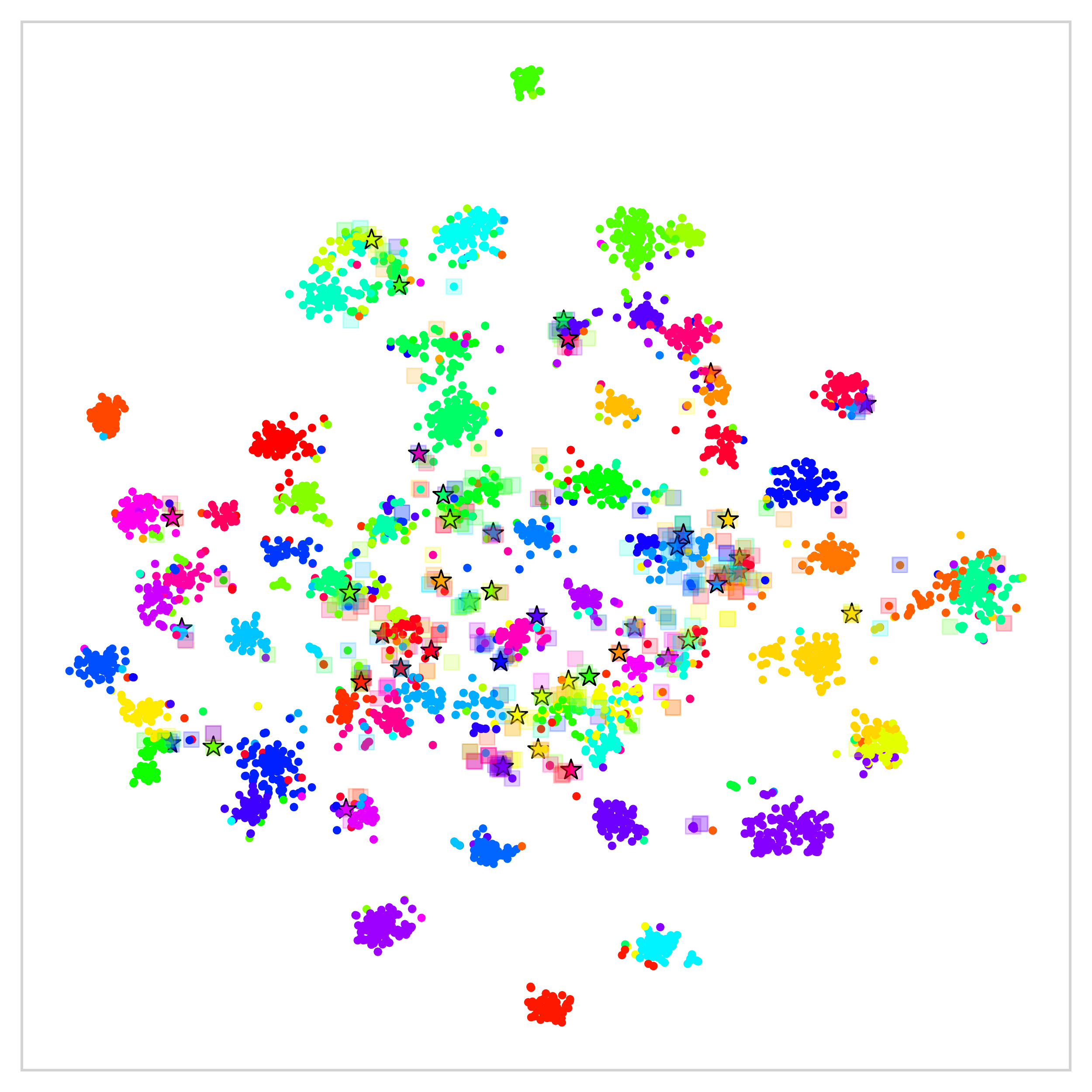}	
    \end{minipage} 
	
	\begin{minipage}{\linewidth}
		\begin{minipage}{0.015\linewidth}
			\centering
			\rotatebox{90}{
				\footnotesize
				Candidates}
		\end{minipage} 
		\hfill   
		\begin{boxedminipage}{0.99\linewidth}
			\includegraphics[width=0.095\linewidth,height=0.1\linewidth]{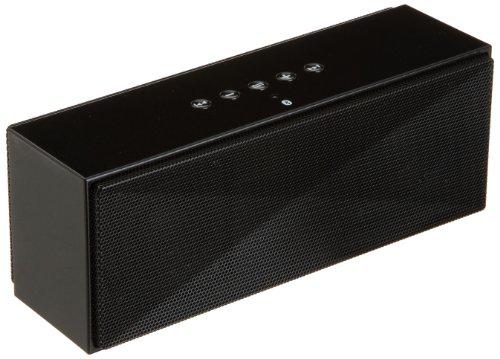}
			\includegraphics[width=0.095\linewidth,height=0.1\linewidth]{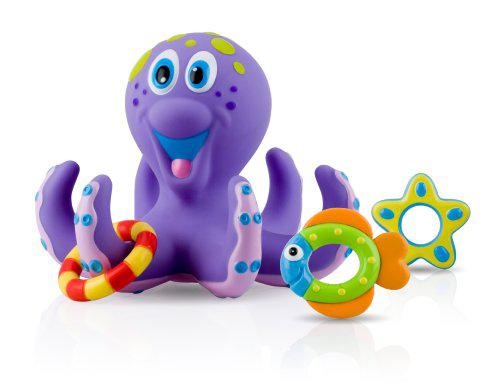}
			\includegraphics[width=0.095\linewidth,height=0.1\linewidth]{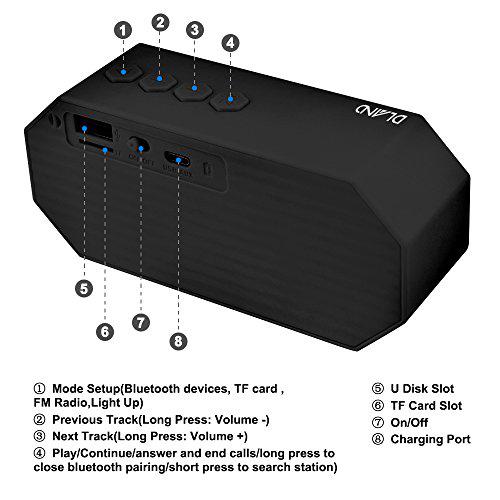}
			\includegraphics[width=0.095\linewidth,height=0.1\linewidth]{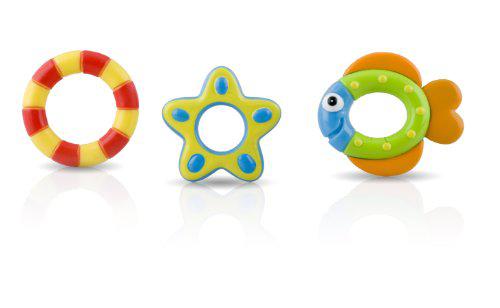}
			\includegraphics[width=0.095\linewidth,height=0.1\linewidth]{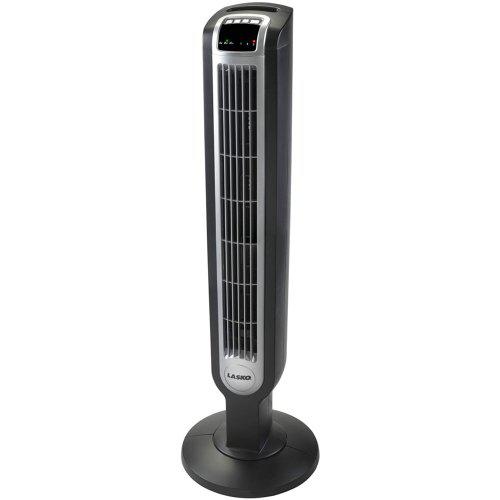}
			\includegraphics[width=0.095\linewidth,height=0.1\linewidth]{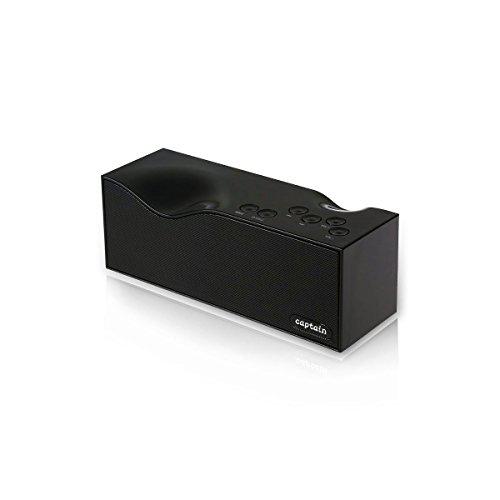}
			\includegraphics[width=0.095\linewidth,height=0.1\linewidth]{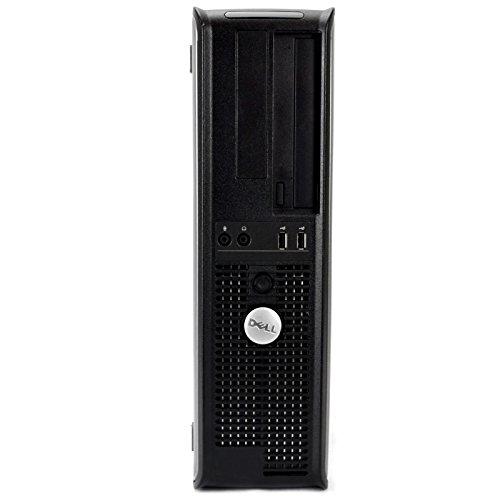}
			\includegraphics[width=0.095\linewidth,height=0.1\linewidth]{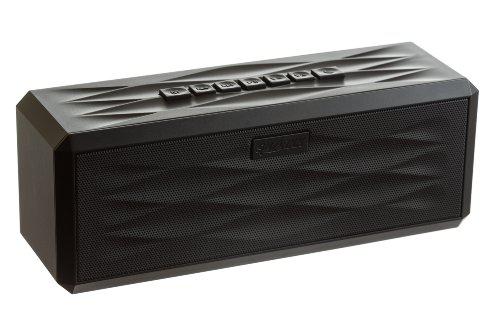}
			\includegraphics[width=0.095\linewidth,height=0.1\linewidth]{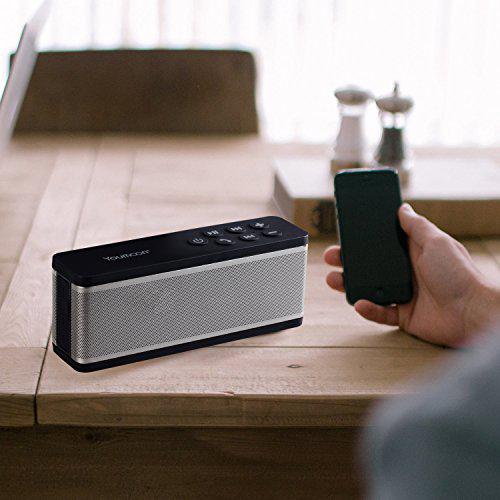}
			\includegraphics[width=0.095\linewidth,height=0.1\linewidth]{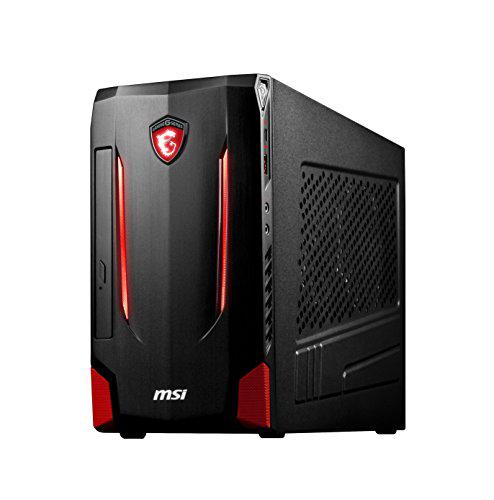}	
			
			\includegraphics[width=0.095\linewidth,height=0.1\linewidth]{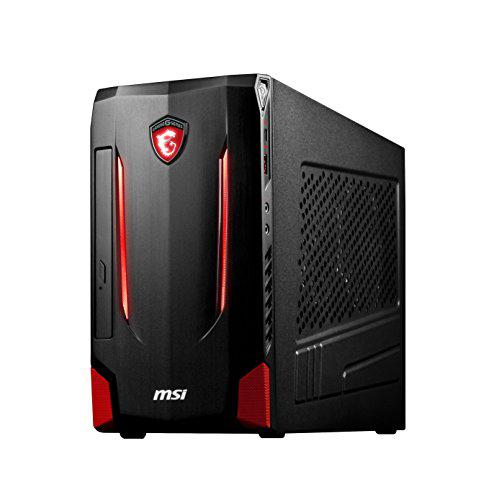}
			\includegraphics[width=0.095\linewidth,height=0.1\linewidth]{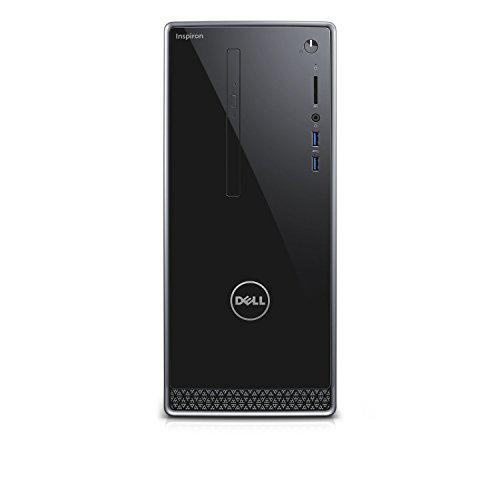}
			\includegraphics[width=0.095\linewidth,height=0.1\linewidth]{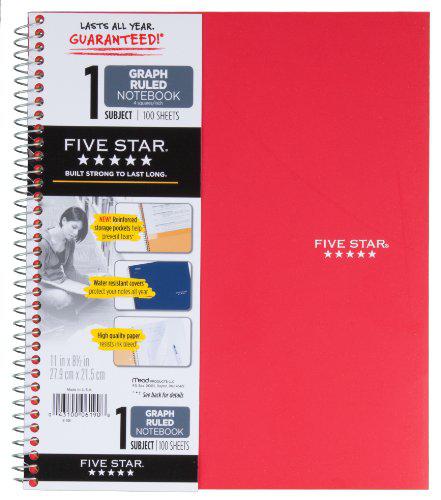}
			\includegraphics[width=0.095\linewidth,height=0.1\linewidth]{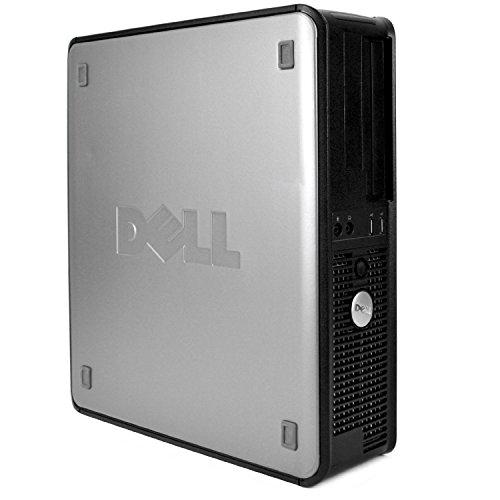}
			\includegraphics[width=0.095\linewidth,height=0.1\linewidth]{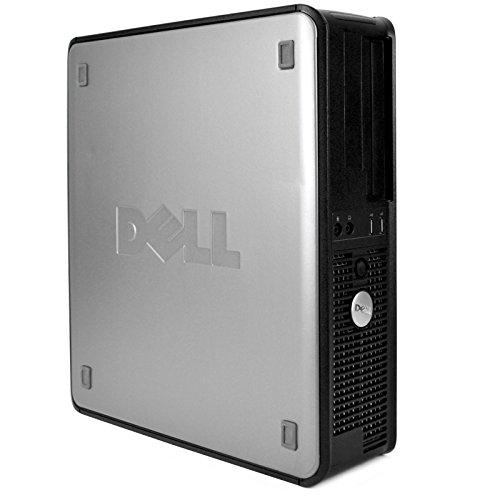}
			\includegraphics[width=0.095\linewidth,height=0.1\linewidth]{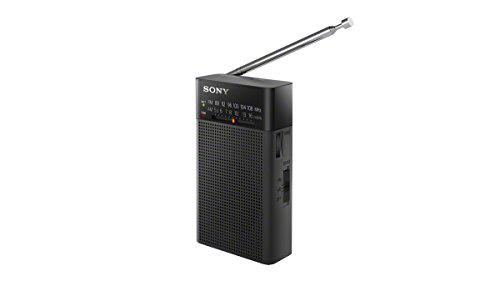}
			\includegraphics[width=0.095\linewidth,height=0.1\linewidth]{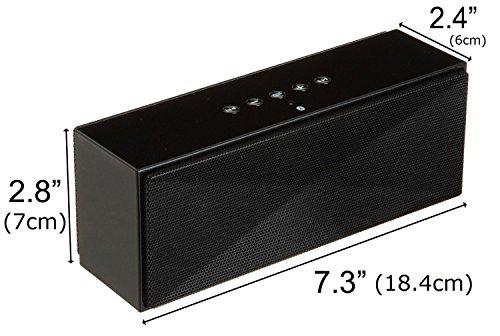}
			\includegraphics[width=0.095\linewidth,height=0.1\linewidth]{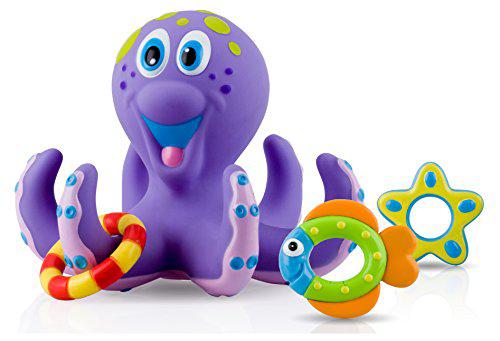}
			\includegraphics[width=0.095\linewidth,height=0.1\linewidth]{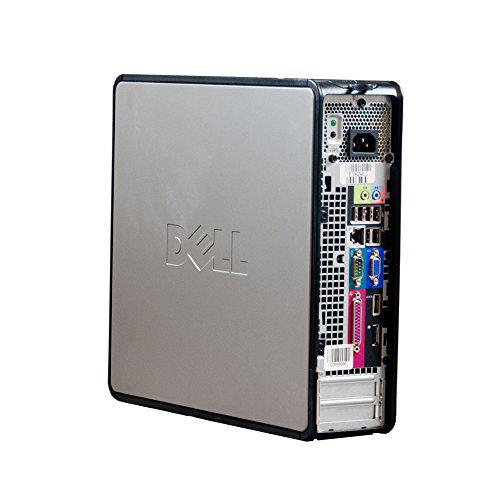}
			\includegraphics[width=0.095\linewidth,height=0.1\linewidth]{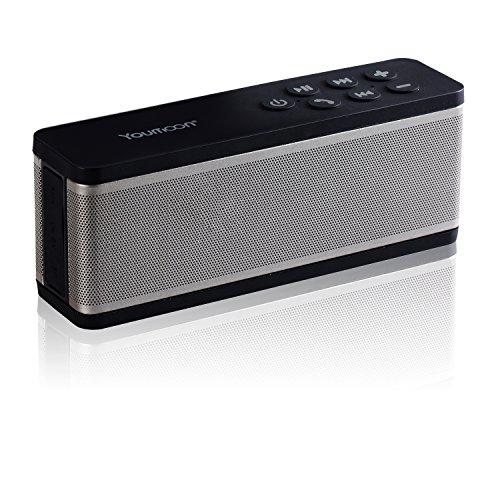}	
		\end{boxedminipage}
	\end{minipage}
	
	\begin{minipage}{\linewidth}
		\begin{minipage}{0.015\linewidth}
			\centering
			\rotatebox{90}{
				\footnotesize
				Queried}
		\end{minipage} 
		\hfill   
		\begin{boxedminipage}{0.99\linewidth}
			\includegraphics[width=0.095\linewidth,height=0.1\linewidth]{fig/Product/Computer_00004.jpg}
			\includegraphics[width=0.095\linewidth,height=0.1\linewidth]{fig/Product/Toys_00008.jpg}
			\includegraphics[width=0.095\linewidth,height=0.1\linewidth]{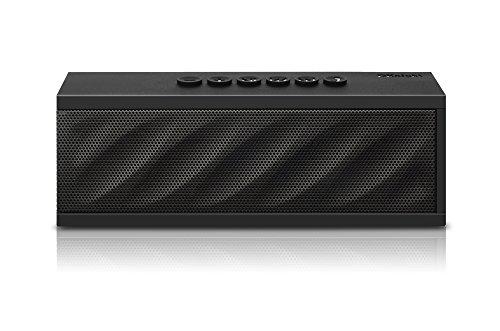}
			\includegraphics[width=0.095\linewidth,height=0.1\linewidth]{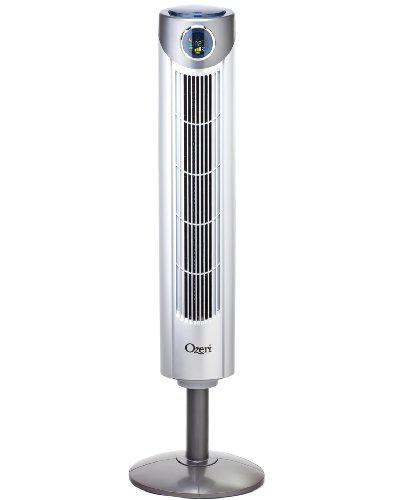}
			\includegraphics[width=0.095\linewidth,height=0.1\linewidth]{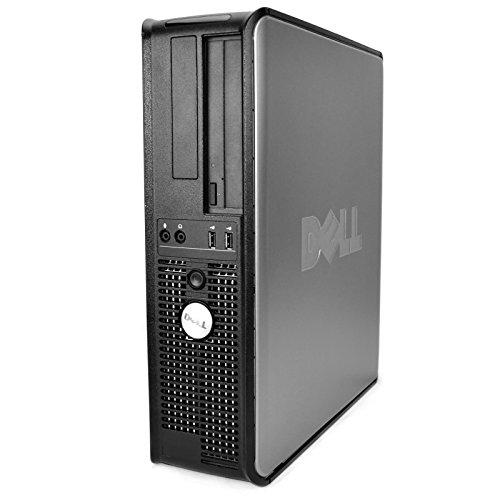}
			\includegraphics[width=0.095\linewidth,height=0.1\linewidth]{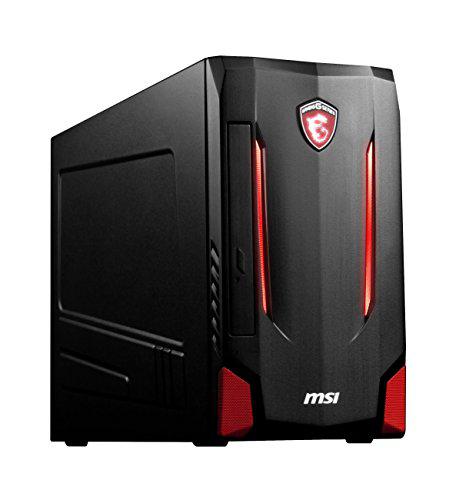}
			\includegraphics[width=0.095\linewidth,height=0.1\linewidth]{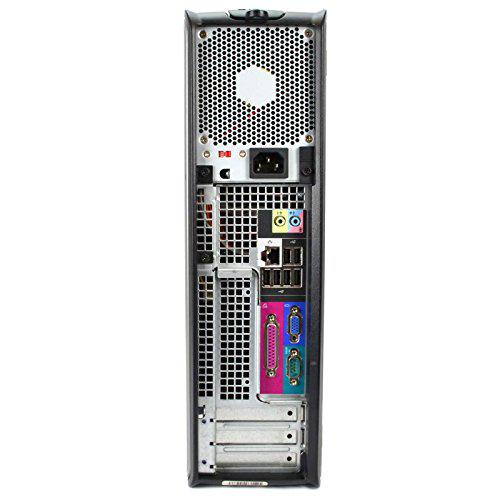}
			\includegraphics[width=0.095\linewidth,height=0.1\linewidth]{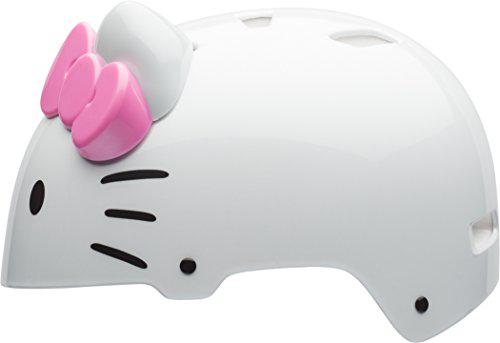}
			\includegraphics[width=0.095\linewidth,height=0.1\linewidth]{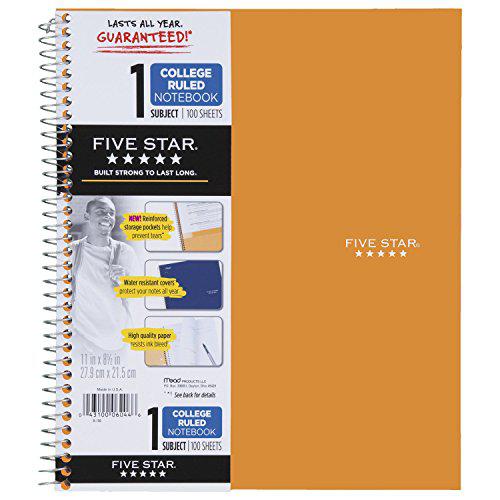}
			\includegraphics[width=0.095\linewidth,height=0.1\linewidth]{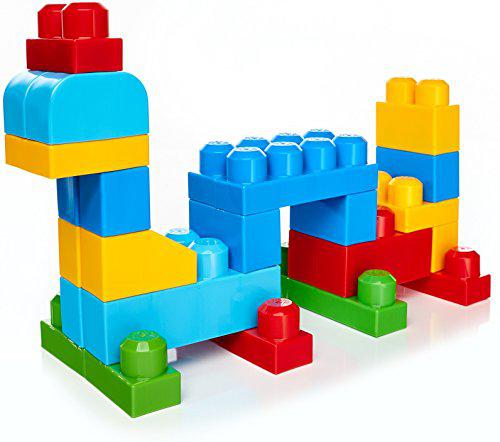}
			
			\includegraphics[width=0.095\linewidth,height=0.1\linewidth]{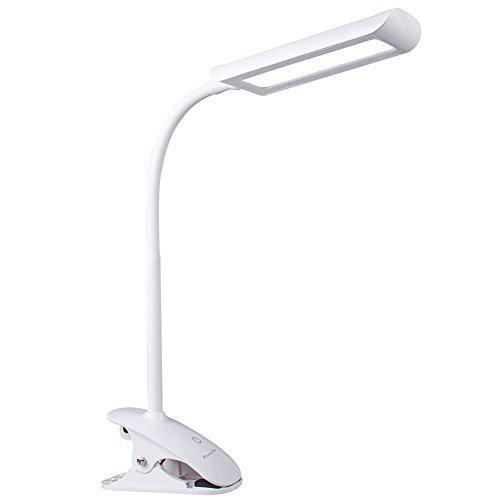}
			\includegraphics[width=0.095\linewidth,height=0.1\linewidth]{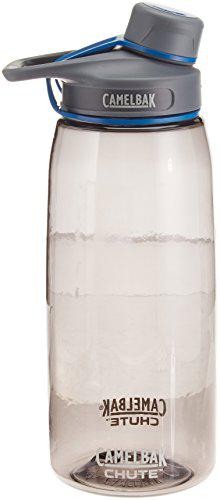}
			\includegraphics[width=0.095\linewidth,height=0.1\linewidth]{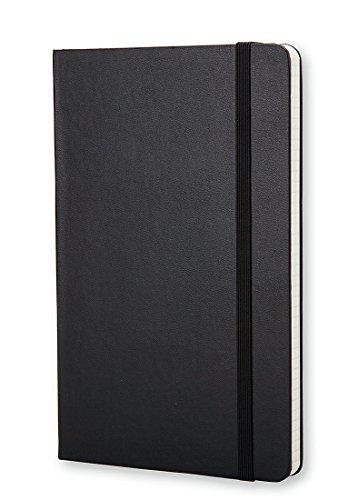}
			\includegraphics[width=0.095\linewidth,height=0.1\linewidth]{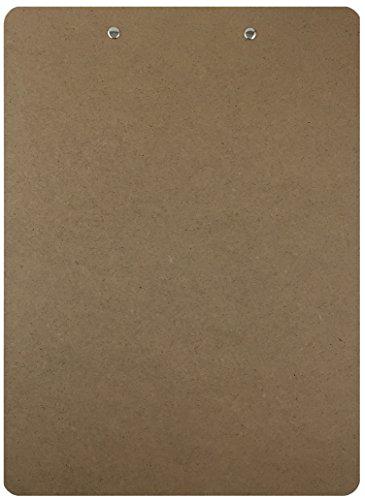}
			\includegraphics[width=0.095\linewidth,height=0.1\linewidth]{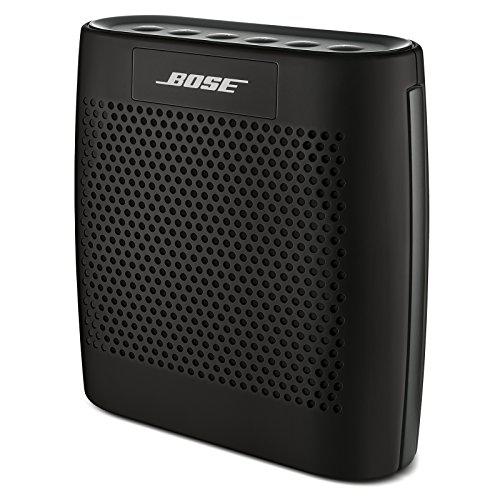}
			\includegraphics[width=0.095\linewidth,height=0.1\linewidth]{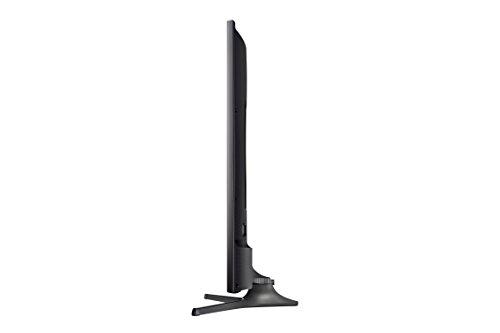}
			\includegraphics[width=0.095\linewidth,height=0.1\linewidth]{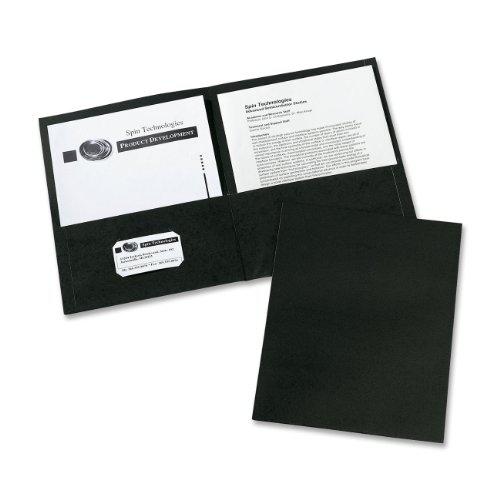}
			\includegraphics[width=0.095\linewidth,height=0.1\linewidth]{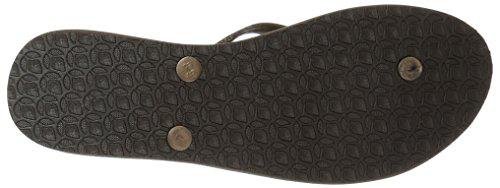}
			\includegraphics[width=0.095\linewidth,height=0.1\linewidth]{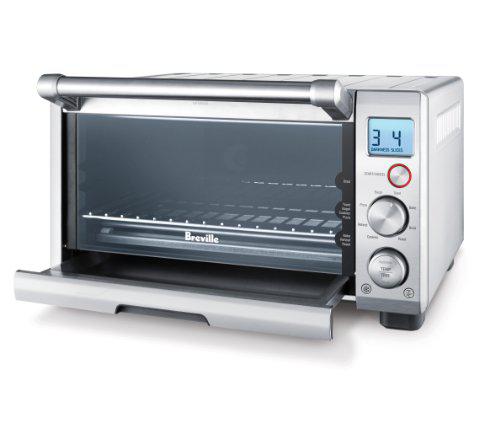}
			\includegraphics[width=0.095\linewidth,height=0.1\linewidth]{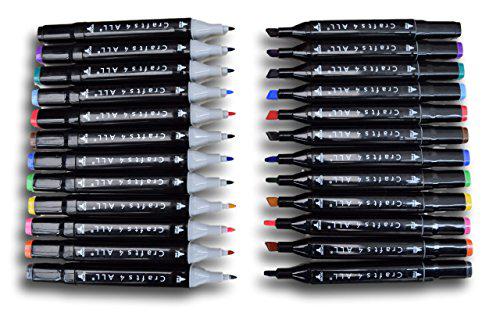}
		\end{boxedminipage}
	\end{minipage}
	
	\caption{Visualization of LAS sampling on Office-Home Ar$\rightarrow$Pr. The first row presents $t$-SNE plots. Squares denote candidate target samples based on LI-scores; stars denote selected target samples for querying labels; and points denote the rest target samples. Each marker is colored according to its (left) ground-truth label and (right) pseudo label from the current model. The last two rows plot top 20 candidate samples and queried samples with largest LI-scores, respectively.}
	\label{fig:sup:ar2pr}
	  \vspace{40mm}
\end{figure*}

\cparagraph{Visualization of LAS.} To visualize how LAS selects target samples, we present $t$-SNE plots of target features on Office-Home Pr$\rightarrow$Ar and Ar$\rightarrow$Pr in Fig.~\ref{fig:sup:pr2ar} and Fig.~\ref{fig:sup:ar2pr}, respectively. We choose the 10th epoch, where 10\% target data are selected as candidates based on LI-scores, of which 1\% target data from cluster centroids are selected for querying labels. Candidate, selected and remaining target samples are marked with squares, stars and points, respectively. The top 20 candidates and queried images are also displayed under the $t$-SNE plots. As can be seen, the candidates (\ie, samples with large LI-scores) generally lie in the regions where model predictions are inconsistent. It is also difficult to distinguish their semantic labels visually, especially for Pr$\rightarrow$Ar, indicating that these images are hard cases. There are some highly similar images in the candidates. After the second step of diverse selection, images selected for querying labels become much more diverse.

{\small
	\bibliographystyle{ieee_fullname}
	\bibliography{ref}
}

\end{document}